%% file: main.tex
\documentclass[sigconf]{acmart}
\usepackage{amsmath,amsfonts} % amssymb
\usepackage[ruled,vlined]{algorithm2e}
\usepackage{graphicx}
\usepackage{textcomp}
\usepackage{xcolor}
\usepackage{tabularx}
\usepackage{caption}
\usepackage{subcaption}
\usepackage[inline]{enumitem}

\newcommand{\X}{$\times$}
\usepackage{bm}
\usepackage{bbm}
\usepackage{url}
\usepackage{amsthm}
\usepackage{wrapfig}
\usepackage{multicol}
\usepackage{multirow}

\SetCommentSty{mycommfont}

\usepackage{xspace}
\newcommand{\Method}{LifeHD\xspace}
\newcommand{\SemiMethod}{LifeHD$_\textrm{semi}$\xspace}
\newcommand{\EffMethod}{LifeHD$_\textrm{a}$\xspace}
\newcommand{\prototype}{cluster HV\xspace}
\newcommand{\prototypes}{cluster HVs\xspace}
\newcommand{\Prototype}{Cluster HV\xspace}

\newcommand{\AccRes}{74.8\%}
\newcommand{\EnergyRes}{34.3x}
\newcommand*{\revise}{\textcolor{black}}

 \renewcommand\footnotetextcopyrightpermission[1]{}

\input{preamble}

\AtBeginDocument{%
  \providecommand\BibTeX{{%
    \normalfont B\kern-0.5em{\scshape i\kern-0.25em b}\kern-0.8em\TeX}}}

\setcopyright{acmcopyright}
\copyrightyear{2023}
\acmYear{2023}
\acmDOI{XXXXXXX.XXXXXXX}

\acmConference[Under review]{Make sure to enter the correct
  conference title from your rights confirmation emai}{}{}
\acmPrice{15.00}
\acmISBN{978-1-4503-XXXX-X/18/06}

\begin{document}

\title{Lifelong Intelligence Beyond the Edge using Hyperdimensional Computing}

\author{Xiaofan Yu}
\email{x1yu@ucsd.edu}
\orcid{0000-0002-9638-6184}
\affiliation{%
  \institution{University of California San Diego}
  \city{La Jolla}
  \state{California}
  \country{USA}
}

\author{Anthony Thomas}
\email{ahthomas@ucsd.edu}
\affiliation{%
  \institution{University of California San Diego}
  \city{La Jolla}
  \state{California}
  \country{USA}
}

\author{Ivannia Gomez Moreno}
\email{ivannia.gomez@cetys.edu.mx}
\affiliation{%
  \institution{CETYS University, Campus Tijuana}
  \city{Tijuana}
  \country{Mexico}
}

\author{Louis Gutierrez}
\email{l8gutierrez@ucsd.edu }
\affiliation{%
  \institution{University of California San Diego}
  \city{La Jolla}
  \state{California}
  \country{USA}
}

\author{Tajana \v{S}imuni\'{c} Rosing}
\email{tajana@ucsd.edu}
\orcid{0000-0002-6954-997X}
\affiliation{%
  \institution{University of California San Diego}
  \city{La Jolla}
  \country{USA}
}


%HDC is a lightweight and hardware-friendly learning paradigm inspired by brain functionalities.

\begin{abstract}
  On-device learning has emerged as a prevailing trend that avoids the slow response time and costly communication of cloud-based learning.
  The ability to learn continuously and indefinitely in a changing environment, and with resource constraints, is critical for real sensor deployments.
  However, existing designs are inadequate for practical scenarios with (i) streaming data input, (ii) lack of supervision and (iii) limited on-board resources. 
  In this paper, we design and deploy the first on-device lifelong learning system called {\Method} for general IoT applications with limited supervision.
  {\Method} is designed based on a novel neurally-inspired and lightweight learning paradigm called Hyperdimensional Computing (HDC).
  We utilize a two-tier associative memory organization to intelligently store and manage high-dimensional, low-precision vectors, which represent the historical patterns as cluster centroids.
  We additionally propose two variants of {\Method} to cope with scarce labeled inputs and power constraints.
  We implement {\Method} on off-the-shelf edge platforms and perform extensive evaluations across three scenarios. Our measurements show that {\Method} improves \revise{the unsupervised clustering accuracy by up to {\AccRes} compared to the state-of-the-art NN-based unsupervised lifelong learning baselines with as much as {\EnergyRes} better energy efficiency.
  Our code is available at \url{https://github.com/Orienfish/LifeHD}.}
\end{abstract}
\keywords{Edge Computing, Lifelong Learning, Hyperdimensional Computing}



\maketitle

%%%%%%%%%%%%%%%%%%%%%%%%%%%%%%%%%%%%%%%%%%%%%%%%%%%%%%%
% Introduction
%%%%%%%%%%%%%%%%%%%%%%%%%%%%%%%%%%%%%%%%%%%%%%%%%%%%%%%
\section{Introduction}

%Along with the rapid expanding speed of edge computing market in recent years~\cite{edge_market}
%The global edge computing market size has been expanding exponentially in recent years,
%Edge intelligence enables in-place learning and quick decision making, while saving the communication cost and preserving user privacy, as edge devices only need to send features/predicted results rather than raw data.

%On-device learning refers to learning that takes place directly on the edge devices, thereby reducing communication cost and improving privacy, by transmitting only summary statistics or prediction results instead of raw data.

The fusion of artificial intelligence and Internet of Things (IoT) has become a prominent trend with numerous real-world applications, such as in smart cities~\cite{chen2016smart}, smart voice assistants~\cite{sun2020alexa}, and smart activity recognition~\cite{weiss2016smartwatch}. However, the predominant current approach is cloud-centric, where sensor devices send data to the cloud for offline training using extensive data sources. This approach faces challenges like slow updates and costly communication, involving the exchange of large sensor data and models between the edge and the cloud~\cite{shunhou2022aiot}.
%However, the prevailing approach in today's AIoT systems is cloud-centralized, where IoT devices transmit data to the cloud for offline training using extensive data sources. %~\cite{su2021aiot,shunhou2022aiot}.
%Such scheme suffers from multiple drawbacks.
%Firstly, transmitting large volumes of data over potentially unreliable wireless networks leads to long communication delays, data loss, and high energy costs, especially for battery-powered devices~\cite{suryavansh2021data}.
%Most importantly, for long-term AIoT deployment, the model requires regular retraining to adapt to dynamically changing environments. However, achieving end-to-end updates within the cloud-centralized scheme requires gathering data from all IoT devices and involves time-consuming offline training.
Instead, recent research has shifted towards edge learning, where machine learning is performed on resource-constrained edge devices right next to the sensors. While most studies focused on inference-only tasks~\cite{lin2020mcunet,lin2021memory,saha2023tinyns}, some recent work has investigated the optimization of computational and memory resources for on-device training~\cite{gim2022memory,lin2022device}. Nevertheless, these efforts often rely on static models for inference or lack the adaptability to accommodate new environments.
%compressing large neural network models~\cite{cai2019once,han2019deep,lin2020mcunet,lin2021memory,saha2023tinyns} or utilizing emerging computing methods~\cite{bakar2022adaptive} to make them suitable for devices as small as microcontrollers.

To fundamentally address these issues, sensor devices should be capable of "lifelong learning"~\cite{parisi2019continual}: to \textit{learn and adapt with limited supervision} after deployment. 
\revise{On-device lifelong learning reduces the need for expensive data collection (including labels) and offline model training, operating in a deploy-and-run manner. This approach enables autonomous learning solely from the incoming samples with minimal supervision, and is thus able to provide real-time decision-making even without a network connection.}
%data transmission by sending only summary statistics or prediction results, facilitating real-time decision-making even without a network connection.
The lifelong aspect is essential for handling dynamic real-world environments, representing the future of IoT.
Although extensive research has investigated lifelong learning across various scenarios~\cite{parisi2019continual}, existing techniques face challenges that render them unsuitable for real-world deployments. These challenges include:
% (for example, supervised~\cite{kirkpatrick2017overcoming} and unsupervised lifelong learning~\cite{fini2021self,madaan2022representational}),
% supervised~\cite{kirkpatrick2017overcoming,lopezpaz2017gradient,chaudhry2018efficient,guo2020improved}
% unsupervised lifelong learning~\cite{jiang2017variational,achille2018life,wu2018memory}

\begin{enumerate}[label=(C\arabic*),noitemsep,nolistsep]
%\vspace{-2mm}
  \item \textbf{Streaming data input.} Edge devices collect streaming data from a dynamic environment.  %Hence in a short amount of time, only one type of data (e.g., the same class) may be observed. 
  This \textit{online} learning with \textit{non-iid} data contrasts with the default \textit{offline} and \textit{iid} setting where multiple passes on the entire dataset are allowed~\cite{grill2020bootstrap}. % chen2020simple,chen2021exploring,gidaris2020online
  
  \item \textbf{Lack of supervision.} Obtaining ground-truth labels and expert guidance is often challenging and expensive. Most lifelong learning methods rely on some form of supervision, such as class labels~\cite{kirkpatrick2017overcoming} or class shift boundaries~\cite{rao2019continual}, which are typically unavailable in real-world scenarios.
  % class shift boundaries~\cite{,he2021unsupervised,taufique2022unsupervised}
  
  %Due to limited on-board memory, the collected samples are fed for learning \textit{only once}. 
  \item \textbf{Limited device resources.} Neural networks (NN) are known for their high resource demands~\cite{wang2019deep}. Furthermore, the main techniques for lifelong learning based on NN, such as regularization~\cite{kirkpatrick2017overcoming} and memory replay~\cite{lopez2017gradient}, add extra computational and memory requirements beyond standard NNs, making them inadequate for edge devices.
  %(i.e., regularization~\cite{kirkpatrick2017overcoming,zenke2017continual,aljundi2018memory,zhang2020class}, distillation~\cite{lin2021continual,fini2021self}, expandable architecture~\cite{rusu2016progressive,ostapenko2019learning,Lee2020A} and experience or generative replay~\cite{rebuffi2017icarl,lopezpaz2017gradient,chaudhry2018efficient,buzzega2020dark,tiwari2022gcr}) 
\end{enumerate}
% regularization: ritter2018online,ahn2019uncertainty,yu2020semantic,
% replay: wang2022memory
% expandable arch: von2020continual,rajasegaran2020itaml,abati2020conditional

\begin{figure}[t]
  \centering
  \includegraphics[width=0.4\textwidth]{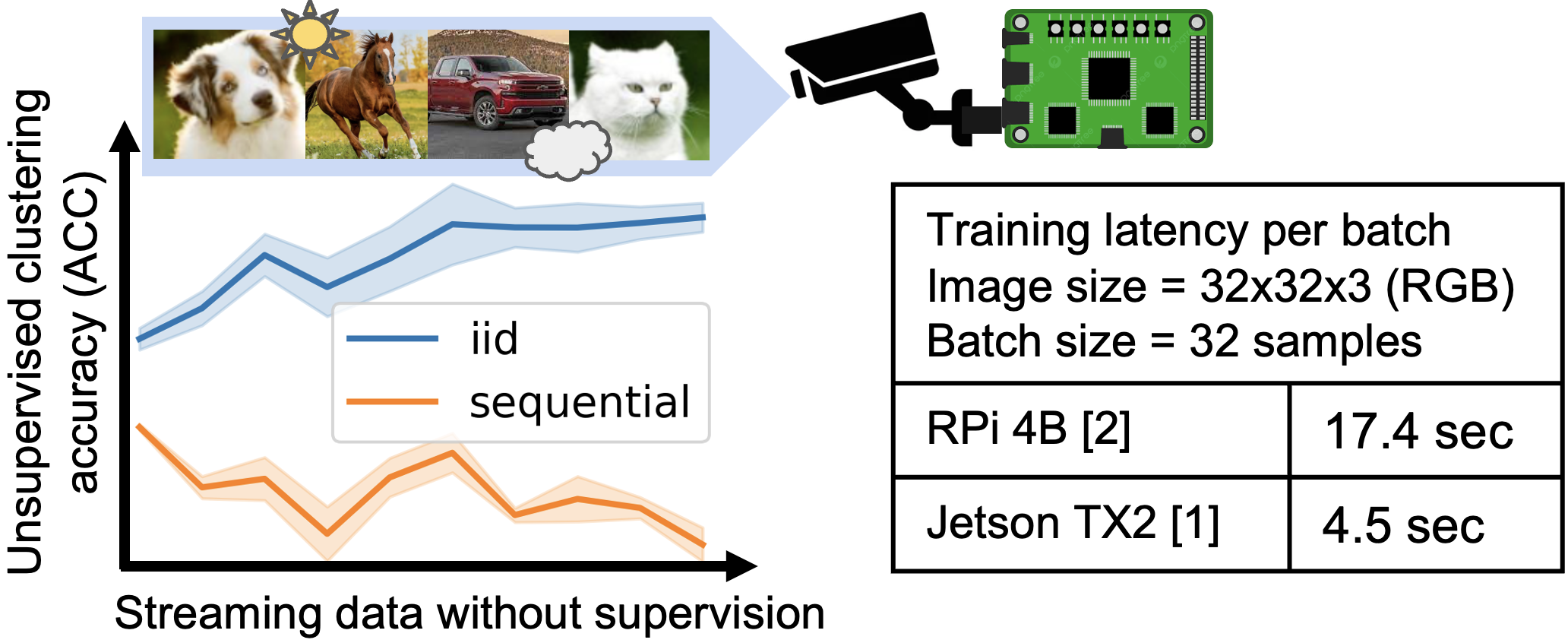}
  \vspace{-4mm}
  \caption{\small Real-world example of on-device lifelong learning evaluated using the unsupervised clustering accuracy metric~\cite{xie2016unsupervised}. The training latency is measured on two typical edge platforms.}
  \vspace{-5mm}
  \label{fig:moti_intro}
\end{figure}

\noindent \textbf{Real-World Example.}
To illustrate the challenges faced, we present a real-world scenario in Fig.\ref{fig:moti_intro}. 
\revise{Consider a camera deployed in the wild continuously collecting data from surrounding environment. Our goal is to train an unsupervised object recognition algorithm on the edge device, purely from the data stream.}
We construct both iid and sequential (one class appears after the other) streams from CIFAR-100~\cite{misc_mhealth_dataset_319}, and adopt the smallest MobileNet V3 model~\cite{howard2019searching} with the popular BYOL unsupervised learning pipeline~\cite{grill2020bootstrap}.
%The sequential stream mimics the data collection in real deployments.
As seen in Fig.~\ref{fig:moti_intro}, while the model shows improved accuracy with iid streams, it has a significant performance loss under sequentially ordered data,
highlighting the NN effect of ``forgetting'' in a streaming and unsupervised setting.
In terms of efficiency, we measure the training latency of MobileNet V3 (small)~\cite{howard2019searching} on two typical edge platforms, Raspberry Pi (RPi) 4B~\cite{rpi4b} and Jetson TX2~\cite{jetsontx2} by running 10 gradient descent steps on a single batch of 32 samples.
Even on these very capable edge platforms, training takes up to 17.4 seconds, clearly unsuitable for real-time processing under 30 FPS. Therefore, a novel approach capable of handling non-iid data and offering more efficient updates is necessary to accommodate the continual changes in data.

To address challenges (C1)-(C3), we draw inspiration from biology, where even tiny insects display remarkable lifelong learning abilities, and do so using ``hardware'' that requires very little energy~\cite{avargues2012simultaneous}. 
Hyperdimensional computing (HDC) is an emerging paradigm inspired by the information processing mechanisms found in biological brains~\cite{kanerva2009hyperdimensional}.
In HDC, all data is represented using high-dimensional, low-precision (often binary) vectors known as ``hypervectors,'' which can be manipulated through simple element-wise operations to perform tasks like memorization and learning. HDC is well-understood from a theoretical standpoint~\cite{thomas2021theoretical} and shares intriguing connections with biological lifelong learning~\cite{shen2021algorithmic}.
Furthermore, its use of basic element-wise operators aligns with highly parallel and energy-efficient hardware, offering substantial energy savings in IoT applications~\cite{kim2018efficient,imani2017voicehd,dutta2022hdnn,xu2023fsl}.
%HDC has been found effective in supervised IoT applications, requiring minimal labeled data~\cite{xu2023fsl}, and offers substantial energy savings on emerging hardware~\cite{dutta2022hdnn}.
While HDC is reported as a promising avenue, the literature to date has not explored weakly-supervised lifelong learning using HDC.

In this work, we design and deploy {\Method}, the \textit{first} system for on-device lightweight lifelong learning in an unsupervised and dynamic environment.
{\Method} leverages HDC's efficient computation and advantages in lifelong learning, while effectively handling unlabeled streaming inputs. These capabilities extend beyond the scope of existing HDC designs, which have focused overwhelmingly on the supervised setting~\cite{kim2018efficient,imani2017voicehd}.
%Following the general principle of HDC, 
Specifically, {\Method} represents the input as high-dimensional, low-precision vectors, and, drawing inspiration from work in cognitive science~\cite{baddeley1992working}, organizes data into a two-tier memory hierarchy: a short-term ``working memory'' and a long-term memory. The working memory processes incoming data and summarizes it into a group of fine-grained clusters that are represented by hypervectors called \textit{\prototypes}. Long-term memory consolidates the frequently appeared \prototypes in the working memory, and will be retrieved for merging and inference occasionally.
We emphasize that \Method is designed to suit a variety of edge devices with diverse resource levels. More efficiency gains can be achieved by employing optimizations such as pruning and quantization~\cite{wang2022melon,gim2022memory}, but this is not the focus of our work.

Our basic approach in {\Method} is fully unsupervised. However, in reality, labels may be available (or could be acquired) for a small number of examples. 
%We consider an extension to {\Method}, that we call {\SemiMethod}, which can exploit a small number of labeled examples to improve performance over the purely unsupervised {\Method}. 
We introduce \SemiMethod to exploit a limited number of labeled samples as an extension to the purely unsupervised \Method. 
Additionally, we propose {\EffMethod}, which uses an adaptive scheme inspired by model pruning, to adjust the HD embedding dimension on-the-fly. {\EffMethod} allows us to further reduce resource usage (power in-particular), where necessary.

%We further propose two variants of {\Method} named {\SemiMethod} and {\EffMethod} to deal with the practical IoT deployments with scarce labeled samples and power constraints respectively. {\SemiMethod} adjusts the \prototypes learned by {\Method} with limited labels coming with the stream. {\EffMethod} offers an extra adaptive masking scheme inspired by model pruning to alter the dimension of HDC on the fly, excelling in efficiency with little accuracy loss. 

%\item We first conduct a motivating study of unsupervised lifelong learning on Stream-51, the latest video-based dataset for classification. The empirical results reveal that existing NN-based approaches are unsatisfactory for real-world deployments under sequential data input and resource constraints.

In summary, the contributions of this paper are:
\begin{enumerate}[noitemsep,nolistsep]
    \item We design {\Method}, the first end-to-end system for on-device unsupervised lifelong intelligence using HDC. 
    {\Method} builds upon HDC's lightweight single-pass training capability and incorporates our novel clustering-based memory design to address challenges (C1)-(C3). %{\Method} excels in efficiency while allowing unsupervised lifelong learning not achieved by previous HDC methods.
    \item We further propose {\SemiMethod} as an extension to fully utilize the scarce labeled samples along with the stream. We devise {\EffMethod} that enables adaptive pruning in {\Method} to reduce real-time power consumption.
    \item We implement {\Method} on off-the-shelf edge devices and conduct extensive experiments across three typical IoT scenarios. {\Method} improves \revise{the unsupervised clustering accuracy up to {\AccRes} with {\EnergyRes} better} energy efficiency compared to leading unsupervised NN lifelong learning methods~\cite{madaan2022representational,fini2021self,smith2019unsupervised}.
    \item {\SemiMethod} improves the unsupervised clustering accuracy by up to 10.25\% over the SemiHD~\cite{imani2019semihd} baseline under limited label availability. {\EffMethod} limits the accuracy loss within 0.71\% using only 20\% of \Method's full HD dimension.

\end{enumerate}

The rest of the paper is organized as follows.
We start by a comprehensive review of related works in Sec.~\ref{sec:related-work}.
We then introduce salient background on HDC in Sec.~\ref{sec:hdc} to help understanding.
We formally define the unsupervised lifelong learning problem we target to solve in Sec~\ref{sec:problem}.
Afterwards, Sec.~\ref{sec:method} describes the details of our major design {\Method}.
Sec.~\ref{sec:variants} introduces {\SemiMethod} and {\EffMethod}.
Sec.~\ref{sec:evaluation-main} presents the implementation and results of {\Method},
while the evaluations of {\SemiMethod} and {\EffMethod} are reported in Sec~\ref{sec:evaluation-variants}.
\revise{We add the discussions and future works in Sec.~\ref{sec:discussion}.} The entire paper is concluded in Sec.~\ref{sec:conclusion}.

%%%%%%%%%%%%%%%%%%%%%%%%%%%%%%%%%%%%%%%%%%%%%%%%%%%%%%%
% Related Work
%%%%%%%%%%%%%%%%%%%%%%%%%%%%%%%%%%%%%%%%%%%%%%%%%%%%%%%
\vspace{-2mm}
\section{Related Work}
\label{sec:related-work}

%In this section, we review the most relevant literature from lifelong learning, on-device training and HDC.

%\subsection{On-Device Learning}
%Offloading NNs from the cloud to resource constrained devices has attracted significant interest, starting a new research trajectory that has come to be known as Tiny Machine Learning (TinyML)~\cite{disabato2022tiny}.
%TinyML advocates for deploying NNs on devices as small as microcontrollers that consume only a few milliwatts of power. Researchers have used techniques such as model compression, weight quantization and pruning to reduce model size without a substantial sacrifice in accuracy~\cite{cai2019once,han2019deep,lin2020mcunet,lin2021memory,saha2023tinyns}.
%Recent efforts also explored new computing paradigms such as tsetlin machine for energy-harvesting devices~\cite{bakar2022adaptive}. 
%However, the only considers inference on tiny devices, while we emphasize on-device adaptive and continual training by proposing {\Method}.

\textbf{Lifelong and On-Device Learning.}
Lifelong learning (or continual learning) is a large and active area of research in the broader machine learning community. 
%Catastrophic forgetting is the main challenge in lifelong learning, that is, updating the model using new samples degrades the knowledge learned in the past~\cite{mccloskey1989catastrophic}.
% \cite{goodfellow2013empirical}
Catastrophic forgetting is a major challenge in lifelong learning, and refers to a commonly observed empirical phenomenon in which updating certain machine learning models with new data severely degrades their ability to perform previously learned tasks~\cite{mccloskey1989catastrophic}.
Previous works proposed techniques such as dynamic architecture~\cite{rusu2016progressive,Lee2020A}, regularization by penalizing important weights~\cite{kirkpatrick2017overcoming,zhang2020class}, knowledge distillation from past models~\cite{fini2021self} and experience replay using a memory buffer~\cite{lopez2017gradient,tiwari2022gcr}.
The lifelong learning literature has examined a wide range of problem settings, ranging from the fully supervised case, in which tasks and class labels are provided, and the fully unsupervised case without any labels and prior knowledge~\cite{madaan2022representational,ijcai2022p0483}.
However, all of these works are based on deep NNs and require backpropagation, which is problematic for resource-constrained devices. 

%TinyOL~\cite{ren2021tinyol}}

Neurally-inspired lightweight algorithms have recently been proposed for lifelong learning applications.  
%Inspired by the ability of biological learning, there has recently been interest in using neurally inspired approaches for lifelong learning.
FlyModel~\cite{shen2021algorithmic} and SDMLP~\cite{bricken2023sparse} use sparse coding and associative memory for lifelong learning. %Justified by theoretical proof and empirical results, they showed the simple combination of sparse coding and associative learning leads to strong continual learning performance.
%SDMLP~\cite{bricken2023sparse} used Sparse Distributed Memory (SDM) to modify a Multi-Layered Perceptron (MLP) with features from a cerebellum-like neural circuit. 
However, both approaches assume full supervision.
%mainly focus on the theoretical aspect, thus are not compatible with IoT deployments with weak supervision.
%Nevertheless, STAM was originally designed for image classification and consume approximately 891KB (465KB) memory footprint for the short-term (long-term) storage. 
STAM~\cite{smith2019unsupervised} is an expandable memory architecture for unsupervised lifelong learning, using layered receptive fields and a two-tier memory hierarchy. It learns via online centroid-based clustering pipeline, novelty detection and memory updates. 
%STAM architecture requires large memory capacity, equivalent to that needed to store 1800 gray-scale images,
Nevertheless, the memory in STAM is solely dedicated to image storage, while our {\Method} additionally emphasizes merging past patterns into coarse groups and shows more effective learning performance.

% \subsection{On-Device Training}

Recent works optimize the resource usage of on-device training via pruning and quantization~\cite{profentzas2022minilearn,lin2022device}, tuning partial weights~\cite{cai2020tinytl,ren2021tinyol}, memory profiling and optimization~\cite{wang2022melon,xu2022mandheling,gim2022memory}, as well as growing the NN on the fly~\cite{zhang2020mdldroidlite}.
All these works optimize training given resource constraints and do not focus on lifelong learning.  They are orthogonal to the contribution of {\Method} which focuses on adaptive and continual training.  {\Method} can be further optimized by combining with such techniques.

%As such , which is orthogonal to the contribution of {\Method} to adaptive and continual training.

%While most of the existing literature only considered inference on tiny devices~\cite{cai2019once,han2019deep,lin2020mcunet,lin2021memory,saha2023tinyns,bakar2022adaptive},
%several works have studied training on IoT devices 
% TinyOL~\cite{ren2021tinyol} finetunes an additional layer to adapt to the changing environments in a supervised manner.

\smallskip \noindent
\textbf{Hyperdimensional Computing.}
HDC has garnered substantial interest from the computer hardware community as an energy-efficient and low-latency approach to learning, and has been successfully applied to problems such as human activity recognition~\cite{kim2018efficient}, voice recognition~\cite{imani2017voicehd}, image recognition~\cite{dutta2022hdnn,xu2023fsl}, to name a few. The large majority of literature on HDC has focused on using the technique to perform supervised classification tasks.
Among the limited literature for weakly-supervised learning with HDC, HDCluster~\cite{imani2019hdcluster} enabled unsupervised clustering in HDC with a new algorithm that is similar to K-Means.
SemiHD~\cite{imani2019semihd} is a semi-supervised learning framework using HDC with iterative self-labeling.
Hyperseed~\cite{osipov2022hyperseed}, C-FSCIL~\cite{hersche2022constrained} and FSL-HD~\cite{xu2023fsl} adopted HDC or similar vector symbolic architectures (VSA) for unsupervised or few-shot learning.
All above works did not consider the lifelong aspect and used offline training on a static dataset.
To the best of the authors' knowledge, {\Method} is the first work that designs and deploys lifelong learning in edge IoT applications especially with zero or minimal amount of labels. 
%{\Method} adopts HDC due to HDC's lightweight nature and advantages in preventing forgetting.

%\subsection{Federated Lifelong Learning}
%Compared to single-device lifelong learning, the federated version increases difficulty as the data streams on different clients might differ.
%BalanceFL~\cite{shuai2022balancefl} spotted the global and local data imbalance in Federated Learning. The authors proposed balanced sampling, feature-level data augmentation, smooth regularization to avoid catastrophic forgetting under missing local classes.

%%%%%%%%%%%%%%%%%%%%%%%%%%%%%%%%%%%%%%%%%%%%%%%%%%%%%%%
% Background on HDC
%%%%%%%%%%%%%%%%%%%%%%%%%%%%%%%%%%%%%%%%%%%%%%%%%%%%%%%
\section{Background on HDC}
\label{sec:hdc}

Hyperdimensional Computing (HDC) is an emerging paradigm for information processing from the cognitive-neuroscience literature \cite{kanerva2009hyperdimensional}. 
%The field is motivated by a wide body of work in empirical and theoretical neuroscience which has established high-dimensional and distributed representations as a fundamental data type for neural information processing \cite{thomas2021theoretical}. 
In HDC, all computation is performed on low-precision and distributed representations of data that accord naturally with highly parallel and low-energy hardware. % \cite{karunaratne2020memory}. 

The first step in HDC is encoding, which maps an input $x \in \mathcal{X}$ to a distributed representation $\phi(x)$ living in some $D$-dimensional inner-product space $\H$, that we call the ``HD-space.'' For instance, one might take $\H \subset \{\pm 1\}^{D}$, or $\H \subset \R^{D}$. 
%Hyperdimensional Computing is inspired by brain activities where dense sensory input is encoded into high-dimensional vectors~\cite{kanerva2009hyperdimensional}.
%We adopt the commonly used bipolar hypervectors (each bit is from $\{ +1, -1 \}$) to maximize hardware efficiency. 
%Suppose $D$ is the dimension of hypervectors.
%Suppose $\theta$ is the encoding function, $d$ is the dimension of raw samples, and $D$ is the dimension of hypervectors, then we have $H_{t,i} = \theta(\mathbf{x}_{t,i}), \mathbf{x}_{t,i} \in \mathbb{R}^d, H_{t,i} \in \{0,1\}^D$.
We refer to points in the HD-space as \textit{hypervectors}.
Encodings of data can be manipulated so as to build more complex composite representations using a set of operators defined as follows:
\begin{enumerate}
    \item \textit{Bind}: $\otimes : \H \times \H \to \H$. Binding takes two hypervectors as inputs and returns a hypervector that is dissimilar to both inputs, and is intuitively used to represent tuples. For bipolar hypervectors (i.e., $\H \subset \{\pm 1\}^{d}$), the binding operator is typically element-wise multiplication.
    \item \textit{Bundle}: $\oplus : \H \times \H \to \H$. Bundling takes two hypervectors as input and returns a hypervector similar to both operands, and is intuitively used to build sets. The bundling operation is implemented through addition.
    \item \textit{Permute}: $\rho : \H \to \H$. Permutation can be used to encode sequential information and is typically implemented using a cyclic shift. 
\end{enumerate}
%Using the above operations, the HDC pipeline, encompassing encoding, training, and inference, is depicted in Fig.~\ref{fig:hd_pipeline}. In the subsequent lines, we provide a more detailed explanation of each step.

\noindent
The encoding function $\phi : \X \to \H$ embeds data from its ambient representation into HD-space. In general, encoding should preserve some meaningful notion of similarity between input points in the sense that $\phi(x)\cdot\phi(x') \approx k(x,x')$, where $k$ is some similarity function of interest on $\X$.
% employ two encoding methods for various datasets. To encode a time series of sensor data, e.g., collected for human activities monitoring,
In this paper, we use spatiotemporal encoding for time series sensor data, and HDnn for more complex data, such as images, which we explain in the following.
% spatial-temporal encoding:~\cite{,zhou2021memory,menon2022efficient}

%Encoding is the first and foremost step in HDC because it determines the manifold in the high-dimensional space.
%Ideally, we want to effectively extract useful patterns from raw time-series sensor readings thus the hypervectors encoded from similar raw samples should also enjoy higher similarity.
%Let $f$ denote the encoding function. For simplicity, in this section, we consider a single sample $\mathbf{x} \in \mathbb{R}^{T \times d}$ and an encoded hypervector $H \in \{0,1\}^D$. The encoding step can be represented as $H = f(\mathbf{x})$. %, where $\mathbf{x}_{t,i} \in \mathbb{R}^d$, $d$ is the dimension of original samples, and $H_{t,i} \in \{0,1\}^D$.

\begin{figure}[t]
  \centering
  \includegraphics[width=0.49\textwidth]{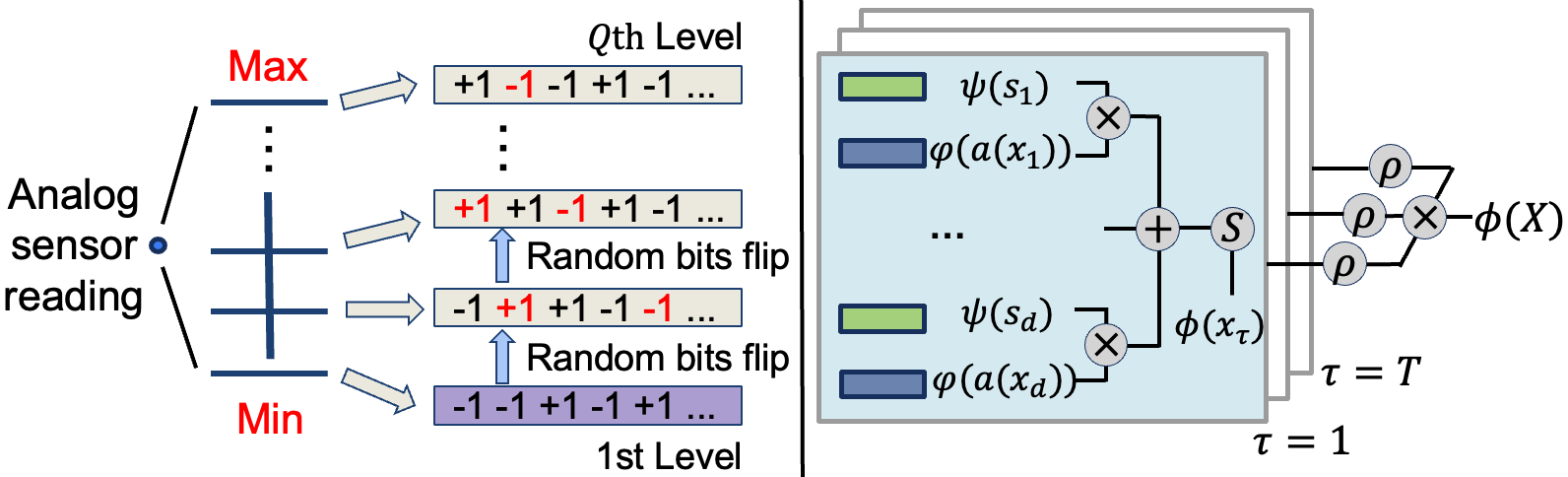}
  \vspace{-6mm}
  \caption{\small Spatiotemporal HDC encoding for time-series data. Left: random generation of level hypervectors. Right: the complete encoding process.}
  \vspace{-5mm}
  \label{fig:encoding}
\end{figure}

\textbf{Spatiotemporal Encoding.} \revise{The spatiotemporal method~\cite{moin2021wearable} jointly encodes the analog information from each sensor (spatial) and at each time stamp (temporal) to a single hypervector.}
Suppose there are $d$-different sensors $s_{1},...,s_{d}$, each of which produce a real-valued reading $x_{1},...,x_{d}$, whereupon we may model the input at a particular moment in time by a set of tuples $\{(s_{i},x_{i})\}_{i=1}^{d}$. 
\revise{We pre-generate a set of base hypervectors to represent the values and sensors respectively.}
To represent a real valued feature $x \in \R$, we quantize the support of $x$ into a set of bins with centroids $a_{1},...,a_{Q}$, and assign each bin an embedding $\varphi(a_i)$, which we call \textit{level hypervectors}, such that $\varphi(a_{i})\cdot\varphi(a_{j})$ is monotonically decreasing in $|a_{i} - a_{j}|$. As shown in Fig.~\ref{fig:encoding} (left), we initially generate a random hypervector for the first level. To maintain similarity between adjacent level hypervectors, for each subsequent level, we randomly flip a fraction of bits from the previous level as described in \cite{thomas2021theoretical}. The fraction of flipping is denoted as $P$. This process is repeated until all $Q$ level hypervectors are generated.
To represent different sensor $s$, we assign each sensor a random embedding $\psi(s_{i})$, which we call \textit{ID hypervectors}, by sampling $\psi(s_{i})\sim \text{Unif}(\{\pm 1\}^{D})$.

The complete spatiotemporal encoding is visualized in Fig.~\ref{fig:encoding} (right). We encode a pair $(s,x)$ via $\psi(s)\otimes\varphi(a(x))$, where $a(x)$ is the centroid of the bin closest to $x$. 
This preserves both the level and sensor ID information.
To encode the readings for all sensors we bundle together their individual embeddings and round to bipolar (e.g. $\{\pm 1\}$) precision:
%\begin{equation}
$\phi(x) = \text{Sign}\left(\bigoplus_{i=1}^{d} \psi(s_{i})\otimes \varphi(a(x_{i})) \right)$.
%\end{equation}
Finally, to represent a sequence of $T$ readings: $X = \{x_{1},...,x_{T}\}$, we use permutation:
%\begin{equation}
$\phi(X) = \bigotimes_{\tau=1}^{T} \rho^{\tau}(\phi(x_{\tau}))$.
%\end{equation}

%The complete encoding process is visualized in .
\iffalse
During encoding, for a given sample $\mathbf{x} \in \mathbb{R}^{T \times d}$, we determine the level hypervector $V_{\tau, j}$ for each sensor reading $\mathbf{x}_{\tau, j}$ at time step $\tau$ from sensor $j$. These level hypervectors are then bound with the corresponding ID hypervectors $ID_{\tau,j}$.
The resulting hypervectors from all sensors $j=1,...,d$ are bundled into a single hypervector and we bipolarize ($\sigma$) it to $\pm 1$ to obtain $H_\tau$ for each $\tau$. Finally, we permute $H_\tau$ based on the order of time steps $\tau = 1, ..., T$, and bind all hypervectors into a unified hypervector $H$. The encoding process is illustrated in Fig.~\ref{fig:encoding} (right).
Mathematically, the complete procedure can be written as:
\begin{subequations}
\label{eq:encoding}
\begin{align}
    H_\tau &= \sigma \left ( \oplus_{j=1}^d \left ( ID_{\tau, j} \otimes V_{\tau, j} \right ) \right ), \\
    H &= \otimes_{\tau=1}^T \rho(H_\tau, \tau).
\end{align}
\end{subequations}
Such procedure effectively incorporates all information at various time steps, from different sensors, while accounting the proximity of real value readings.
\fi

\begin{figure}
  \includegraphics[width=0.46\textwidth]{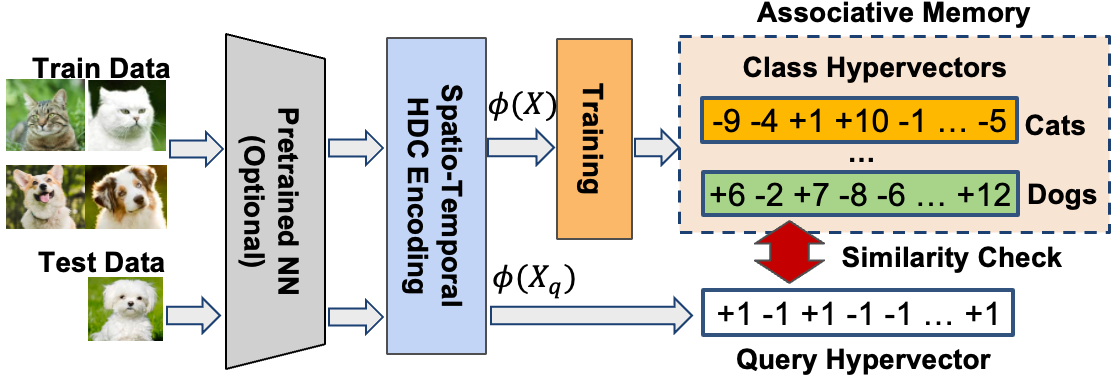}
  \vspace{-4mm}
  \caption{\small An overview of supervised HDC pipeline including encoding, training and inference (similarity check for classification). An additional pretrained NN is used as a feature extractor in the HDnn.}
  \label{fig:hd_pipeline}
  \vspace{-5mm}
\end{figure}

\textbf{HDnn Encoding.}
In this work we use the recently proposed HDnn style encoding~\cite{dutta2022hdnn,xu2023fsl} that combines a pretrained and frozen NN feature extractor with HDC's spatiotemporal encoding to obtain state of the art accuracy for sound and images.
In HDnn the inputs to the spatio-temporal encoding, $s_{1},...,s_{d}$, are intermediate feature outputs of the pretrained and frozen NN (Fig.~\ref{fig:hd_pipeline}). For example, a section of MobileNet pretrained on ImageNet creates features which are then encoded into HD hypervectors for object recognition tasks. This only marginally increases the computational costs as no training is performed on NN, all the training happens in HD.

\textbf{Supervised Training and Inference.} A common use case of HDC, summarized in Fig.~\ref{fig:hd_pipeline}, is to fit classifiers. In particular, let us suppose that we see a set of $N$ labeled samples $\{(X_{i},y_{i})\}_{i=1}^{N}$, where $x_{i}$ is an input, and $y_{i} \in \{c_{1},...,c_{J}\}$ is a class label. In the traditional approach to classification, one simply represents each class via the bundle of its training data. That is: $\phi(c_{j}) = \bigoplus_{i : y_{i} = c_{j}} \phi(X_{i})$. We store the trained class hypervectors in an associative memory. 
For example, in Fig.~\ref{fig:hd_pipeline}, we compute and store the class hypervectors of cats and dogs.
During inference, we first encode the testing sample $X_q$ into a query hypervector $\phi(X_q)$ using the same encoding procedure as for training. We then predict the label corresponding to the most similar class as measured by the cosine similarity, i.e., $\hat{y} = \argmax_{j} \cos(\phi(X_q), \phi(c_{j})) \propto \argmax_{j} (\phi(X_{q}) \cdot \phi(c_{j})) / \|\phi(c_{j})\|$.

%Due to the high dimensionality, the cosine distance between two randomly generated hypervectors should be close to zero, i.e., $\cos(\phi(x), \phi(x')) \approx 0$. Hypervectors belonging to the same class should enjoy much larger similarity and can be easily distinguished.

%For bipolar hypervectors, the cosine distance is reduced to hamming distance.

%%%%%%%%%%%%%%%%%%%%%%%%%%%%%%%%%%%%%%%%%%%%%%%%%%%%%%%
% Motivation
%%%%%%%%%%%%%%%%%%%%%%%%%%%%%%%%%%%%%%%%%%%%%%%%%%%%%%%
%\section{Motivation}
%\label{sec:motivation}

\section{Problem Definition}
\label{sec:problem}
Before diving into our method, we first rigorously formulate the unsupervised lifelong learning problem using streaming sources, driven by real-world IoT applications.

\textbf{Streaming Data.}
%\textcolor{red}{Add a plot.}
%Consider an edge device deployed in the wild where the surrounding environment can change without notice over time.
%For example, a self-driving vehicles is traveling around while the on-board sensors are continuously sampling new data.
%We denote the \textit{unlabelled} data readings by $\mathcal{D}^s = \{ \mathbf{x}_t \}, \mathbf{x}_t \in \mathbb{R}^d$, where $t$ is the sampling time stamp and $d$ is the total number of sensors.
%We envision two data sources:
%\begin{enumerate}[label={(\roman*)}]
%    \item The \textit{unlabelled} data readings collected from the on-board sensors, denoted by $\mathcal{D}^s = \{ \mathbf{x}_t \}, \mathbf{x}_t \in \mathbb{R}^d$, where $t$ is the sampling time stamp and $d$ is the total number of sensors.
%    \item The \textit{labelled} data samples transmitted from the cloud, denoted by $\mathcal{D}^c = \{(\mathbf{x}_t, y_t)\}, \mathbf{x}_t \in \mathbb{R}^d, y_t \in \mathbb{R}$. Here $t$ is the receiving time stamp. These samples might originally come from other edge devices or the Internet, and are labelled by experts thus appear to be rare.
%\end{enumerate}
%The two data sources are independent, while all data come at the same time $t$ is combined to be fed to training.
%We define $S = |\mathcal{D}^c| / |\mathcal{D}^s|$ to represent the level of supervision, which can be adjusted to match different scenarios. In this paper, we focus on the case where supervision is rare thus $S$ is small. We also allow $S=0$ for the completely unsupervised scenarios.
% and indefinitely
To represent continuously changing environment, we assume a well-known \textit{class-incremental} model in lifelong learning, in which new classes emerge in a sequential manner~\cite{rao2019continual}. \revise{We also allow data distribution shift within one class.} This setting models a scenario in which a device is continuously sampling data while the surrounding environment may change implicitly over time, e.g., the self-driving vehicle as shown in Fig.~\ref{fig:moti_intro}.
%Two real-world examples are shown in Fig.~\ref{fig:stream}.
We require that all samples appear \textit{only once} (i.e., single-pass streams).
%As presented in the introduction, class-incremental streams make the algorithm more prone to catastrophic forgetting than using iid data input.

%\revise{Need to be updated to include data distribution drift.} 
Formally, we consider a scenario involving $d$ sensors, each producing a real valued reading. We group readings into sliding windows of length $T$, and treat one such batch $X_{i} \in \R^{T \times d}$ as an input sample. Each input $X_{i}$ is associated with an unknown label $y_{i}$. Importantly, the labels are not made available during training, nor the boundaries of class shift. Therefore the entire process is unsupervised.
We represent the data stream associated with each class by $\D_{j} = \{X_{1},X_{2},...\}$, and the set of streams for all classes by $\D = \{\D_{1},...,\D_{J}\}$. Note that the class-incremental streams can have imbalanced classes, i.e., $|D_i| \neq |D_j|, i \neq j$, \revise{and gradual distribution shift within each class.}

\iffalse
Mathematically, suppose there are $J$ classes in total, we represent the data sequence from on-board sensors as $\mathcal{D} = \{ \mathcal{D}_1, ..., \mathcal{D}_J \}$, where $\mathcal{D}_j = \{X_{j}\}$ denotes a series of samples from class $j$. 
%$u$ is the global batch index.
Each sample is defined as $X_j \in \mathbb{R}^{T \times d}$ where $T$ is the length of sliding time window and $d$ is the number of sensor sources.
Note that we do not assume an equal amount of samples from each class. In other words, the class-incremental streams can have imbalanced classes, i.e., $|D_i| \neq |D_j|, i \neq j$.
The ground-truth label and the boundaries of class shift are not available to the learning algorithm, making it a completely unsupervised problem.
\fi
%For the cloud-transmitted labelled samples, $\mathcal{D}^c$ can have imbalanced classes and be sporadic due to the availability of experts.
%While we recognize that $\mathcal{D}^c$ can be imbalanced and sporadic in reality, this situation adds non-trivial complexity and we leave it for future works. 

\iffalse
\begin{figure}[h]
  \vspace{-3mm}
  \includegraphics[width=0.49\textwidth]{figs/stream2.png}
  \vspace{-6mm}
  \caption{\small Two real-world examples that produce class-incremental streams, i.e., only one class can be observed during a certain period.} % for designing continual intelligence in a complex and dynamic environment under limited resources.}
  \label{fig:stream}
  \vspace{-3mm}
\end{figure}
\fi

\textbf{Learning Protocol.}
Our goal is to build a classification algorithm that maps $\mathcal{X} \rightarrow \mathcal{Y}$.
For evaluation, we use the common evaluation protocol in state-of-the-art lifelong learning works~\cite{fini2021self,madaan2022representational,smith2019unsupervised}, 
in which we construct an iid dataset $\mathcal{E} = \{ (X_k, y_k)\}$ for periodic testing, by sampling labeled examples from each class in a manner that preserves the overall (im)balance between the classes.
Note, that even when one class has not appeared in the training data stream, it is always included in $\mathcal{E}$. Hence $\mathcal{E}$ is a global view of all classes that can potentially exist in the environment.

\revise{
\textbf{Unsupervised Clustering Accuracy.}
Since we do not give class labels or the total number of classes during training, the predicted label can be different from the ground-truth label.
Therefore, for evaluation metric, we cannot adopt the simple prediction accuracy that requires exact label matching.
Instead, we employ a widely used clustering metric known as unsupervised clustering accuracy (ACC)~\cite{xie2016unsupervised}, which mirrors the conventional accuracy evaluation but within an unsupervised context.}
%to quantify the quality of clustering regardless of the total number of clusters.

\revise{Suppose $\omega_k$ is the predicted cluster of testing sample $(X_k, y_k)$ in $\mathcal{E}$. ACC is computed as:
%\begin{equation}
$ACC = \max_m \frac{1}{|\mathcal{E}|} \sum_{k=1}^{|\mathcal{E}|}\mathbf{1} \left \{y_k = m(\omega_k)\right \}$,
%\end{equation}
where $m$ ranges over all possible one-to-one mappings between predicted clusters and ground-truth classes. Intuitively, this metric computes the accuracy under the ``best'' mapping between clusters and labels.}
\revise{The biggest advantage of ACC is that it does not require the number of clusters and classes to be equal. For instance, a cluster of pines and a cluster of redwood both belong to the ground truth label of trees. We treat such clustering result as a valid learning outcome, with a concrete visualization shown in Sec.~\ref{sec:accuracy}.
}

\begin{figure*}[t]
  \includegraphics[width=0.84\textwidth]{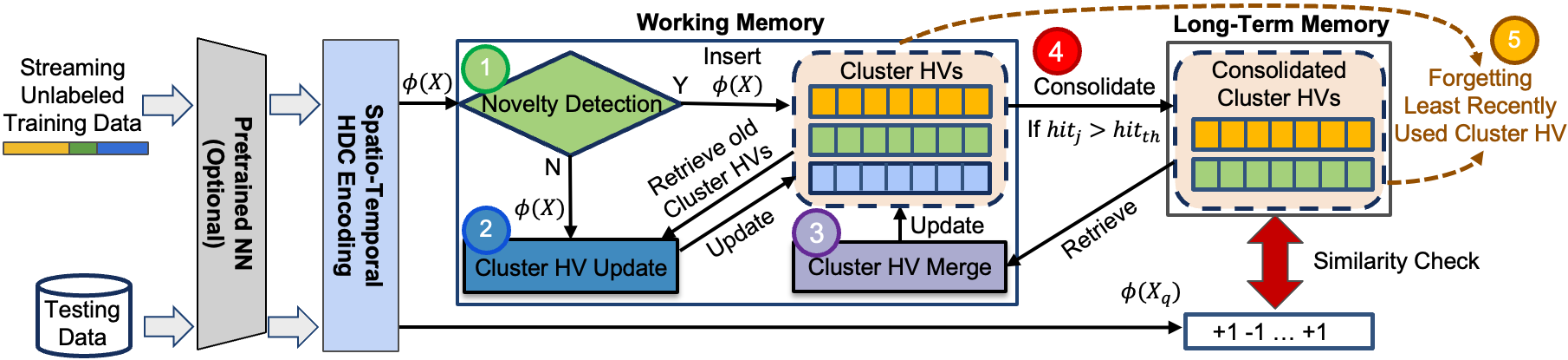}
  \vspace{-3mm}
  \caption{\small The end-to-end algorithm flow of \Method.} % with an optional NN feature extractor (for HDnn), spatio-temporal HDC encoding, and two-tier memory architecture.}} %In the working memory, {\Method} manipulate the seen prototype hypervectors with three components: novelty detection, prototype update and prototype merging.}
  \label{fig:method}
  \vspace{-3mm}
\end{figure*}
%%%%%%%%%%%%%%%%%%%%%%%%%%%%%%%%%%%%%%%%%%%%%%%%%%%%%%%
% Method
%%%%%%%%%%%%%%%%%%%%%%%%%%%%%%%%%%%%%%%%%%%%%%%%%%%%%%%
%\vspace{-2mm}
\section{{\Method}}
\label{sec:method}

%In this section, we introduce our design {\Method} thoroughly.
%which fully leverages HDC's efficiency while achieving effective learning and memorization with minimal supervision.

In this section, we present the design of {\Method}, the first unsupervised HDC framework for lifelong learning in general edge IoT applications.
Compared to operating in the original data space, HDC improves pattern separability through sparsity and high dimensionality, making it more resilient against catastrophic forgetting~\cite{shen2021algorithmic}.
{\Method} preserves the advantages of HDC in computational efficiency and lifelong learning, while handling the input of unlabeled streaming data, which has not been achieved in previous work~\cite{imani2019hdcluster,imani2019semihd,osipov2022hyperseed,hersche2022constrained,xu2023fsl}.
%\revise{
%One major novelty of \Method is its \textit{maintenance} and \textit{manipulation} of working and long term memories, which is very important to lifelong intelligence in a dynamic environment.
%For instance, it allows the system to learn to group Chihuahua and Bulldog into a more general "dog" category as it encounters more animals. Such learning surpasses the capabilities of previous memory architectures designed solely for storage~\cite{smith2019unsupervised}.}

\iffalse
For example, when a child first sees a Chihuahua and a Border Collie, she might think that they are two distinct species. As the child grows up and sees more animals, she may realize one day that both Chihuahua and Border Collie are dogs, which are different from horses. 
To achieve such a learning goal without supervision, \Method is designed with a novel two-tier associative memory architecture inspired by cognitive science studies~\cite{baddeley1992working}.
\fi

\iffalse
\begin{figure}[h]
  \vspace{-3mm}
  \includegraphics[width=0.48\textwidth]{figs/overview.png}
  \vspace{-6mm}
  \caption{\small The end-to-end design of \Method.} % with an optional NN feature extractor (for HDnn), spatio-temporal HDC encoding, and two-tier memory architecture.}} %In the working memory, {\Method} manipulate the seen prototype hypervectors with three components: novelty detection, prototype update and prototype merging.}
  \label{fig:method}
  \vspace{-5mm}
\end{figure}
\fi

%\vspace{-2mm}
\subsection{\Method Overview}
\label{sec:overview}

Fig.~\ref{fig:method} gives an overview of how \Method works.  The first step is HDC encoding of data into hypervectors as described in Sec.~\ref{sec:hdc}.   
Training samples $X$ are organized into batches of size $bSize$ and input into an optional fixed NN for feature extraction (e.g. for images and sound) and the encoding module. 
The encoded hypervectors $\phi(X)$ are input to \Method's two-tier memory design inspired by cognitive science studies~\cite{baddeley1992working}, consisting of working memory and long-term memory.
This memory system \textit{intelligently} and \textit{dynamically} manages historical patterns, stored as hypervectors and referred to as \textit{\prototypes}.
As shown in Fig.~\ref{fig:method}, the working memory is designed with three components: novelty detection, \prototype update and \prototype merge.
$\phi(X)$ is first input into novelty detection step (\textcircled{1}). An insertion to the \prototypes is made if a novelty flag is raised, otherwise $\phi(X)$ updates the existing \prototypes (\textcircled{2}).
The third component, \prototype merge (\textcircled{3}), retrieves the \prototypes from long-term memory, and merges similar \prototypes into a supercluster via a novel spectral clustering-based merging algorithm~\cite{von2007tutorial}.
The interaction between working and long-term memory happens as commonly encountered \prototypes are copied to long-term memory, which we call \textit{consolidation} (\textcircled{4}).
Finally, when the size limit of either working or long-term memory is reached, the least recently used \prototypes are forgotten (\textcircled{5}).
%All modules in {\Method} work collaboratively to achieve a good performance, as verified by experimental results in Sec.~\ref{sec:ablation-studies}.

\revise{All modules in {\Method} work collaboratively, making it adaptive and robust to continuously changing streams without relying on any form of prior knowledge. For example, in scenarios of distribution drift, \Method may generate new \prototypes upon encountering drifted samples initially, which can later be merged into coarse clusters. This approach ensures that \Method can efficiently capture and retain historical patterns.}

In the following, we discuss more details about the major components of \Method: novelty detection (Sec.~\ref{sec:novelty-detection}), \prototype update (Sec.~\ref{sec:prototype-update}), and \prototype merging (Sec.~\ref{sec:prototype-merging}). We summarize the important notations used in this paper in Table~\ref{tbl:notation}.
%For simplicity, we explain {\Method} assuming batch size $b=1$. $u$ refers to the incoming batch index thus is also the sample index.
%Suppose $\mathcal{M} = \{ m_1, m_2, ..., m_M\}$ is the set of prototypes stored in the working memory, while $\mathcal{L} = \{ l_1, l_2, ..., l_L\}$ is the prototypes stored in the long-term memory.
%The maximal allowed number of \prototypes are $M$ and $L$ in the long-term and working memory, with $M > L$.
%Each stored \prototype is a normalized hypervector.
%
%We summarize the important notations used in this paper in Table~\ref{tbl:notation}.
%The pseudocode of {\Method} is listed in Algorithm~\ref{alg:edgehd}.

\begin{table}[tb]
\small
\caption{\small List of important notations.}
\label{tbl:notation}
\vspace{-3mm}
\begin{center}
\hspace{-2mm}\begin{tabular}{p{3em} p{25em}} 
 \toprule
 \hspace{-2mm}Symbol & Meaning \\
 \midrule
 $d$ & Number of sensor sources \\
 $T$ & Time window length of one input sample $X$ \\
 %$J, K$ & Number of ground-truth and predicted classes \\
 $D$ & Dimension of the HD-space \\
 $Q$ & Number of quantized level for encoding \\
 $P$ & Fraction of random bit flip to generate level hypervector \\
 $\phi$ & HDC encoding function \\
 $\varphi, \psi$ & Level and ID hypervector encoding function \\
 $bSize$ & Batch size of input samples \\
 $\mathcal{M}, \mathcal{L}$ & Set of \prototypes stored in the working and long-term memory \\
 $M, L$ & Maximum number of \prototypes in the working and long-term memory \\
 $\mu, \hat{\sigma}$ & Mean similarity and standard difference of between each \prototype and its assigned inputs in the working memory \\
 $hit$ & The number of times that each \prototype is hit in the working memory \\
 $hit_{th}$ & The hit frequency threshold to consolidate \prototype from working to long-term memory \\
 $p, q$ & The most recent batch index when each \prototype is accessed, for the working and long-term memory \prototypes \\
 $\gamma$ & Hyperparameter for novelty detection sensitivity \\
 $\alpha$ & Moving average update rate during \prototype update \\
 $g_{ub}$ & \Prototype merge sensitivity \\
 $f_{merge}$ & Cluster HV merge frequency \\
 $r$ & Average labeling ratio in \SemiMethod \\
 $D_a$ & Dimension of the mask used in \EffMethod \\
 \bottomrule
\end{tabular}
\end{center}
\vspace{-4mm}
\end{table}

\subsection{Novelty Detection}
\label{sec:novelty-detection}
%The novelty detection and prototype update components are designed for processing new hypervectors $\phi(X)$ and deciding how to update the existing prototypes in the working memory.
%Incoming hypervector $\phi(X)$ is first assessed to determine if it should be considered as a new \prototype

The initial novelty detection step (\textcircled{1} in Fig.~\ref{fig:method}) is crucial for identifying emerging patterns in the environment.
Suppose $\mathcal{M} = \{ m_1, ..., m_M\}$ is the set of \prototypes stored in the working memory.
We gauge the "radius" of each cluster by tracking two scalars for each \prototype $i$: $\mu_i$ and $\hat{\sigma}_i$, which represent the mean cosine difference and standard difference between the \prototype and its assigned inputs. 
Given $\phi(X)$, we first identify the most similar \prototype, denoted by $j$.
\Method marks $\phi(X)$ as ``novel" if it substantially differs from its nearest \prototype. Specifically, this dissimilarity is measured by comparing $\cos(\phi(X), m_j)$ with a threshold based on the historical distance distribution of \prototype $j$:
\begin{equation}
\textrm{If } \cos(\phi(X), m_j) < \mu_j - \gamma \hat{\sigma}_j \textrm{, then flag novel.}
\end{equation}
The hyperparameter $\gamma$ fine-tunes the sensitivity to novelties. %, and is analyzed in Sec.~\ref{sec:sensitivity-analysis}.
%i.e., $\mu_j - \gamma \hat{\sigma}_j$. 

{\Method} recognizes new $\phi(X)$ as prototypes and inserts them into the working memory. 
When reaching its size limit $M$, the working memory experiences forgetting (\textcircled{5} in Fig.~\ref{fig:method}). The least recently used (LRU) \prototype, represented by $LRU = \argmin_{i=1}^{M} p_i$, is replaced. Here $p$ corresponds to the latest batch index where the \prototype was accessed. A similar forgetting mechanism is configured for the long-term memory, where the last batch accessed is marked with $q$.
%, with a forgetting mechanism when the size limit of the memory is reached.
%$p, q$ record the latest batch index when the corresponding \prototype is hit in working and long-term memory.

%, (iii) $c_i$ records the number of times that each prototype is hit by a new sample, and (iv) $p_i$ 

% line 2-6 cover the complete procedure of novelty detection and prototype update.

%compares the cosine distance between $H$ and its nearest prototype $\mathcal{M}_{w,j}$, with its the historical distance distribution of prototype $j$ defined by $\mu_j, \hat{\sigma}_j$ (line 14).
%$H$ will be flagged as novelty if its distance to the  

%\vspace{-2mm}
\subsection{\Prototype Update}
\label{sec:prototype-update}
If novelty is not detected, indicating that $\phi(X)$ closely matches \prototype $j$, we proceed to update the \prototype and its associated information (\textcircled{2} in Fig.~\ref{fig:method}). This update process involves bundling $\phi(X)$ with \prototype $m_j$, akin to how class hypervectors are updated as described in Sec.~\ref{sec:hdc}, and updaing $\mu_j$ and $\hat{\sigma}_j$ with their moving average:
\begin{subequations}
\begin{align}
    m_j &\leftarrow m_j \oplus \phi(X) \\
    \mu_j &\leftarrow (1 - \alpha) \mu_j + \alpha \cos(\phi(X), m_j) \\
    \hat{\sigma}_j &\leftarrow (1 - \alpha) \hat{\sigma}_j + \alpha |\cos(\phi(X), m_j) - \mu_j| \\
    hit_j &\leftarrow hit_j + 1, p_j \leftarrow idx
\end{align}
\end{subequations}
The hyperparameter $\alpha$ adjusts the balance between historical and recent inputs, where a higher $\alpha$ gives more weight to recent samples. Properly maintaining $\mu_j$ and $\hat{\sigma}_j$ is vital for tracking the ``radius'' of each \prototype, affecting future novelty detection. We also increase the hit frequency $hit_j$ and refresh $p_j$ with current batch index $idx$. $hit_j$ is further used to compared with a predetermined threshold $hit_{th}$ to decide when a working memory \prototype appears sufficiently frequently to be consolidated to long-term memory (\textcircled{4} in Fig.~\ref{fig:method}). $p_j$ determines forgetting as described in the previous section. With this lightweight approach, {\Method} continually records temporal \prototypes from the environment, while the most prominent \prototypes are transferred to long-term memory.

%\vspace{-2mm}
\subsection{\Prototype Merging}
\label{sec:prototype-merging}

%\revise{It is important to merge several \prototypes into a coarser group at appropriate times to improve lifelong learning performance and save memory costs.
%Intuitively, a group of \prototypes can be merged if they are highly similar to each other, and dissimilar from the other \prototypes. For instance, one might connect the \prototypes for Chihuahua and Border Collie, but not the \prototypes for Border Collie and Tabby Cat.}
%\revise{To enhance lifelong learning performance and conserve memory, it is crucial to merge similar \prototypes into coarser groups as needed. This process involves combining \prototypes that share high similarity within the group while maintaining distinctness from other \prototypes. For example, you might group Chihuahua and Border Collie \prototypes together, but keep Border Collie and Tabby Cat prototypes separate.}

\Prototype merge (\textcircled{3} in Fig.~\ref{fig:method}) has the dual benefit of reducing memory use and of elucidating underlying similarity structure in the data. Intuitively, a group of \prototypes can be merged if they are similar to each other and dissimilar from other \prototypes. For instance, one might merge the \prototypes for Bulldog and Chihuahua into a single ``Dog'' cluster HV, that remains distinct from the cluster HV for ``Tabby Cat''.

%Over time, clusters of closely related \prototypes can be consolidated into a coarser group. This has the dual benefit of reducing memory use and of elucidating underlying similarity structure in the data. Intuitively, a group of \prototypes can be merged if they are similar to each other, and dissimilar from other \prototypes. For instance, one might merge the prototypes for Border Collie and Chihuahua into a single ``dog'' cluster HV, that remains distinct from the cluster HV for ``Tabby Cat.''

\begin{figure}[t]
  \includegraphics[width=0.44\textwidth]{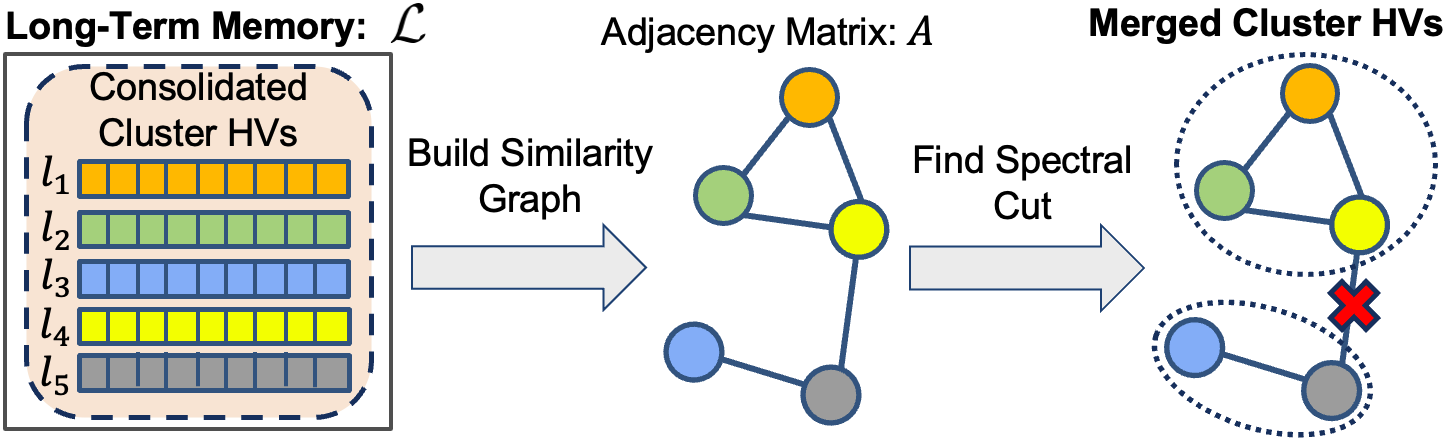}
  \vspace{-3mm}
  \caption{\small An intuitive visualization of \prototype merging.}
  \label{fig:merging}
  \vspace{-5mm}
\end{figure}

To merge the \prototypes, we first construct a similarity graph defined over the \prototypes from the long-term memory.
The \prototypes correspond to nodes, and a pair of \prototypes are connected by an edge if they are sufficiently similar. We then merge the \prototypes by computing a particular type of cut in the graph in a manner similar to spectral clustering~\cite{ng2001spectral}.
This graph based formalism for clustering is able to capture complex types of cluster geometry and often substantially outperforms simpler approaches like K-Means~\cite{von2007tutorial}. 
We detail the steps of \prototype merging in \Method below, while Fig.~\ref{fig:merging} offers an illustrative overview.

%The \prototype merge module is the key component for manipulating the memory, which consolidates several \prototypes into a coarser group at appropriate times.
%The \prototypes represent fine-grained patterns learned by the system. However, over time, it may be possible to

\smallskip
\textbf{Step 1: Preprocessing.} 
Given the set of long-term memory \prototypes $\mathcal{L} = \{ l_1, ..., l_L\}$, we construct a graph $\G$ using the adjacency matrix $A \in \{0,1\}^{L \times L}$. Here, $A_{ij} = A_{ji} = \mathbbm{1}[\cos(l_i,l_{j}) \geq \beta]$, with $\beta$ as an adaptive threshold. In other words, an edge connects \prototypes $l_{i}$ and $l_{j}$ if their similarity in HD-space surpasses $\beta$. A larger $\beta$ implies that \prototypes must be more similar to be considered for merging. In practice, we set $\beta = \frac{1}{M} \sum_{i=1}^M \mu_i$, representing the mean of the observed \prototypes.

\textbf{Step 2: Decomposition.} We compute the Laplacian $W = D - A$, where $D$ is the diagonal matrix in which $D_{ii} = \sum_{j} A_{ij}$.
We then compute the eigenvalues $\lambda_{1},..,\lambda_{L}$, sorted in increasing order, and eigenvectors $\nu_{1},...,\nu_{L}$ of $W$.

\textbf{Step 3: Grouping.}
We infer $k = \max_{i \in [L]} \lambda_{i} \leq g_{ub}$, and merge the \prototypes by running K-Means on $\nu_{1},...,\nu_{k}$. 
The upperbound $g_{ub}$ is a hyperparameter that adjusts the granularity of merging, with
a smaller $g_{ub}$ leading to smaller $k$ thus encouraging merging more aggressively.
%We conduct sensitivity analysis on $g_{ub}$ in Sec.~\ref{sec:sensitivity-analysis}.

\smallskip

Our merging approach is formally grounded, as discussed in \cite{von2007tutorial}. It is a well-known fact that the eigenvectors of $W$ encode information about the connected components of $\G$. When $\G$ has $k$ connected components, the eigenvalues $\lambda_{1} = \lambda_{2} = ... = \lambda_{k} = 0$. To recover these components, K-Means clustering on $\nu_{1},...,\nu_{k}$ can be employed, as explained in \cite{von2007tutorial}. However, practical scenarios may have a few inter-component edges that should ideally be distinct. For instance, when the similarity threshold is imprecisely set, erroneous edges may appear in the graph, causing $\lambda_{1},...,\lambda_{k}$ to be only approximately zero. Our merging approach is designed to handle this situation by introducing $g_{ub}$. % the sensitivity to which is further evaluated in experiments.
The \prototype merging is evaluated every $f_{merge}$ batches, where $f_{merge}$ is a hyperparameter that controls the trade-off between merging performance and computational latency. Both $g_{ub}$ and $f_{merge}$ are analyzed in Sec.~\ref{sec:sensitivity-analysis}, along with other key hyperparameters in \Method.
%In particular, if the graph described by $A$ contains $k$-connected components, then there exists a set of eigenvectors $\nu_{1},...,\nu_{k}$, such that $\nu_{i}^{j} > 0$ if and only if $m_{j}$ is a member of the $i$-th connected component. That is to say, $\nu_{i} > 0$ can be regarded as an indicator vector on the $i$-th connected component. The connected components themselves can then be recovered by running $k$-means on $\nu_{1},...,\nu_{k}$.

\textbf{Time Complexity of Merging.} A potentially limiting issue with spectral clustering is its time complexity, which is, in the worst case $O(L^{3})$. However, this is not a concern in our setting. First, the number of \prototypes in long-term memory ($L$) is typically small, around $50$ in practice, resulting in modest worst-case complexity. Secondly, worst-case analysis is overly pessimistic, assuming a full eigendecomposition of the graph Laplacian ($W$). In practice, $W$ is nearly always approximately low rank, meaning that only the first $k \ll L$ eigenvectors are needed. In such cases, fast randomized eigendecomposition algorithms can reduce the time complexity to linear in $L$ \cite{halko2011finding}. Thus, while spectral clustering is sometimes colloquially thought of as an ``expensive'' procedure, this is true only in very unfavorable ``worst-case'' settings. In practice, its complexity is modest and acceptable for our situation, as shown in Sec.~\ref{sec:latency}.
%\revise{While spectral clustering is known for its computational complexity, this is not a concern in {\Method}. To begin with, the merging of \prototypes takes long-term memory as input, which typically contains around 50 hypervectors. Moreover, merging is only triggered every $u_{merge}$ data batches during training. This ensures that {\Method} has sufficient time to accumulate prototypes in the long-term memory and mitigates the computational overhead. Additionally, spectral clustering exhibits computational efficiency when the similarity graph is sparse, a common characteristic of HD space. In practice, various randomized algorithms can be employed for approximate eigendecompositions~\cite{}, further reducing the time complexity to $O(M k^2)$, where $M$ represents the long-term memory size.}

%To make the \prototype work effectively and efficiently, two important considerations should be taken into account. Firstly, the \prototypes are extracted solely from the long-term memory $\mathcal{L}$, with subsequent updates to both $\mathcal{M}$ and $\mathcal{L}$. This step is crucial in preventing corruption during merging within the working memory, as discussed in more detail in Sec.~\ref{sec:ablation-studies}. 

%%%%%%%%%%%%%%%%%%%%%%%%%%%%%%%%%%%%%%%%%%%%%%%%%%%%%%%
% Variants of EdgeHD
%%%%%%%%%%%%%%%%%%%%%%%%%%%%%%%%%%%%%%%%%%%%%%%%%%%%%%%
\section{Variants of {\Method}}
\label{sec:variants}
While {\Method} is designed to cater to general IoT applications with streaming input and without supervision, real-world scenarios may vary. Some scenarios might have a few labeled samples in addition to the unlabeled stream, while others may require operation within strict power constraints. {\Method} offers extensibility to address these diverse needs.
In this section, we introduce two software-based extensions:
\SemiMethod, which adds a separate processing path to manage labeled samples, and \EffMethod, which adaptively prunes the HDC model using masking to handle low-power scenarios.

\begin{figure}[b]
  \vspace{-4mm}
  \includegraphics[width=0.43\textwidth]{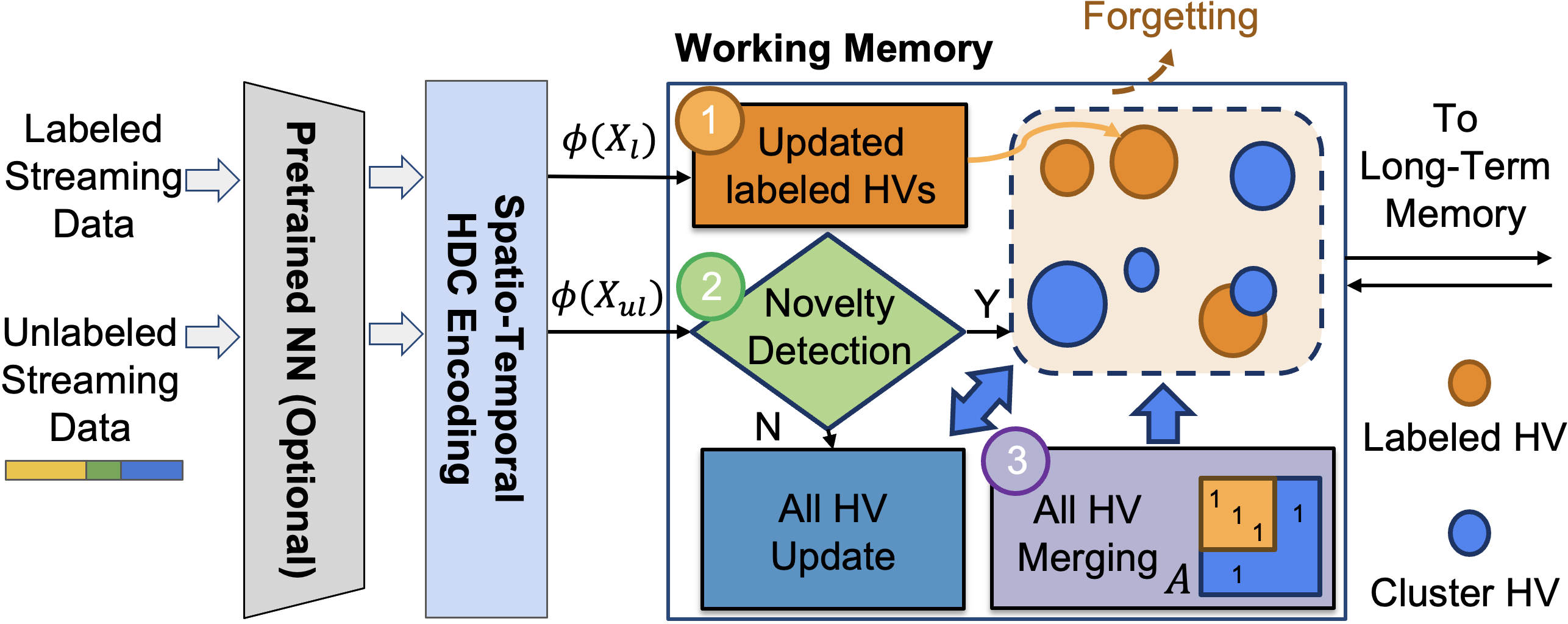}
  \vspace{-4mm}
  \caption{\small An overview of {\SemiMethod} which is designed to handle scarce labeled samples.}
  \label{fig:semihd}
  %\vspace{-4mm}
\end{figure}

\vspace{-2mm}
\subsection{\SemiMethod}
\label{sec:semiedgehd}

While {\Method} excels in unsupervised scenarios, it does not harness labeled data when available. To address this limitation, we introduce {\SemiMethod} as an extension to enhance accuracy utilizing the limited labels. In Fig.~\ref{fig:semihd}, we provide an overview of {\SemiMethod}. For each input batch $idx$, we consider two subsets: one labeled $(X_{l,idx}, y_{l,idx})$ and one unlabeled $X_{ul,idx}$. We denote the average labeling ratio throughout the data stream as $r = \frac{\sum_{idx} |X_{l,idx}|}{\sum_{idx} |X_{l,idx}| + |X_{ul,idx}|}$. Since obtaining external supervision is often challenging in dynamic environments, we focus on cases where $r \leq 0.01$.
%We extend {\Method} to {\SemiMethod} to handle the scarce labeled samples along with the unlabeled stream. 

{\SemiMethod} retains the two-tier memory structure of {\Method} but introduces modifications to the working memory components. In the {\SemiMethod} pipeline, the working memory undergoes three key steps. Firstly, labeled samples $(X_{l}, y_{l})$ update labeled class hypervectors following the conventional HDC methods outlined in Sec.~\ref{sec:hdc}. Next, we process unlabeled samples $X_{ul}$ through novelty detection and HV update modules, mirroring {\Method}. Importantly, in {\SemiMethod}, these operations are applied to both labeled HVs and \prototypes. Lastly, we introduce a merging step to group labeled HVs and \prototypes that are closely related. To handle labeled HVs, we modify the adjacency matrix $A$ by making it diagonal for labeled entries. For example, if the first $J$ HVs correspond to labeled HVs, we ensure that $A_{1:J, 1:J} = \text{diag}([1,...,1])$, while calculating the remaining values following {\Method} procedures. This strategy prevents the merging of labeled HVs with each other.
With these adjustments, {\SemiMethod} offers a solution that retains the core elements of {\Method} while handling scarce labeled inputs.

%\vspace{-2mm}
\subsection{\EffMethod}
\label{sec:effedgehd}
\begin{figure}[t]
  \centering
  %\vspace{-2mm}
  \begin{subfigure}[t]{0.2\textwidth}
     \centering
     \includegraphics[width=\textwidth,height=2cm]{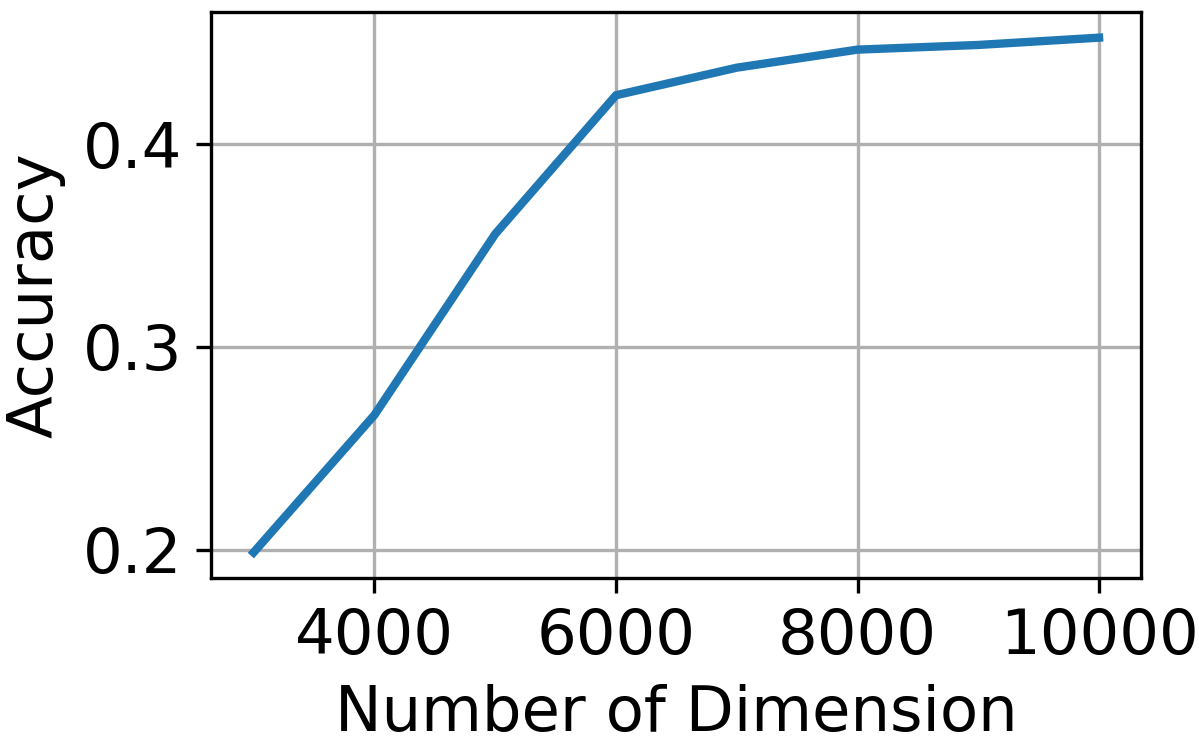}
     %\caption{\small An example to show the impact of masking on HDC's supervised performance.}
     %\label{fig:behnam}
 \end{subfigure}
 %\hfill
 \begin{subfigure}[t]{0.2\textwidth}
     \centering
     \includegraphics[width=\textwidth,height=2cm]{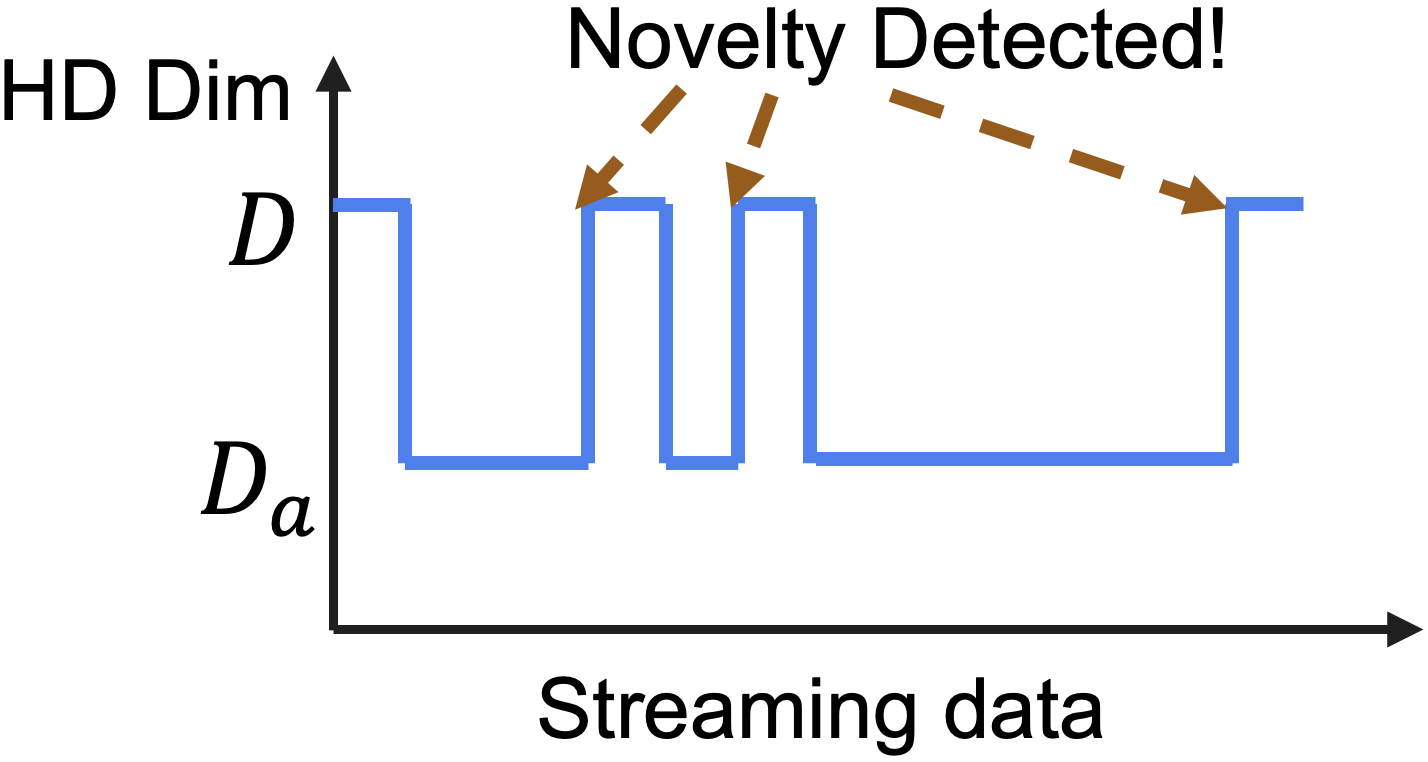}
     %\caption{\small How {\EffMethod} adaptively adjusts the HD dimension.}
     %\label{fig:eff_intuition}
 \end{subfigure}
 %\hfill
 \vspace{-4mm}
  \caption{\small Left: An example of the impact of masking on HDC. Test on CIFAR-10~\cite{krizhevsky2009learning}. Right: The intuitions behind the design of {\EffMethod}.}
  \vspace{-5mm}
  \label{fig:effhd}
\end{figure}

While HDC computation is typically lightweight, there may be instances of energy scarcity (e.g., when powered by a solar panel) that call for a balance between accuracy and power efficiency. Similar to neural networks, one approach is to prune the HDC model using a mask, retaining the most crucial HDC dimensions post-encoding~\cite{khaleghi2020prive}. Dimension importance can be determined by aggregating all class hypervectors into one and sorting the values across all dimensions. Notably, direct reduction of the encoding dimension should be avoided, as it can degrade HDC's expressive capability and end up with corruption. Fig.~\ref{fig:effhd} (left) visually demonstrates the impact of masking in supervised HDC tasks: retaining the top 6000 bits incurs only a 3\% accuracy loss compared to using the full 10K-bit precision.

%Previous work~\cite{khaleghi2020prive} has pointed out that HDC model pruning can be done by retaining the most significant HDC dimensions after encoding, while discarding out the rest.
%a similar pruning approach can be applied to HDC by simply masking the hypervectors.
%Notably, effective pruning can be done by retaining the most significant HDC dimensions after encoding, while discarding out the rest.
%For neural networks, previous works have conducted multiple attempts on model pruning and quantization to obtain a lighter model~\cite{cai2019once,han2019deep,lin2020mcunet,lin2021memory,saha2023tinyns}.

While the concept of masking has been employed in prior HDC studies~\cite{khaleghi2020prive}, they are not directly applicable to {\Method} due to their offline training setting with iid data.
With streaming non-iid data in {\Method}, the set of observed \prototypes may only represent a subset of the potential classes, and the less significant bits could become crucial as new classes are introduced.
%These works compute a fixed mask from the class hypervectors after one epoch of data. However, with streaming non-iid data in {\Method}, the set of observed \prototypes may only represent a subset of the potential classes, and the less significant bits could become crucial as new classes are introduced.

We introduce {\EffMethod}, which enhances {\Method} through an adaptive masking approach applied to all \prototypes in working and long-term memory. Let $D_a$ represent the target dimension for reduction. The rationale behind {\EffMethod} is depicted in Fig.~\ref{fig:effhd} (right). Whenever an original {\Method} detects novelty, we temporarily revert to the full dimension $D$ for 2 batches, which is sufficient for \Method to consolidate new patterns in its memory. After these two batches, we assess the long-term memory \prototypes by aggregating them and ranking the dimensions, and derive a mask retaining $D_a$ dimensions with the largest absolute values. This mask is then applied to the following batches of $\phi(X)$ immediately after encoding, up until the next novelty is detected. 
Novelty detection is executed with the masked hypervectors. Importantly, {\EffMethod} can utilize the same novelty detection sensitivity as {\Method} since the most significant dimensions dominate the similarity check. In other words, the similarity results in \EffMethod using $D_a$ dimension are similar as using the full dimension.
{\EffMethod} offers an adaptive HDC model pruning interface with minimal accuracy loss and overhead.

\begin{table*}[t]
%\vspace{-4mm}
\small
\caption{\small Experimental setup of \Method across all datasets.}
\label{tbl:datasets}
\vspace{-4mm}
\begin{center}
\begin{tabular}{c|cccc|ccc} % note: no vertical bars at all
\toprule % not \hline
\small
\textbf{Dataset} & \textbf{Application} & \textbf{Classes} & \textbf{Total} & \textbf{Training Data Order} & \textbf{HDnn?} & \textbf{Pretrained} &\textbf{\# of} \\
& \textbf{Category} & \textbf{(Balanced?)} & \textbf{Samples} & & & \textbf{Models} & \textbf{ Params} \\ 
\midrule % not \hline
MHEALTH~\cite{misc_mhealth_dataset_319} & Activity & 12 (N) & 9K & Temporal order during collection & N & - & - \\
ESC-50~\cite{piczak2015esc} & Sound & 50 (Y) & 2K & Class-incremental, random within class & Y &ACDNet~\cite{mohaimenuzzaman2023environmental} & 4.7M \\
%CIFAR-10~\cite{krizhevsky2009learning} & Image & 10 (Y) & 50K & Class-incremental, random within class & ResNet-18 & 11.2M & Y \\ 
\multirow{2}{*}{CIFAR-100~\cite{krizhevsky2009learning}} & \multirow{2}{*}{Image} & \multirow{2}{*}{20 (Y)} & \multirow{2}{*}{60K} & Class-incremental, random within class & \multirow{2}{*}{Y} & \revise{MobileNet V2~\cite{sandler2018mobilenetv2} or} & \revise{2.2M} \\
& & & & \revise{or gradual rotation within class} & & \revise{MobileNet V3 small~\cite{howard2019searching}} & \revise{927K} \\
\bottomrule % not \hline
\end{tabular}
\end{center}
\vspace{-2mm}
\end{table*}
% mobilenet v3~\cite{howard2019searching}, params 927k

%%%%%%%%%%%%%%%%%%%%%%%%%%%%%%%%%%%%%%%%%%%%%%%%%%%%%%%
% Evaluation
%%%%%%%%%%%%%%%%%%%%%%%%%%%%%%%%%%%%%%%%%%%%%%%%%%%%%%%
%\vspace{-2mm}
\section{Evaluation of {\Method}}
\label{sec:evaluation-main}

\subsection{System Implementation} 
We implement {\Method} with Python and PyTorch~\cite{paszke2019pytorch} and deploy it on three standard edge platforms: Raspberry Pi (RPi) Zero 2 W~\cite{rpi0}, Raspberry Pi 4B~\cite{rpi4b}, and Jetson TX2 module~\cite{jetsontx2}. The selection of edge platforms represent three tiers with small, medium and abundant resources.
%Our experiments reflect the typical performance in practical edge deployments.

RPi Zero 2 W has a 1GHz quad-core Cortex-A53 CPU and 512MB SDRAM.
RPi 4B enjoys a 1.8GHz quad-core Cortex-A72 CPU and 4GB SDRAM. The Jetson TX platform is equipped with a dual-core NVIDIA Denver 2 CPU, a
quad-core ARM Cortex-A57 MPCore, an NVIDIA Pasca family GPU with 256 NVIDIA CUDA cores, and 8GB RAM. We measure the training latency per batch and the energy consumption using the Hioki 3334 powermeter~\cite{powermeter}. 
%The complete setup is shown in Fig.~\ref{fig:setup}.

\iffalse
\begin{figure}[t]
  \includegraphics[width=0.4\textwidth]{figs/setup.png}
  \vspace{-4mm}
  \caption{\small System implementation and setup.}
  \label{fig:setup}
  \vspace{-6mm}
\end{figure}
\fi

We are aware that all NN-based feature extractors can be pruned and quantized to attain more efficient deployment on edge platforms~\cite{dutta2022hdnn}, same for the NN-based baselines we compare to~\cite{madaan2022representational,fini2021self}.
However, NN model compression is not the primary focus of {\Method}. Existing compression techniques~\cite{neill2020overview,kim2019efficient} can be applied directly to the feature extractor in {\Method}.
We leave {\Method} with acceleration design and emerging hardware deployment for future works.

\vspace{-2mm}
\subsection{Experimental Setup}
We conduct comprehensive experiments to evaluate {\Method} on three typical edge scenarios.
All three scenarios incorporate continuous data streams and expect lifelong learning over time.
We summarize the experimental setup in Table~\ref{tbl:datasets}.

\textbf{Application \#1: Personal Health Monitoring.}
%With the prevalence of personal healthcare devices, 
Continuous health monitoring has emerged as a popular use case for IoT. 
%It is essential to perform on-device learning and classification to ensure users' privacy. 
%To simulate a personal healthcare equipment, such as a smartwatch, 
We utilize the MHEALTH~\cite{misc_mhealth_dataset_319} dataset which includes measurements of acceleration, rate of turn, and magnetic field orientation on a smartwatch. MHEALTH differentiates 12 activities in daily lives and is collected from 10 subjects. Notably, MHEALTH employs raw time-series signals rather than processed frequency components as inputs. We use time windows of 2.56s ($T=128$) with 75{\%} overlap to generate the samples. In contrast to previous datasets, we strictly adhere to the temporal order during data collection. %, ensuring that the constructed data streams reflect the gradual transition of activities. This is similar to the class-incremental scenario but with possibly recurrent class appearance.

\textbf{Application \#2: Sound Characterization.}
Continuous sound detection contributes to the characterization of urban environments.
%as well as real-time noise monitoring and response.
%Using microphones on edge devices, sound classification acts as a common IoT application. 
We choose the ESC-50~\cite{piczak2015esc} dataset to emulate this scenario. 
%, a dataset specifically designed for environmental sound classification. 
This dataset comprises 5-second-long recordings categorized into 50 semantically diverse classes, including animals, human sounds, and urban noises. We construct the class-incremental streams by arranging the data in random order within each class.

\textbf{Application \#3: Object Recognition.} 
Object recognition is a common use case for camera-mounted mobile systems, e.g., self-driving vehicles.
%The on-board algorithm is expected to learn continuously from the streaming images.
%We select CIFAR-10~\cite{krizhevsky2009learning} and Stream-51~\cite{Roady_2020_Stream51}, featuring 32$\times$32 RGB images of 10 and 51 classes respectively, to assess this scenario. For CIFAR-10, we randomly sample instances within each class to construct the class-incremental streams. In the case of Stream-51, we maintain the temporal order of the videos within each class, making it a more challenging setting compared to CIFAR-10 and closely resembling human perception.
%We select Stream-51~\cite{Roady_2020_Stream51}, a video-based dataset consisting of 32$\times$32 RGB images of 51 classes. We strictly maintain the temporal order of the videos within each class, making it closely resemble human perception.
We set up a class-incremental stream from CIFAR-100~\cite{krizhevsky2009learning}, consisting of 32$\times$32 RGB images of 20 coarse classes.
\revise{We further evaluate the case of data distribution drift by examining gradual rotations occurring within each CIFAR-100 class.}

%For all settings, we apply the spatiotemporal encoding in {\Method} after the frozen feature extractor, using a time window length of $T=1$.
On MHEALTH, {\Method} is fully dependent on the HDC spatiotemporal encoder to process the raw time-series signals.
For ESC-50 and CIFAR-100, {\Method} utilizes the HDnn framework with a pretrained feature extractor before HDC encoding, same as in the state-of-the-art HDC works~\cite{shen2021algorithmic,hersche2022constrained}. Specifically, we adapt a pretrained ACDNet with quantified weights~\cite{mohaimenuzzaman2023environmental} for ESC-50. ACDNet is a compact convolutional neural network architecture designed for small embedded devices.
\revise{For CIFAR-100, we use a MobileNet V2~\cite{sandler2018mobilenetv2} for accuracy evaluation and MobileNet V3 small~\cite{howard2019searching} for efficiency evaluation, both pretrained on ImageNet~\cite{russakovsky2015imagenet}.}
For all pretrained NNs, we remove the last fully connected layer used for classification and keep the remaining weights frozen.

%For the MHEALTH dataset, we directly employ the spatiotemporal HDC encoder on the raw time-series samples. This setup fully utilizes the expressiveness of the HDC encoder for time-series IoT data without relying on recurrent neural network-based feature extractors.
Table~\ref{tbl:parameters} summarizes the key hyperparameters in {\Method}, which are selected based on a separate validation set.
We configure $\alpha=0.1$ for moving-average update, $hit_{th}=10$ for long-term memory consolidation.
The long-term memory size $L$ is set to 50 in all cases.

\begin{table}[t]
\small
%\vspace{-4mm}
\caption{\small Important hyperparameters configuration of {\Method}.}
\label{tbl:parameters}
\vspace{-4mm}
\begin{center}
\begin{tabular}{c|ccc|ccccc} % note: no vertical bars at all
 \toprule % not \hline
 \small
 \textbf{Dataset} & \multicolumn{3}{c|}{\textbf{HDC Encoding}} & \multicolumn{5}{c}{\textbf{\Method Design}} \\
 & $D$ & $Q$ & $P$ & $bSize$ & $M$ & $\gamma$ & $g_{ub}$ & $f_{merge}$\\ 
 \midrule % not \hline
% CIFAR-10 & 10000 & 100 & 0.01 & 32 & 100 & 3.0 & 0.1 \\
 MHEALTH & 1000 & 5 & 0.01 & 32 & 50 & 3.0 & 0.2 & 25 \\
 ESC-50 & 10000 & 100 & 0.02 & 32 & 100 & 1.0 & 0.1 & 5 \\
 CIFAR-100 & 10000 & 100 & 0.01 & 32 & 100 & 1.0 & 0.1 & 150 \\
 \bottomrule % not \hline
\end{tabular}
\end{center}
\vspace{-3mm}
\end{table}

%\textbf{Evaluation Metrics.} We use NMI and RI as the major metric for evaluating the performance of unsupervised lifelong learning. 
%The higher the NMI/RI, the better the quality of unsupervised clustering.
%Additionally, on the edge platforms, 

%\vspace{-2mm}
\subsection{State-of-the-Art Baselines}
We conduct a comprehensive comparison between {\Method} and state-of-the-art NN-based unsupervised lifelong learning baselines, which continuously train a NN for representation learning. The loss functions in these setups are defined in the feature space without relying on label supervision. %For instance, CaSSLe~\cite{fini2021self} utilizes distillation of feature representations from a previous network to the current network during training. 
During testing, we freeze the neural network and apply K-Means clustering on the testing feature embeddings to generate predicted labels. $k$ is set to 50 which is the same number of \prototypes as in \Method. Such a pipeline is widely used for lifelong learning evaluations~\cite{rao2019continual,smith2019unsupervised}.
%For the image- and audio-based applications,

Fig.~\ref{fig:exp-pipeline} presents a comparison of the pipeline setup using both the baselines and {\Method} on HDnn and non-HDnn frameworks respectively.
%It is important to note that HDC is a new computing paradigm that operates without gradient descent. 
%Consequently, the setup of {\Method} differs from existing solutions in unsupervised lifelong learning. 
%In particular, {\Method} incorporates an optional frozen feature extractor and trains the \prototypes in the memory, while the baselines train the NN feature extractors for better representations in the feature space.
To ensure fair comparisons, in HDnn framework on ESC-50 and CIFAR-100, we initialize the NN with the same pretrained weights for {\Method} and NN baselines. 
\revise{For the NN baselines on MHEALTH, we randomly initialize a one-layer LSTM of 64 units followed by a fully connected layer of 512 units. This architecture has achieved competitive accuracy as the Transformers-based designs on MHEALTH~\cite{essa2023temporal}.}
% with around 83K randomly initialized weights.
%We have verified the expressibility of the model - the LSTM converges to about 80\% accuracy after 20 epochs of supervised training.
%which is a commonly used architecture for human activity recognition with time-series data~\cite{chen2016lstm}. The LSTM is designed to have compatible complexity with our HDC encoding, and we have verified the expressibility of the model - the LSTM converges to about 80\% accuracy after 20 epochs of supervised training. The weights of LSTM are randomly initialized since {\Method} does not rely on any NN-based feature extractor.

We compare \Method with the following baselines, which include all main lifelong learning techniques:

\begin{figure}[t]
  \includegraphics[width=0.45\textwidth]{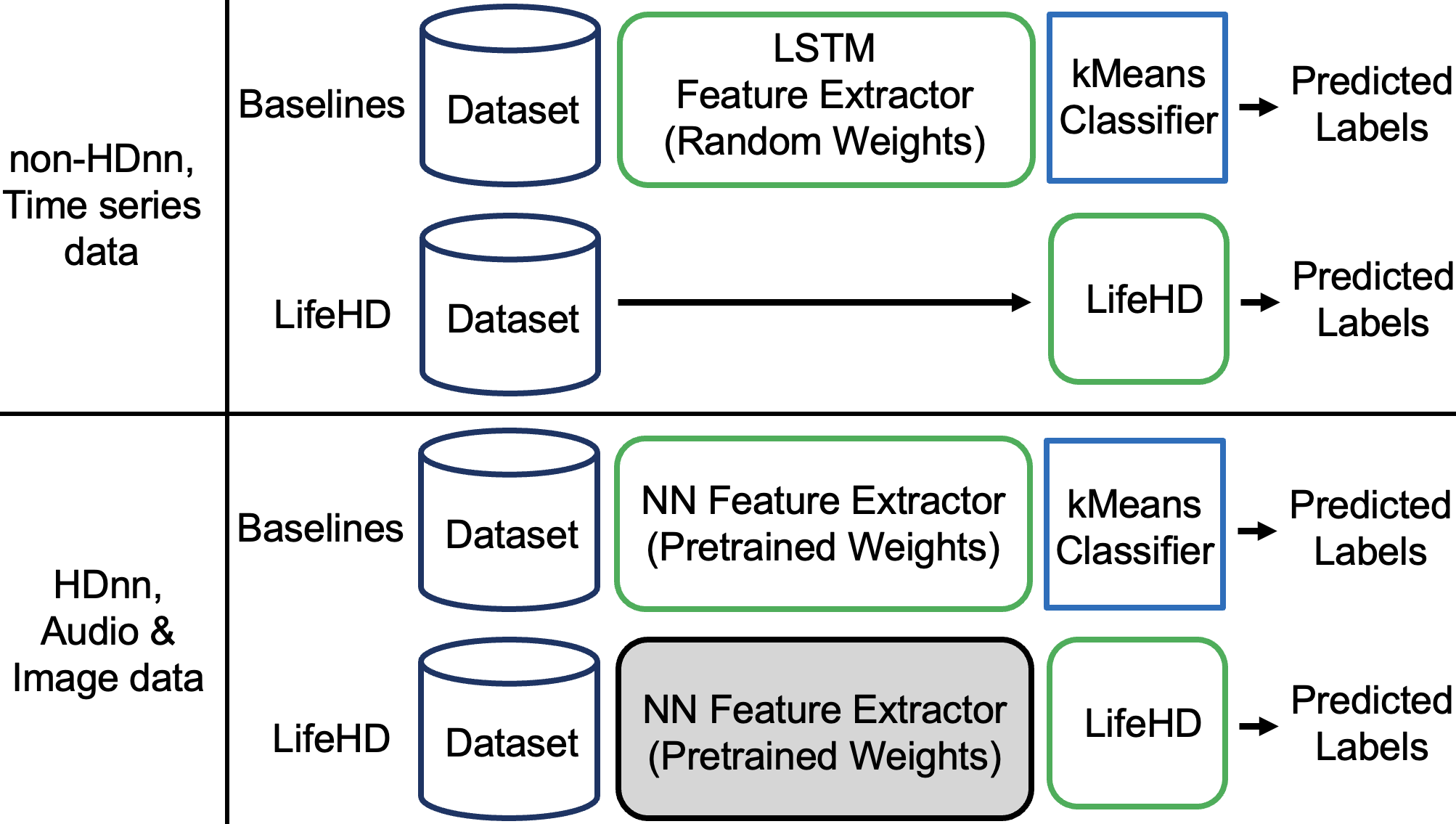}
  \vspace{-3mm}
  \caption{\small A graphical explanation of the pipeline setup. Green outlines denote the module trained with the streaming data. Blue outlines denote the unsupervised classifier trained during testing. Gray denotes when the module is frozen \& not trained further.}
  \label{fig:exp-pipeline}
  \vspace{-3mm}
\end{figure}

\begin{figure*}[t]
\begin{center}
\begin{tabular}{c c}
    \vspace{-1mm}
    \includegraphics[width=0.6\textwidth, height=3.2cm]{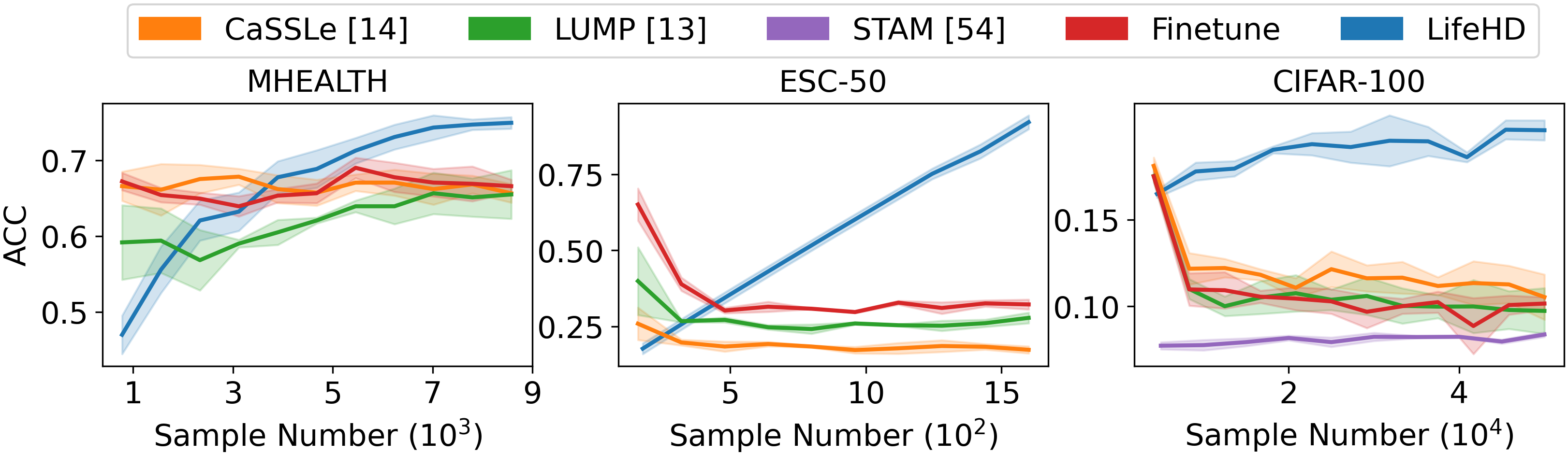} &
    \includegraphics[width=0.37\textwidth, height=2.5cm]{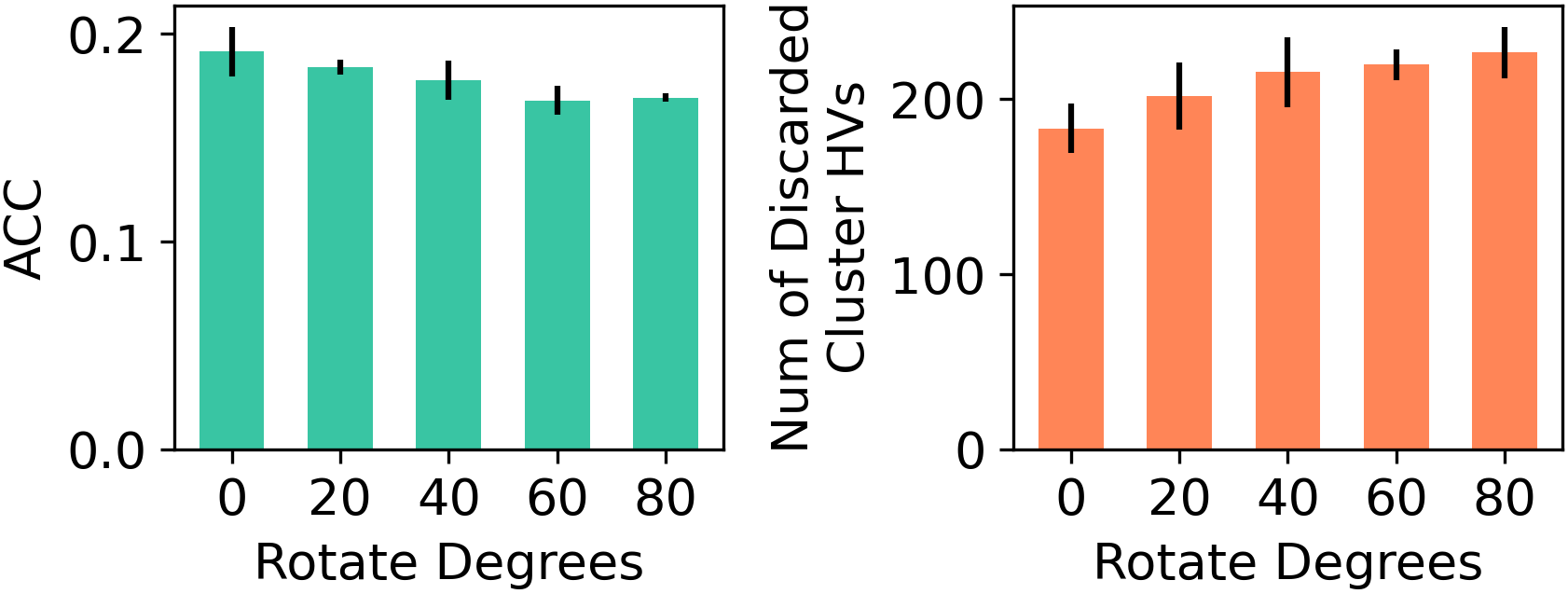} \\
    (a) {\small The ACC curve of \Method vs. state-of-the-art NN baselines on three scenarios.} & (b) {\small The final results of \Method in CIFAR-100 with rotation.} \\
\end{tabular}
\vspace{-4mm}
\caption{\small The unsupervised clustering accuracy (ACC) results of \Method on various input data streams.}
\label{fig:acc}
\end{center}
\vspace{-3mm}
\end{figure*}

%we employ the state-of-the-art unsupervised lifelong learning methods which are representative, including:

\begin{itemize}
%\item \textbf{PNN}~\cite{rusu2016progressive}: Progressive Neural Network gradually expands the network architecture to learn new representations.
%\item \textbf{SI}~\cite{zenke2017continual} performs online per-synapse consolidation as a typical regularization technique. 
    \item \textbf{Finetune} is a na\"ive baseline that optimizes the NN model using the current batch of data without any lifelong learning techniques.
    \item \textbf{CaSSLe}~\cite{fini2021self} is a distillation-based framework that utilizes self-supervised losses. It leverages distillation between the representations of the current model and a past model. In the original paper, the past model is captured at the end of the previous task and prior to the introduction of a new task. However, since we do not assume awareness of task shifts, we simply freeze the model from the previous batch.
    \item \textbf{LUMP}~\cite{madaan2022representational} employs a memory buffer for replay and mitigates catastrophic forgetting by interpolating the current batch with previously stored samples in the memory.
    %, with add-ons to distill the best possible representations invariant to the underlying shifts. %CaSSLe represents the knowledge distillation-based approaches.
    \item \textbf{STAM}~\cite{smith2019unsupervised} is brain-inspired expandable memory architecture using online clustering and novelty detection. We exclusively apply STAM to CIFAR-100 due to its demand for intricate dataset-specific tuning (e.g., number of receptive fields), and because the authors only released the implementation for the CIFAR datasets. %and memory update. 
    \item \textbf{SupHDC}~\cite{kim2018efficient,hersche2022constrained} is the fully supervised HDC pipeline.
\end{itemize}

%\textbf{CaSSLe}~\cite{fini2021self} and \textbf{LUMP}~\cite{madaan2022representational}, as introduced in Sec.~\ref{sec:motivation}. Additionally, we compare with \textbf{STAM}~\cite{smith2019unsupervised}, a recently proposed brain-inspired expandable memory architecture. 
All baselines are adapted from their original open-source code.
For CaSSLe and LUMP, we employ BYOL~\cite{grill2020bootstrap} as the self-supervised loss function because it has showed superb empirical performance in lifelong learning tasks compared to other self-supervised learning backbones ~\cite{fini2021self}. %~\cite{,zbontar2021barlow,bardes2021vicreg}.
%Unlike other contrastive learning methods~\cite{chen2020simple,chen2021exploring}, BYOL achieves state-of-the-art performance without using any negative samples.
We use the memory buffer size of 256 for LUMP which is the same as in the original paper.
We employ the Stochastic Gradient Descent (SGD) optimizer with a learning rate of 0.03 across all methods, training each batch for 10 steps.
All experiments are executed for 3 random trials. 
\subsection{\Method Accuracy}
\label{sec:accuracy}

%We present the results of unsupervised clustering accuracy (ACC) in Fig.~\ref{fig:acc}.
%All experiments are executed for 3 random trials.  
%It is also worth noting that problem's scale (e.g., the image size) is limited due to the essential difficulty of unsupervised lifelong learning, including single-pass non-iid data and no supervision. For the same reason, there remains a disparity in accuracy between unsupervised lifelong learning and fully supervised NNs, as substantiated by prior research~\cite{smith2019unsupervised,madaan2022representational}.
%We next review the results of time series data, followed by images and audio, as they exhibit different patterns.

%\textbf{Results on Time Series Data.}
\revise{\textbf{Results on Three Application Scenarios.}}
%When learning from time series data, such as MHEALTH, initial knowledge across all methods is limited, leading to lower ACCs at the beginning as shown in the leftmost plot in Fig.~\ref{fig:acc}.
Fig.~\ref{fig:acc} (a) details the ACC curve of all methods as streaming samples are received.
All NN baselines start at higher accuracy, especially in ESC-50 (sounds) and CIFAR-100 (images), owing to the presence of a pretrained NN feature extractor within the HDnn framework.
Meanwhile, \Method begins with lower accuracy as both the working and long-term memories are empty, needing to learn the \prototypes and the optimal number of clusters. 
Notably, as streaming samples come in, all NN baselines experience a decline in ACC, underscoring the inherent challenges of unsupervised lifelong learning with streaming non-iid data and a lack of supervision.
This is primarily due to the demand for extensive iid data and multi-epoch offline training for finetuning NNs,  which is not feasible in our setting. 
CaSSLe~\cite{fini2021self} leads to forgetting due to its inability to identify suitable past models from which to distill knowledge.
Similarly, LUMP~\cite{madaan2022representational} exhibits reduced ACC in ESC-50 and CIFAR-100, with only marginal ACC improvement in MHEALTH (time series), suggesting that its memory interpolation strategy may not be universally suitable for all applications.
While the memory-based design of STAM~\cite{smith2019unsupervised} can mitigate forgetting, its efficacy in distinguishing patterns and acquiring new knowledge remains unsatisfactory.
On the contrary, \Method demonstrates incremental accuracy across all three different scenarios, achieving \textbf{up to 9.4\%, {\AccRes} and 11.8\% accuracy increase} on MHEALTH (time series), ESC-50 (sound) and CIFAR-100 (images), compared with the NN-based unsupervised lifelong learning baselines at the end.
Such outcome can be attributed to HDC's lightweight but meaningful encoding and the effective memorization design of \Method.

\begin{table}[t]
\small
%\vspace{-2mm}
\caption{\small \revise{The gap of ACCs at the end of the stream between {\Method} and Supervised HDC~\cite{kim2018efficient,hersche2022constrained}}.}
\label{tbl:suphd}
\vspace{-4mm}
\begin{center}
\begin{tabular}{cccc} % note: no vertical bars at all
 \toprule % not \hline
 \textbf{Method} & \textbf{MHEALTH} & \textbf{ESC-50} &  \textbf{CIFAR-100} \\ % \textbf{CIFAR-10} &
 \midrule
 \textbf{\Method} & 0.75 & 0.92 & 0.20 \\ % 0.850 &
 \textbf{Supervised HDC}~\cite{kim2018efficient,hersche2022constrained} & 0.90 & 0.95 & 0.26 \\ \hline % 0.861 & 
 Gap & -0.15 & -0.03 & -0.06 \\ % +0.011 &
 \bottomrule % not \hline
\end{tabular}
\end{center}
\vspace{-3mm}
\end{table}

\begin{figure}[t]
\begin{center}
\includegraphics[width=0.46\textwidth]{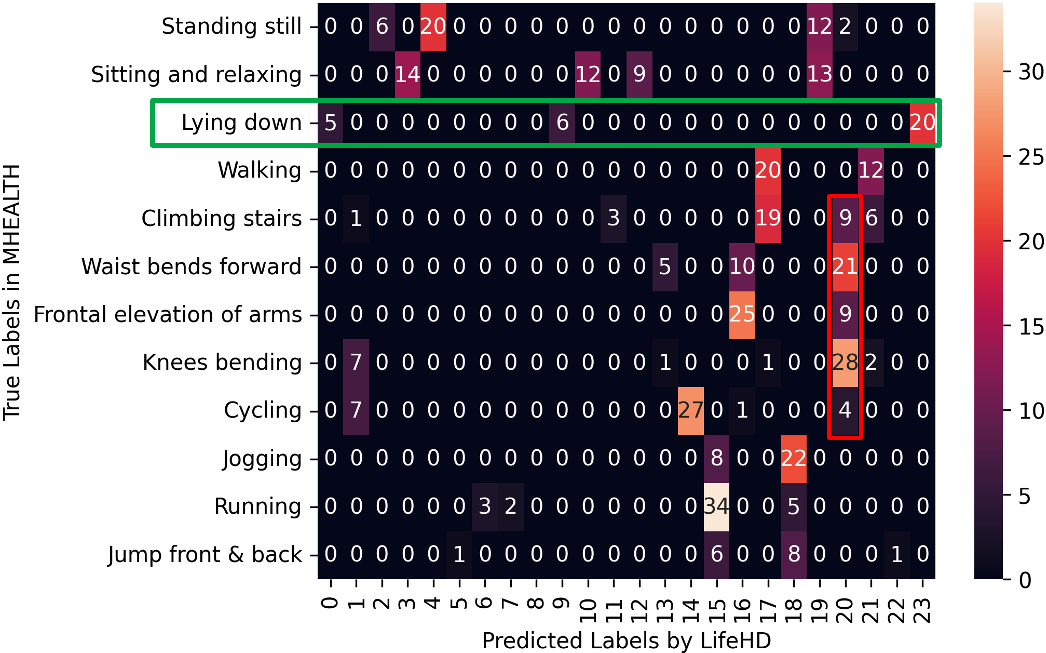}
\vspace{-4mm}
\caption{\small The confusion matrix of \Method on MHEALTH. The green box highlights smaller \prototypes that form a single large ground truth class, which is a valid learning outcome. The red box highlights "boundary" \prototypes that span multiple true labels, leading to lower ACCs.}
\label{fig:cm}
\end{center}
\vspace{-5mm}
\end{figure}

\textbf{Results under Data Distribution Drift.}
\revise{We further evaluate \Method's performance under drifted data and present the final ACC along with the number of discarded \prototypes in Fig.~\ref{fig:acc} (b).
Specifically, we introduce gradual rotation to the CIFAR-100 samples within each class, ranging from no rotation to a substantial rotation angle of $80^{\circ}$.
The other parameter settings remain the same as in Table~\ref{tbl:parameters}.
The number of discarded \prototypes accounts for those that are either forgotten or merged.
From Fig.~\ref{fig:acc} (b), we can observe the remarkable resilience of \Method to drifted data, with an ACC loss of less than 2.3\% even under a severe rotation of $80^{\circ}$. This robustness stems from the general and uniform design of \Method to accommodate various types of continuously changing data streams. In cases of slight or minimal distribution drift, \Method updates existing \prototypes; in instances of severe drift, new \prototypes are created and subsequently merged if deemed appropriate. 
However, due to the finite memory capacities, more \prototypes are subject to forgetting or merging under larger drifts, as shown in Fig.~\ref{fig:acc} (b) by the number of discarded \prototypes, leading to ACC loss. In our experiments, \Method demonstrates minimal ACC loss even under substantial rotation shifts.}

\begin{figure*}[t]
\begin{center}
\begin{tabular}{c|c|c}
    \multicolumn{3}{c}{\includegraphics[width=0.65\textwidth]{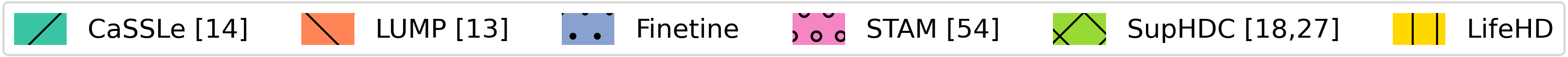}} \vspace{-1mm} \\
    \vspace{-1mm}
    \includegraphics[width=0.12\textwidth, height=1.8cm]{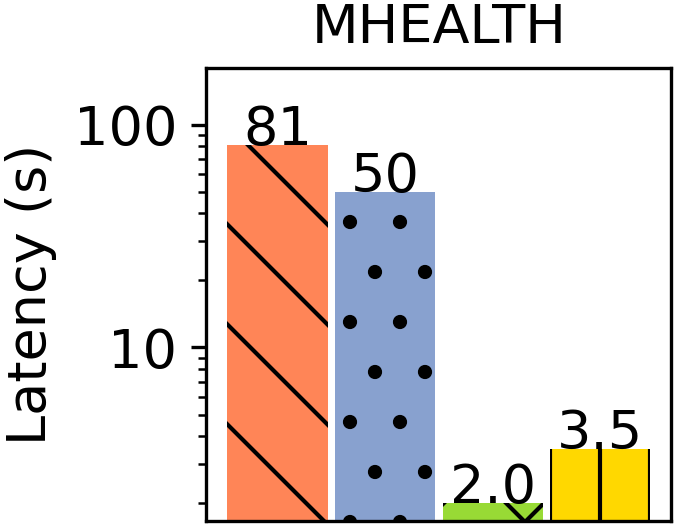} &
    \includegraphics[width=0.4\textwidth, height=1.8cm]{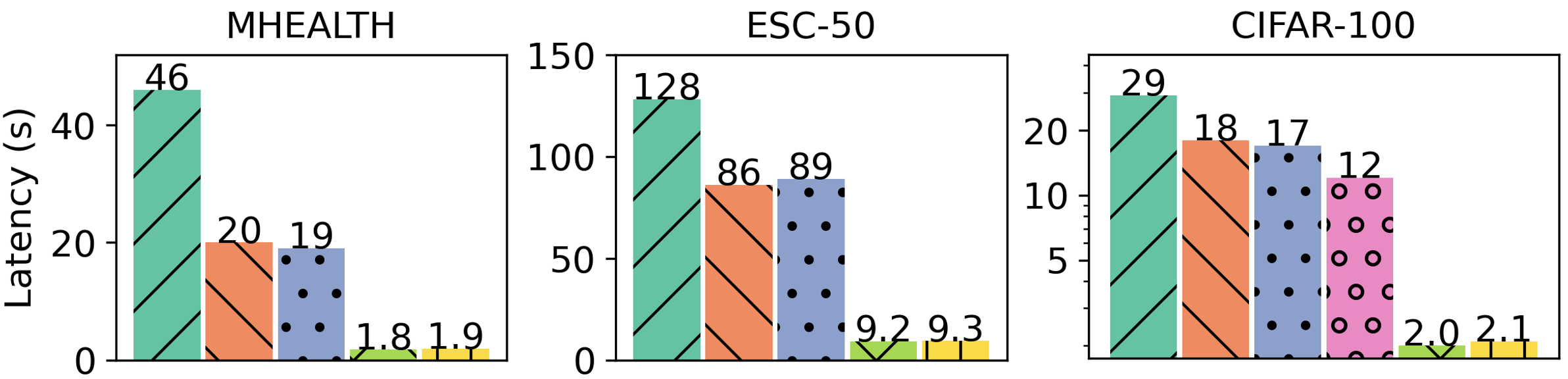} &
    \includegraphics[width=0.4\textwidth, height=1.8cm]{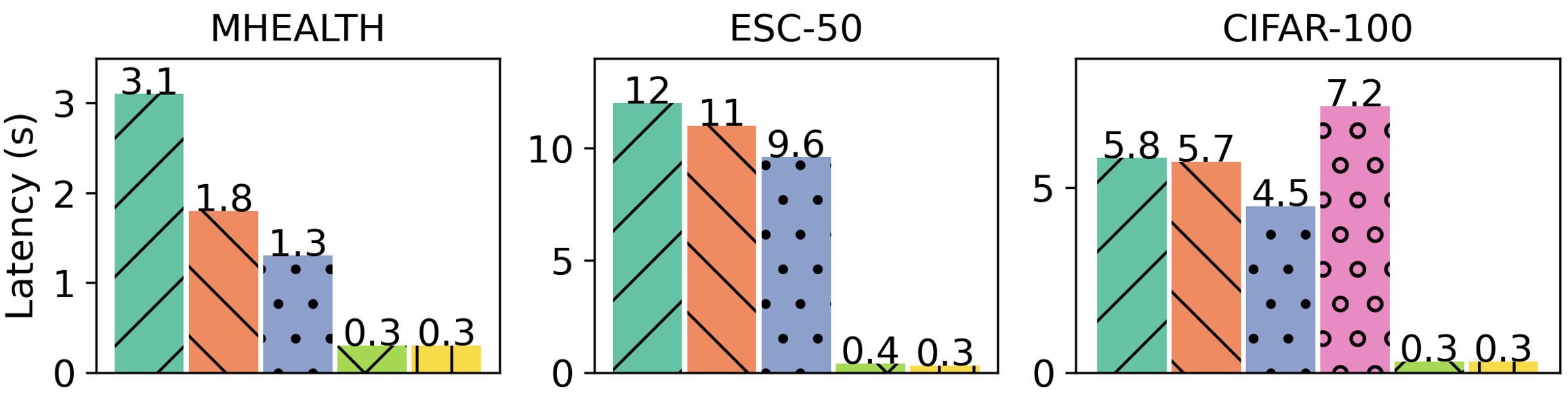} \\
    (a) {\footnotesize Latency on RPi Zero} & (b) {\footnotesize Latency on RPi 4B} & (c) {\footnotesize Latency on Jetson TX2} \\ %\hline
    \vspace{-1mm}
    \includegraphics[width=0.12\textwidth,height=1.8cm]{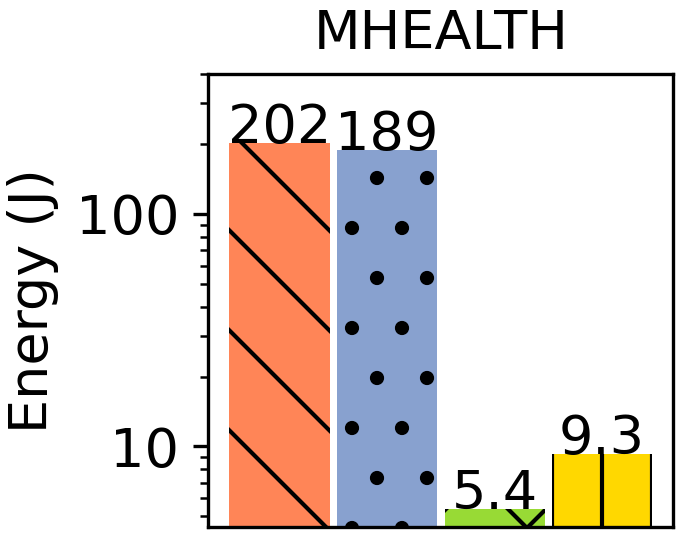} &
    \includegraphics[width=0.4\textwidth,height=1.8cm]{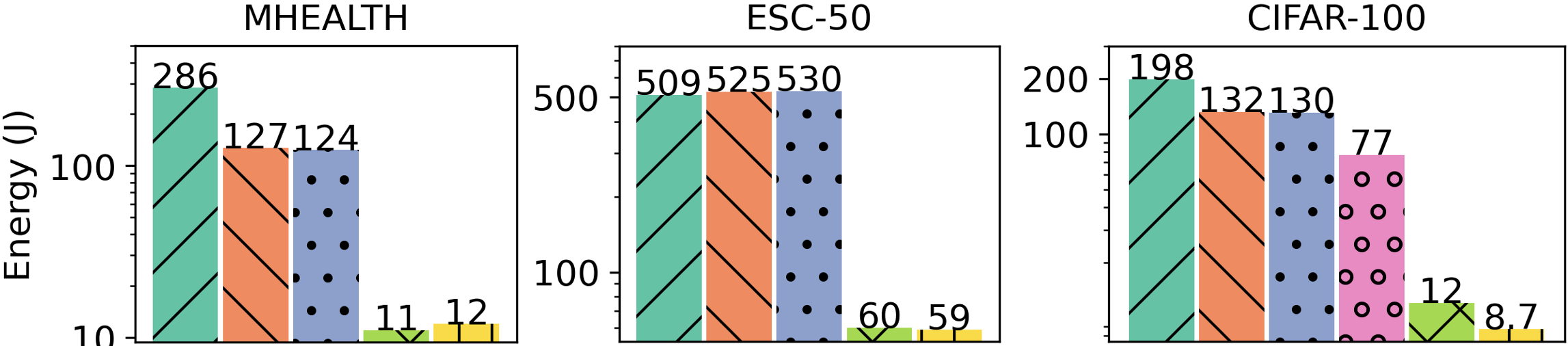} &
    \includegraphics[width=0.4\textwidth,height=1.8cm]{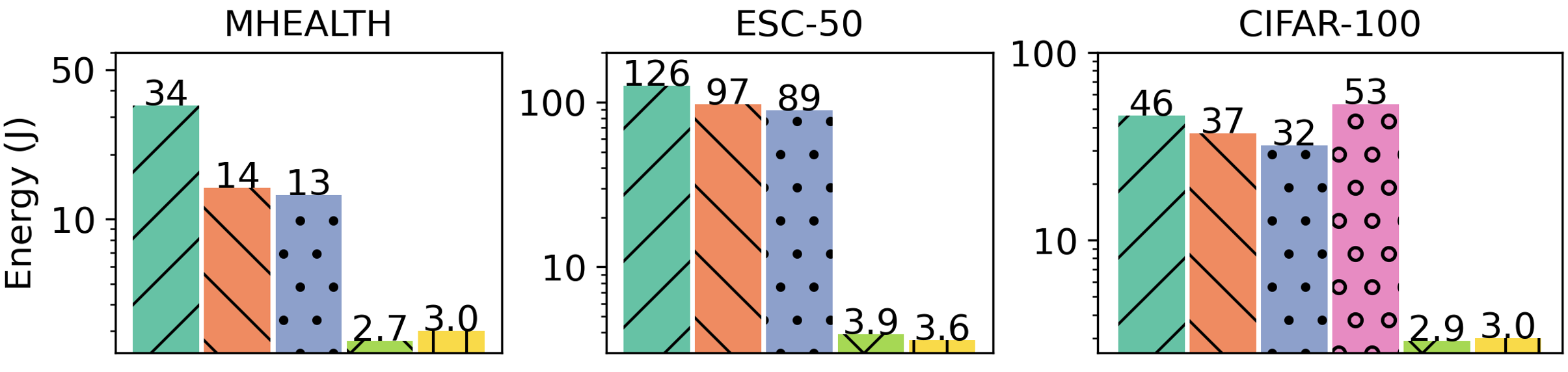} \\
    (d) {\footnotesize Energy on RPi Zero} & (e) {\footnotesize Energy on RPi 4B} & (f) {\footnotesize Energy on Jetson TX2} \\
\end{tabular}
\vspace{-4mm}
\caption{\small Latency and energy consumption to train one batch of data using {\Method} and all baselines on off-the-shelf edge platforms.}
\label{fig:energy}
\end{center}
\vspace{-2mm}
\end{figure*}

\textbf{Comparison with Fully Supervised HDC.}
Table~\ref{tbl:suphd} compares the average ACCs of supervised HDC method~\cite{kim2018efficient,hersche2022constrained} and {\Method}. Even without any supervision, {\Method} approaches the ACC of supervised HDC with \textbf{a gap of 15\%, 3\% and 6\%} on MHEALTH, ESC-50 and CIFAR-100. 
%ACC allows for mapping unsupervised clusters to supervised classes prior to accuracy calculation. 
A minimal ACC gap confirms the effectiveness of {\Method} in separating and memorizing key patterns.
%RI offers a concrete assessment by judging pairwise correctness in clustering, as discussed in Sec.\ref{sec:problem}. Table~\ref{tbl:suphd} illustrates that {\Method} makes pairwise decisions with a correctness level \textit{nearly matching} that of supervised HDC. 
%
To help explain the small ACC loss even without supervision, we visualize the confusion matrix of \Method on MHEALTH in Fig.~\ref{fig:cm}.
%the difference between fully supervised approach vs. \Method.
MHEALTH has 12 true classes (y axis), whereas {\Method} maintains 23 \prototypes in its long-term memory (x axis).
%In Fig.~\ref{fig:cm}, {\Method} has 23 \prototypes in its long-term memory to match the 12 true labels in MHEALTH.
%Similar RIs mean that even when true labels are unknown, \Method effectively separates data from various classes into different \prototypes.
ACC is evaluated by mapping the unsupervised \prototypes to true labels.
Although {\Method} cannot achieve precise label matching with true classes, it can preserve the essential patterns by using finer-grained clusters.
For example, the green box in Fig.~\ref{fig:cm} highlights a valid learning outcome, where \Method uses predicted \prototype No. 0, 9 and 23 to represent a bigger true class of ``Lying down''.

\subsection{Training Latency and Energy}
\label{sec:latency}
Fig.~\ref{fig:energy} provides comprehensive latency and energy consumption results to train one batch of samples on all three edge platforms.
\revise{For CIFAR-100, we use the most lightweight MobileNet version, V3 small~\cite{howard2019searching}, as HDnn feature extractor and NN baseline, to assess \Method's efficiency gain over the most competitive mobile computing setup.}
On RPi Zero, we report results for the relatively lightweight NN-based baselines, Finetune and LUMP~\cite{madaan2022representational}, using the smallest dataset, MHEALTH, while running CaSSLe~\cite{fini2021self} on MHEALTH would result in out-of-memory errors.
As shown in Fig.~\ref{fig:energy}, \Method is \textbf{up to 23.7x, 36.5x and 22.1x faster} to train on RPi Zero, RPi 4B and Jetson TX2, respectively, while being \textbf{up to 22.5x, {\EnergyRes} and 20.8x more energy efficient} on each, compared to the NN-based unsupervised lifelong learning baselines.
%On the RPi 4B, RPi Zero and Jetson TX2, \Method has much faster training time, achieving up to 62.4x, 23.1x, and 37.5x, faster training respectively, while consuming up to 43.2x, 20.8x, and 35.2x less energy than the NN-based unsupervised lifelong learning baselines.
%On RPi 4B (RPi Zero, Jetson TX2), \Method achieves up to 62.4x (23.1x, 37.5x) speedup in training latency while being up to 43.2x (20.8x, 35.2x) more energy efficient than the NN-based unsupervised lifelong learning baselines.
In most settings, CaSSLe~\cite{fini2021self} is the most time-consuming because of the expensive distillation. LUMP~\cite{madaan2022representational} is slightly more expensive than Finetune due to its replay mechanism. 
STAM~\cite{smith2019unsupervised}, implemented only on CPU, incurs the longest training latency on Jetson TX2, as it does not use GPU's acceleration capabilities.
\Method is clearly faster and more efficient than all NN-based unsupervised lifelong learning baselines~\cite{fini2021self,madaan2022representational,smith2019unsupervised} due to \Method's lightweight nature. 
The overhead of \Method alongside fully supervised HDC, SupHDC~\cite{kim2018efficient,hersche2022constrained}, is negligible on more powerful platforms like RPi 4B and Jetson TX2.
%These results confirm the efficiency of {\Method}, attributed to HDC's lightweight encoding and \Method's streamlined design, which contrasts with the resource-intensive gradient descent operations. 
%Even using the HDnn framework (as shown on ESC-50 and CIFAR-100), \Method only runs one set of inference on the frozen NN thus is much cheaper to execute on edge platforms.
%
Notably, in \Method, the \prototype merging step for processing about 40 LTM elements takes 7.4, 0.86 and 0.66 seconds to run on RPi Zero, RPi 4B and Jetson TX2, respectively, which only executes once every $f_{merge}$ batches. Further enhancements can be achieved using the acceleration techniques mentioned in Sec.~\ref{sec:prototype-merging}.

Fig.~\ref{fig:energy} indicates \Method improves latency and energy efficiency the most on RPi 4B, as compared to RPi Zero and Jetson TX2 that represent more limited or powerful devices.
%This outcome might seem counterintuitive, but it is logical. 
This is because the high-dimensional nature of \Method requires a fair amount of memory, thus it cannot run efficiently on the highly restricted RPi Zero.
The GPU resources on Jetson TX2 boost the NN-based baselines, narrowing the gap between them and \Method.
We expect much larger efficiency improvements  when {\Method} is accelerated using emerging in-memory computing hardware~\cite{dutta2022hdnn,xu2023fsl}.
%RPis represent the general-purpose edge computer without a GPU, while the Jetson platform, equipped with an 8GB GPU, is designed for more demanding image and audio processing.
%As a result, the Jetson TX2 exhibits significantly shorter latency and lower energy consumption compared to RPi.
%while LUMP is slightly slower than Finetune due to the replay mechanism.
%On RPi, {\Method} achieves 1.84-87.1x speedup in training while being up to 63.5x more energy efficient than the NN-based baselines on all datasets.
%On the Jetson platform, {\Method} demonstrates speedups of up to 20.0x while exhibiting energy efficiency gains of up to  42.6x. 
%These results confirm the efficiency of {\Method}.
%The only exception is observed with  MHEALTH on Jetson, where {\Method} takes approximately 2 seconds longer to train a single batch. We believe this can be attributed to the combined influence of the lightweight LSTM backbone and the complexity of the spatiotemporal encoder when processing timeseries data.

\begin{figure}[t]
\begin{center}
\includegraphics[width=0.47\textwidth]{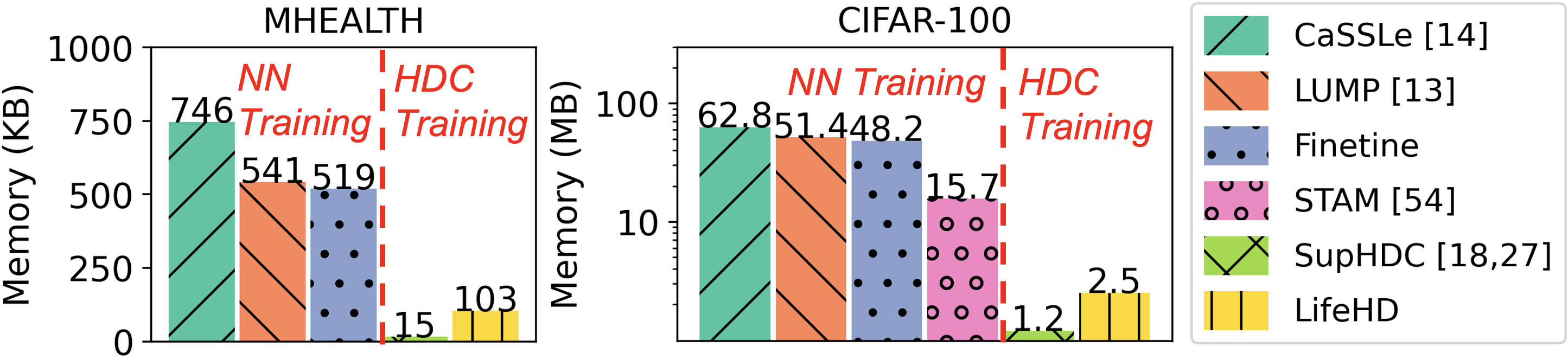}
\vspace{-3mm}
\caption{\small Peak memory footprint of all methods on MHEALTH (left) and CIFAR-100 (right) with batch size of 1. The results are representative for time series data and image data.}
\label{fig:memory}
\end{center}
\vspace{-4mm}
\end{figure}

%\vspace{-1mm}
\subsection{Memory Usage}
Fig.~\ref{fig:memory} provides a comprehensive summary of peak memory footprint for all methods on MHEALTH and CIFAR-100. 
%We also assess \Method's memory overhead by comparing it with NN inference and SupHDC.
We categorize the methods into NN training (Finetune, LUMP~\cite{madaan2022representational}, CaSSLe~\cite{fini2021self}, STAM~\cite{smith2019unsupervised}) and HDC training (Supervised HDC~\cite{kim2018efficient,hersche2022constrained} and our \Method). %we calculate NN inference memory based on the largest-sized activation and 
\revise{Following~\cite{kwon2023lifelearner}, we calculate the peak memory of NN training as the sum of model, optimizer and activation memories, plus additional memory consumption for lifelong learning.} Specifically, CaSSLe~\cite{fini2021self} requires additional memory for training a predictor and inference from a frozen model, LUMP~\cite{madaan2022representational} needs extra memory for replay. 
\revise{For HDC-based methods, each dimension of the \prototype is represented as a signed integer and stored in a byte. In addition to the working and long-term memories, we also consider the storage of bipolar level and ID hypervectors for encoding, and the frozen MobileNet for HDnn encoding in CIFAR-100.}
%\Method uses integer numbers for full-precision \prototypes in the working memory and binary digits for long-term memory. 
Notice that our focus here is on comparing full-precision memory usage, and optimization techniques like quantization can be applied to all methods in the future.

The results in Fig.~\ref{fig:memory} highlight \Method's memory efficiency.
\Method conserves 80.1\%-86.2\% and 84.1\%-96.0\% of memory compared to NN training baselines on MHEALTH (non-HDnn) and CIFAR-100 (HDnn), respectively. % and even 33.3\% memory in contrast to NN inference. 
%because it does not require gradient descent.
This remarkable efficiency stems from \Method's HDC design, which dispenses with the memory-intensive gradient descents in NNs.
STAM~\cite{smith2019unsupervised}, with its hierarchical and expandable memory structure, consumes 6.3x the memory of \Method, as it stores raw image patches across all hierarchies. 
%\Method requires the same 2.4 MB for frozen ResNet-18 inference and an extra 4 MB for cluster HV storage. 
Compared to fully supervised HDC, SupHDC~\cite{kim2018efficient,hersche2022constrained}, \Method introduces a modest memory increase to accomplish the challenging task of organizing label-free \prototypes.
\Method proves advantageous for edge applications with only 103 KB and 2.5 MB of peak memory required for MHEALTH and CIFAR-100.

%\vspace{-2mm}
\subsection{Ablation Studies}
\label{sec:ablation-studies}
\begin{figure}[t]
   \centering
    \setlength{\tabcolsep}{3pt}
\begin{tabular}{cc}
        \includegraphics[height=2.1cm]{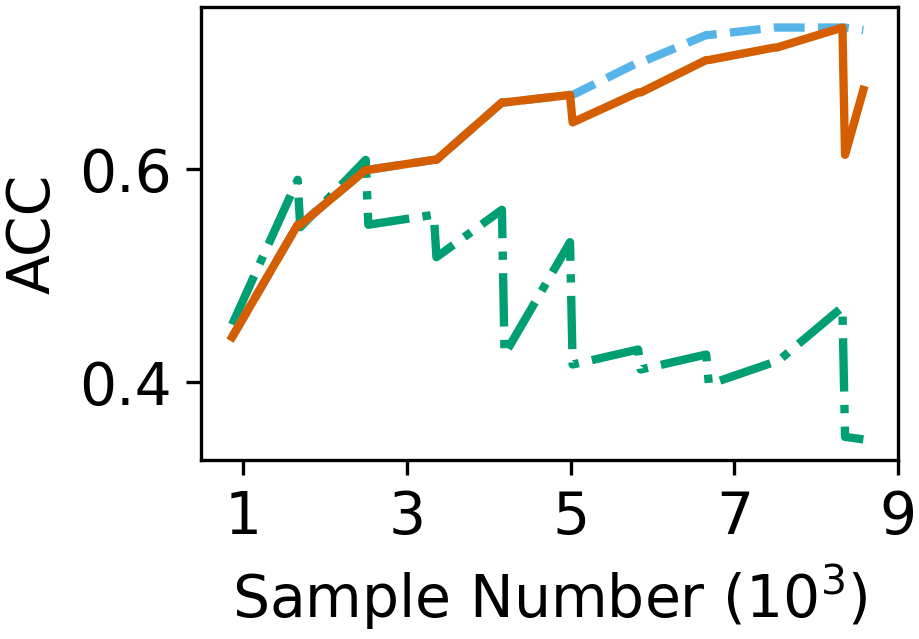} &
        \includegraphics[height=2.1cm]{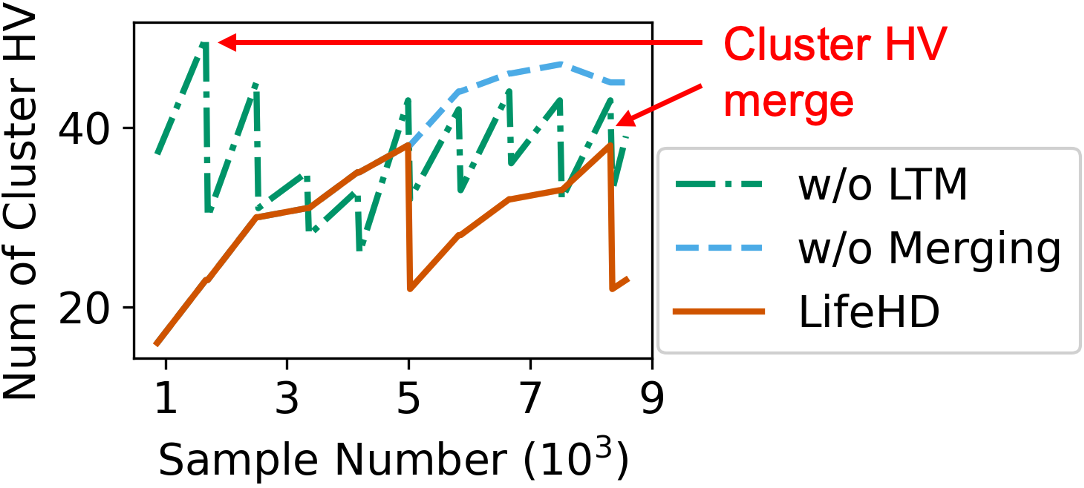} \\
\end{tabular}
\vspace{-4mm}
    \caption{ \small Ablation study of {\Method} on MHEALTH. The number of \prototypes reported for {\Method} without long-term memory (LTM) is for the working memory, since no LTM is allowed.}
        \label{fig:ablation}
        \vspace{-4mm}
\end{figure}

The design of {\Method} consists of several key elements: the two-tier memory organization, novelty detection and online update, and \prototype merging that manipulates past patterns.
We conduct experiments to assess the contribution of each element.
Using the configuration in Table~\ref{tbl:parameters}, we evaluate the performance of (i) {\Method} without long-term memory, using only a single layer memory, (ii) {\Method} without merging, employing only novelty detection, online update and forgetting, and (iii) complete {\Method}.
We present the ACC and the number of \prototypes in LTM during MHEALTH training in Fig.~\ref{fig:ablation}, chosen as a representative scenario.
{\Method} without LTM (green dashdot line) forces \prototype merging to take place in working memory, where the large number of temporary \prototypes creates less important nodes in the graph and corrupts the graph-based merging process, as shown in Fig.~\ref{fig:ablation} (left). This necessitates the design of the two-tier memory architecture and merging with LTM elements.
{\Method} without merging (blue dashed line) consumes 1x more memory in the LTM, making it unsuitable for resource-constrained edge devices.
Our design of \Method (red solid line) strategically combines similar \prototypes with minor loss on the clustering quality, achieving ACC similar to those without merging while conserving memory storage.
%This is because, NMI allows matching multiple smaller clusters to one big class. Hence NMI tends to favor having more clusters as long as the clusters remain distinct.
%The value of merging is to strategically combine similar \prototypes with minor loss on the clustering quality, while saving the memory storage for future samples.

%\vspace{-2mm}
\subsection{Sensitivity Analysis}
\label{sec:sensitivity-analysis}

\begin{figure}[!t]
   \centering
    \setlength{\tabcolsep}{0.2pt}
\begin{tabular}{cccc}
        \vspace{-2mm}
        \includegraphics[width=0.15\textwidth, height=1.6cm]{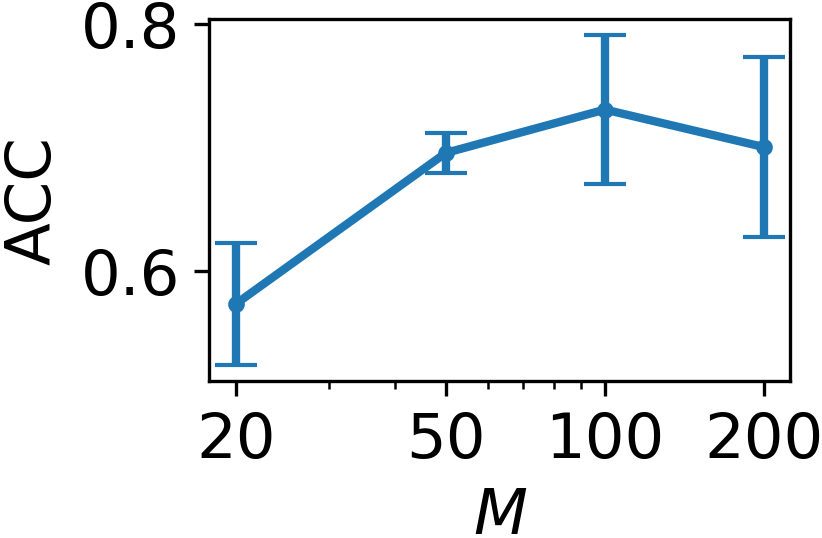} &
        \includegraphics[width=0.15\textwidth, height=1.6cm]{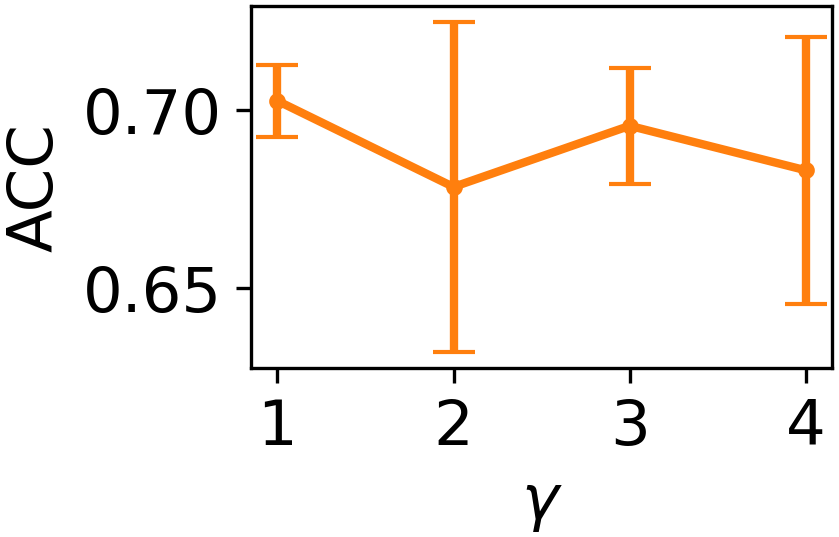} &
        \includegraphics[width=0.15\textwidth, height=1.6cm]{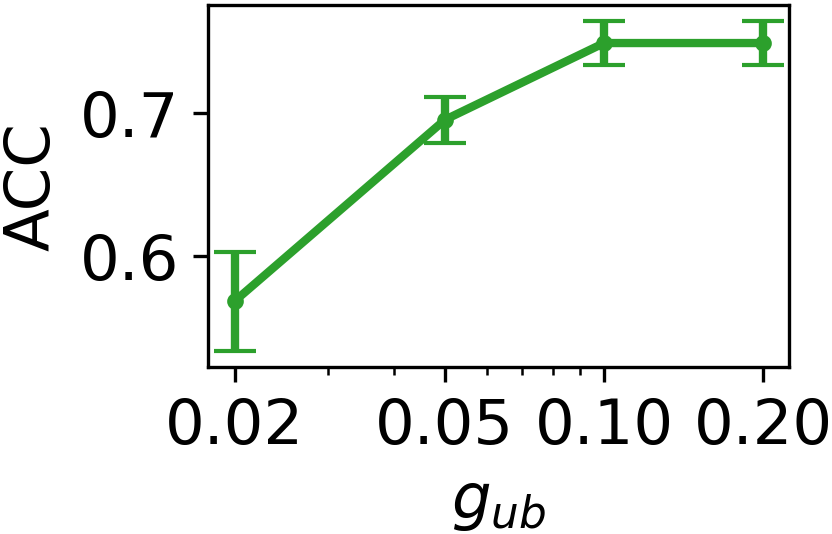} \\ 
        {\footnotesize (a) Working memory size} &
        {\footnotesize (b) Novelty threshold} &
        {\footnotesize (c) Merging sensitivity} \\
        \vspace{-2mm}
        \includegraphics[width=0.15\textwidth, height=1.6cm]{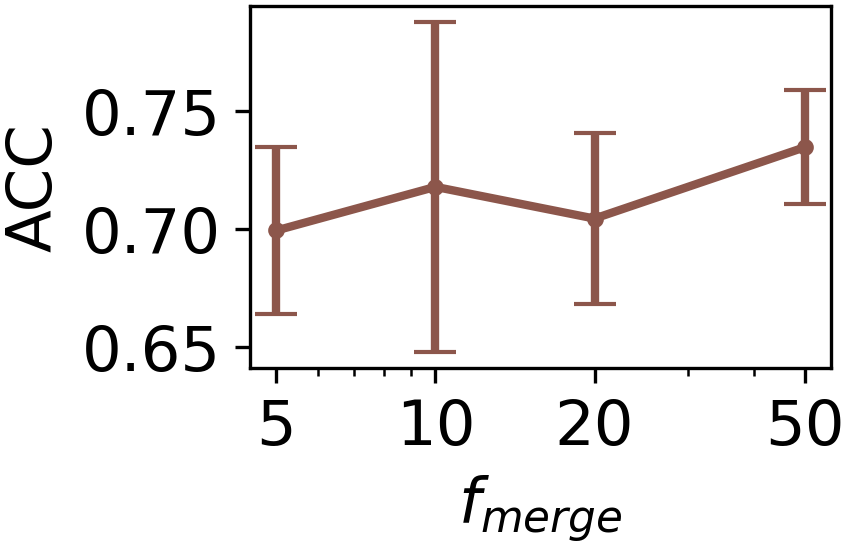} &
        \includegraphics[width=0.15\textwidth, height=1.6cm]{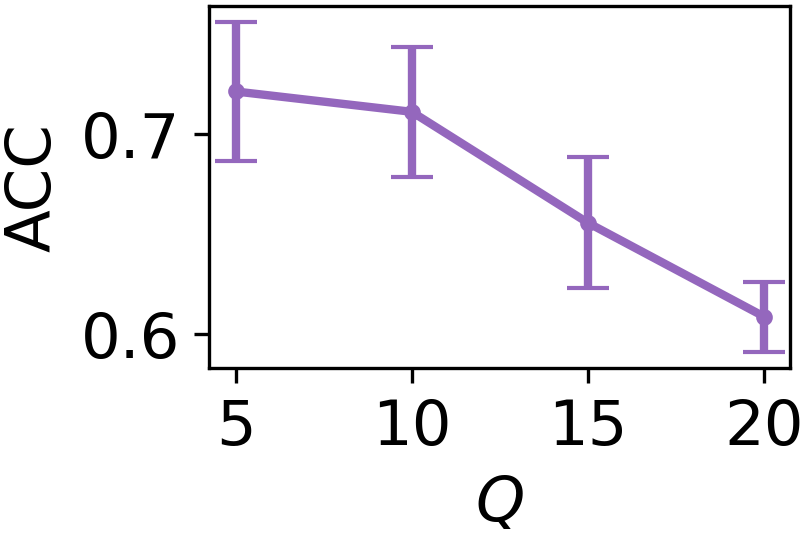} &
        \includegraphics[width=0.15\textwidth, height=1.6cm]{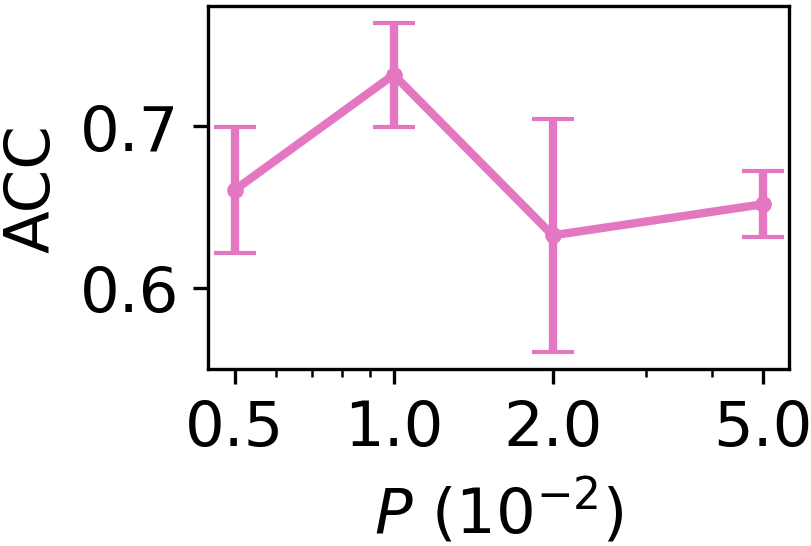} \\ 
        {\footnotesize (d) Merge frequency} &
        {\footnotesize (e) Encoding level} &
        {\footnotesize (f) Encoding flipping ratio} \\
\end{tabular}
\vspace{-4mm}
    \caption{ \small Sensitivity of various hyperparameters in {\Method}, using MHEALTH dataset as a representative.}
    \label{fig:sensitivity}
    \vspace{-5mm}
\end{figure}

%We conduct comprehensive sensitivity analyses on the important hyperparameters in {\Method}. For all the experiments, we repeat three random trials.
Fig.~\ref{fig:sensitivity} summarizes the sensivitity results of key parameters in \Method, while the less sensitive ones such as $\alpha$ and $hit_{th}$ are omitted due to space limitation. The default setting is the same as in Table~\ref{tbl:parameters}.

\textbf{Working Memory Size.}
%We mainly focus on the working memory size as the long-term memory is accessed less frequently. A long-term memory with size $L=50$ is satisfied for all experiments.
Fig.~\ref{fig:sensitivity} (a) shows ACCs using working memory sizes of 20, 50, 100 and 200. %We fix $\gamma=3.0$ and $g_{ub}=0.9$.
In general, a larger working memory allows more temporary \prototypes at the cost of higher memory consumption.
$M=100$ produces optimal results, while further increasing the memory size reduces clustering quality. This occurs because excessively large working memory retains outdated prototypes, degrading lifelong learning performance.
% In all four scenarios, using $M=100$ or $M=200$ enjoy the best ultimate results. It can be observed that further increasing the working memory size will not bring additional gain, but may even degrade the clustering quality. 
%When the limit of memory is reached, {\Method} replaces the least recently used unit.

\textbf{Novelty Threshold.}
In Fig.~\ref{fig:sensitivity} (b), we present the final ACCs for different novelty detection thresholds ($\gamma$). 
\revise{A lower $\gamma$ results in more frequent novelty detections and increased loads on the working memory, while a higher $\gamma$ may lead to overlooking significant changes.
Remarkably, {\Method} demonstrates resilience to variations in $\gamma$, a phenomenon that we attribute to the combined impact of novelty detection and merging processes.}
%A smaller $\gamma$ results in finer-grained \prototypes because a new one is created whenever a novelty flag is triggered. However, this increases the load on the working memory, where only a limited number of \prototypes can be stored. Conversely, a higher $\gamma$ detects fewer new items, potentially overlooking significant changes in patterns. It is essential to find the right balance by jointly tuning the working memory size and the novelty detection threshold to align with the expected changes in the environment.
%, as they collaboratively determine the lifelong learning performance.

\textbf{Merging Sensitivity.}
%$g_{ub}$ is the major nob that controls merging sensitivity. 
% Used in the last step in \prototype merging (Sec.~\ref{sec:prototype-merging}), $g_{ub}$ determines the number of clusters $k$ to merge into based on the eigenvalues of the graph Laplacian, i.e.,  $k = \max_{i \in [L]} \lambda_{i} \leq g_{ub}$.
Fig.~\ref{fig:sensitivity} (c) shows ACC using various merging thresholds ($g_{ub})$.
$g_{ub}$ determines the number of clusters ($k$) to merge in the \prototype merging step (Sec.~\ref{sec:prototype-merging}).
A low value for $g_{ub}$ results in overly aggressive merging, leading to the fusion of dissimilar \prototypes and a degraded ACC.
A larger $g_{ub}$ adopts a conservative merging strategy and encourages finer-grained clusters, albeit at the expense of increased resource demands.
%A more moderate merging strategy with $g_{ub}=0.2$ preserves overall clustering quality using more clusters, as NMI encourages finer-grained clusters. Nevertheless, maintaining a large number of \prototypes imposes a burden on both computational and memory resources.

\textbf{Merging Frequency.}
Fig.~\ref{fig:sensitivity} (d) shows the final ACCs for different merging frequencies ($f_{batch}$). \Method shows its robustness across various $f_{batch}$ values, partly due to the presence of $g_{ub}$ to prevent aggressive merging. Less frequent merging (larger $f_{batch}$) raises the risk of forgetting important patterns as of memory constraints. More frequent merging (smaller $f_{batch}$) increases the computational burden due to the spectral clustering-based algorithm.

\textbf{Encoding Level and Flipping Ratio for Spatiotemporal Encoding.}
Fig.~\ref{fig:sensitivity} (e) and (f) show the ACCs for various quantization encoding levels ($Q$) and flipping ratios ($P$) during the spatiotemporal encoding.
Both parameters are important for preserving the similarity in HD-space after encoding.
Optimal $Q$ depends on the sensor sensitivity, with finer-grained sensors requiring more quantization levels.
$P$ determines the similarity between adjacent levels of hypervectors. For personal health monitoring, such as MHEALTH, $Q=10, P=0.01$ usually gives the best results.

%%%%%%%%%%%%%%%%%%%%%%%%%%%%%%%%%%%%%%%%%%%%%%%%%%%%%%%
% Evaluation
%%%%%%%%%%%%%%%%%%%%%%%%%%%%%%%%%%%%%%%%%%%%%%%%%%%%%%%
%\vspace{-2mm}
\section{Evaluation of {\SemiMethod} and {\EffMethod}}
\label{sec:evaluation-variants}

In this section, we compare {\SemiMethod} and {\EffMethod}, our proposed extensions from {\Method}, with existing designs that are similar. % during certain conditions in IoT edge applications.
%, i.e., scarce label availability and power constraint.
%Without such condition, {\SemiMethod} and {\EffMethod} offer identical performance as {\Method} in the major evaluation setup.

\begin{figure}[t]
\begin{center}
\begin{tabular}{c}
    \vspace{-1mm}
    \includegraphics[width=0.4\textwidth, height=2.4cm]{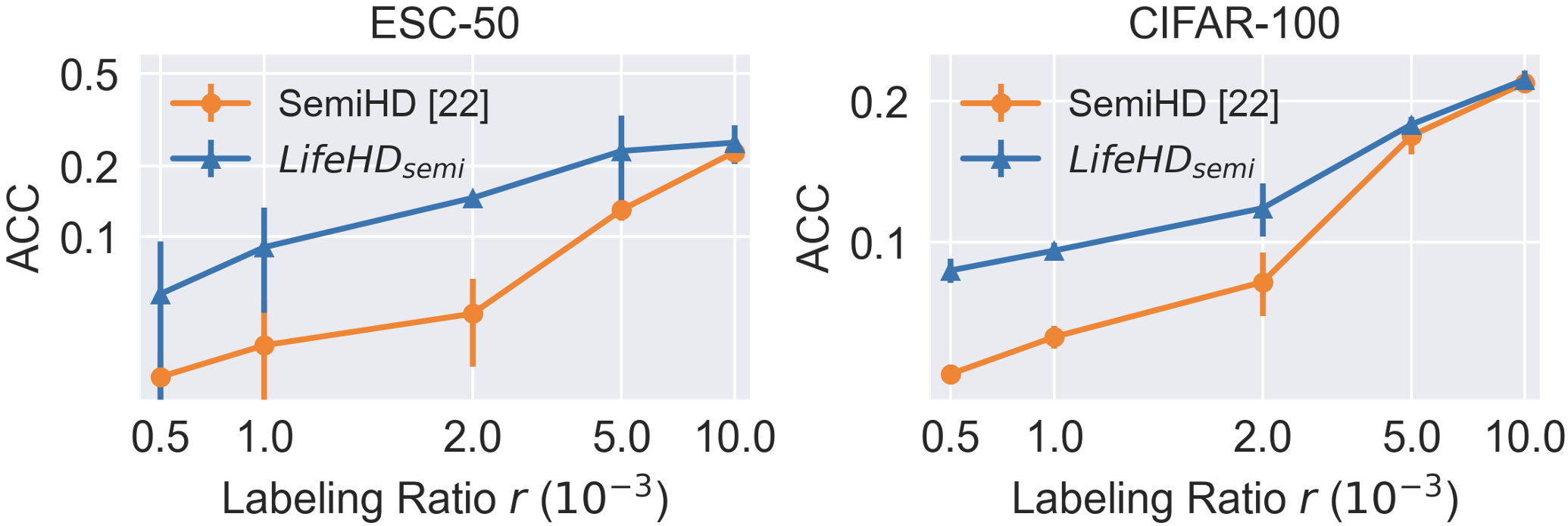} \\
    (a) {\small Gains of {\SemiMethod} over SemiHD~\cite{imani2019semihd} in lower labeling ratios.} \\
    \vspace{-1mm}
    \includegraphics[width=0.4\textwidth, height=2.4cm]{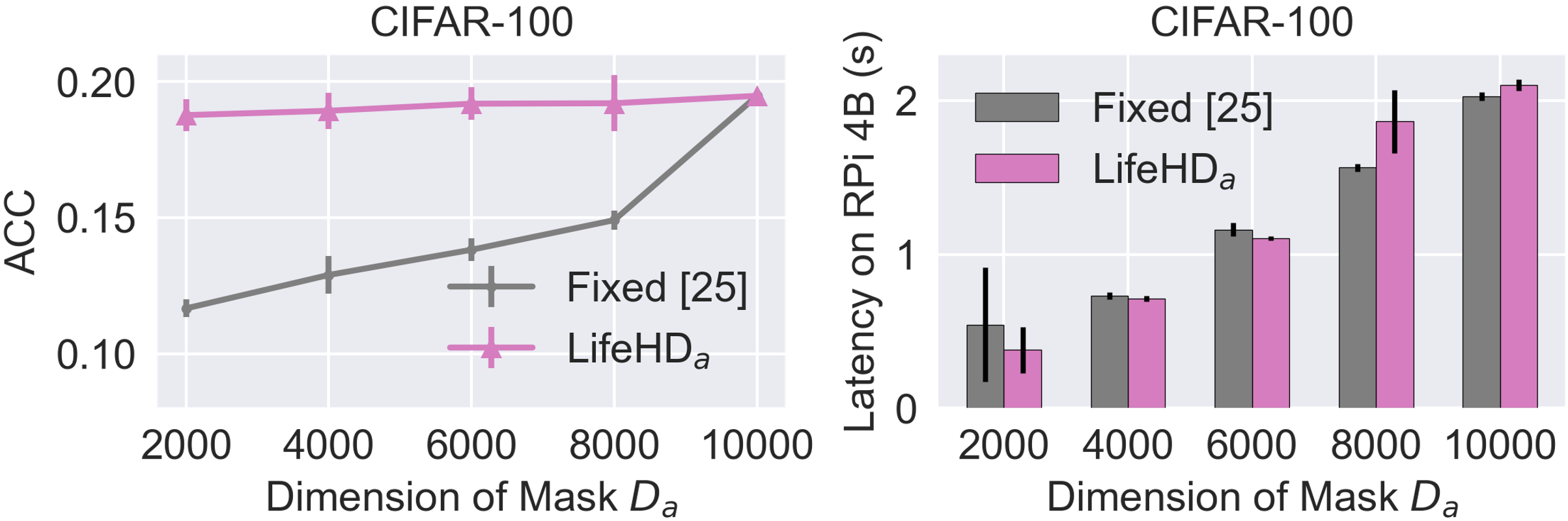} \\
    (b) {\small {\EffMethod} vs. using fixed mask (Fixed)~\cite{khaleghi2020prive} under various $D_a$.}
\end{tabular}
\vspace{-4mm}
\caption{\small Results of \SemiMethod and \EffMethod compared to existing HDC techniques for similar goals.}
\label{fig:variants}
\end{center}
\vspace{-4mm}
\end{figure}

%\vspace{-2mm}
\textbf{Performance of {\SemiMethod}.}
%\label{sec:semi_perf}
To evaluate {\SemiMethod} in a low-label scenario, we compare it with SemiHD~\cite{imani2019semihd}, which is the state-of-the-art HDC method for semi-supervised learning.
We adapt SemiHD~\cite{imani2019semihd} for single-pass settings, introducing a pseudolabel assignment threshold. When the cosine similarity of an unlabeled sample to the nearest class hypervector surpasses the threshold, we assign that class as its pseudolabel. The sample is then employed to update the class hypervector in SemiHD. We explore various threshold values and choose the optimal result for comparison.
Fig.~\ref{fig:variants} (a) compares {\SemiMethod} and SemiHD~\cite{imani2019semihd} on ESC-50 and CIFAR-100 across various labeling ratios $r < 0.01$.
The advantages of {\SemiMethod} are most prominent when labels are limited, the weakly supervised scenario is \SemiMethod's major focus. %With a labeling ratio of $r=0.002$ on ESC-50, {\SemiMethod} improves RI by 0.37 compared to SemiHD, representing a 37\% increase in pairwise correctness. Similarly, on CIFAR-100 with $r=0.001$, \SemiMethod brings a 50\% increase in pairwise clustering decisions.
\SemiMethod improves ACC by up to 10.25\% and 3.6\% on ESC-50 and CIFAR-100 respectively.
This outcome arises from the unsupervised nature of {\Method}, allowing it to autonomously organize prominent \prototypes, especially when all samples from a class lack labels. As the labeling ratio increases, \SemiMethod's advantage over SemiHD diminishes, because more labels bolster SemiHD's performance. %, while unsupervised self-organization's relative effectiveness wanes.
%While SemiHD demonstrates an advantage with more labeled samples, the benefits of {\SemiMethod} become evident when labels are scarce, particularly in scenarios where all samples from one class are unlabeled.
%The observations validate {\SemiMethod} in utilizing the supervision power from very few samples.

%\begin{figure}[t]
%    \centering
%    \setlength{\tabcolsep}{0.2pt}
%\begin{tabular}{cc}
%        \includegraphics[width=0.21\textwidth, height=2.5cm]{figs/RI_ESC50.png} &       
%        \includegraphics[width=0.21\textwidth, height=2.5cm]{figs/RI_CIFAR100.png} \\
%\end{tabular}
%     \vspace{-6mm}
%      \caption{\small Results of {\SemiMethod} vs. SemiHD~\cite{imani2019semihd}.}
%      \vspace{-2mm}
%    \label{fig:result_semi}
% \end{figure}

%\vspace{-2mm}
\textbf{Performance of {\EffMethod}.}
%\label{sec:eff_perf}
\EffMethod provides an interface to trade minimal performance loss for efficiency gains, by adaptively pruning out the insignificant dimensions.
We compare {\EffMethod} with previous HDC works employing a fixed mask throughout training~\cite{khaleghi2020prive}, and the results are presented in Fig.~\ref{fig:variants} (b) for CIFAR-100, including ACC and training latency per batch on RPi 4B.
Fixed masks negatively impact HDC learning, especially with smaller dimensions. Such masks fail to adapt to new hypervectors in class-incremental streams, where less significant dimensions may become crucial later in training. {\EffMethod} addresses this issue by adjusting the mask upon novelty detection, leading to a degradation of only 0.71\% in ACC and 4.5x efficiency gain compared to the complete \Method, using only 20\% of the full HD dimension of \Method.
%Additionally, Fig.~\ref{fig:variants} (b) illustrates that reducing the dimension of the mask linearly significantly saves training time on RPi 4B, demonstrating near-linear efficiency improvements. 
The overhead of adaptively adjusting the mask is negligible when novelty detection occurs infrequently.

%\begin{figure}[!t]
%   \centering
%    \setlength{\tabcolsep}{0.2pt}
%\begin{tabular}{ccc}     
        %\includegraphics[width=0.21\textwidth, height=2.3cm]{figs/esc50-eff.png} &
%        \includegraphics[width=0.21\textwidth, height=2.3cm]{figs/stream51-eff.png} &
        %\includegraphics[width=0.21\textwidth, height=2.3cm]{figs/eff-time-esc50.png} &  
 %       \includegraphics[width=0.21\textwidth, height=2.3cm]{figs/eff-time-stream51.png}  \\
%\end{tabular}
%\vspace{-5mm}
%    \caption{\small Left: NMI results of {\EffMethod} vs. using fixed mask (Fixed)~\cite{khaleghi2020prive} under various dimension $D_a$. Right: training latency per epoch on RPi 4B.}
%        \label{fig:result_eff}
%        \vspace{-6mm}
%\end{figure}

%%%%%%%%%%%%%%%%%%%%%%%%%%%%%%%%%%%%%%%%%%%%%%%%%%%%%%%
% Discussion
%%%%%%%%%%%%%%%%%%%%%%%%%%%%%%%%%%%%%%%%%%%%%%%%%%%%%%%
%\vspace{-2mm}
\section{Discussions and Future Works}
\label{sec:discussion}

\revise{\textbf{Problem Scale.}}
\revise{One limitation of \Method is the relative small problem scale (e.g., the image size of CIFAR-100 is restricted to 32x32) due to the essential difficulty of unsupervised lifelong learning problem, including single-pass non-iid data and no supervision. For the same reason, there remains a disparity in accuracy between unsupervised lifelong learning and fully supervised NNs, as substantiated by prior research~\cite{smith2019unsupervised,madaan2022representational}. In order to scale \Method to more challenging applications such as self-driving vehicles, one possible direction is to leverage the pretrained foundation model as a frozen feature extractor in the HDnn framework, which we leave for future investigation.}

\revise{\textbf{Hyperparameter Tuning.}
While we recognize that hyperparameters can influence the performance of \Method, such an issue is not exclusive to \Method, but has persistently been a challenge in machine learning research~\cite{bischl2023hyperparameter}.
In \Method, the impact of hyperparameters can be mitigated through pre-deployment evaluation and component co-design.
For example, encoding parameters such as $Q,P$ can be tuned on similar health monitoring data sources prior to deployment. Meanwhile, the component of \prototypes merging can increase \Method's resiliency to the novelty detection threshold $\gamma$, as a higher quantity of novel clusters can be merged in later stage of learning.}

\revise{\textbf{Limitations of HDC.}
HDC serves as the fundamental core of \Method.
While HDC shows promise with its notable lightweight design, it is burdened by several limitations that remain active areas of research.
First, for complex datasets like audio and images, HDC requires a pretrained feature extractor (the HDnn encoding) which may not exist for certain applications.
Moreover, akin to any other architecture, HDC vectors face capacity limitations determined by the dimension of HD space, encoding method, and potential noise levels in the input data~\cite{thomas2021theoretical}. Due to these factors, careful evaluation and sometimes manual feature engineering are required to successfully deploy HDC for new applications.}
% Feature extractor, HDC capacity

%\textbf{Discussion of Hardware Platform.}

\revise{\textbf{Future Works.} Although \Method focuses on single-device lifelong learning for classification tasks, the method can be extended for other types of tasks and learning settings, such as federated learning and reinforcement learning. We leave the investigation of these topics for future work.}
%Other types of classification, like object detection. Federated learning, other applicatioins like IIoT, reinforcement learning.

%%%%%%%%%%%%%%%%%%%%%%%%%%%%%%%%%%%%%%%%%%%%%%%%%%%%%%%
% Conclusion
%%%%%%%%%%%%%%%%%%%%%%%%%%%%%%%%%%%%%%%%%%%%%%%%%%%%%%%
%\vspace{-2mm}
\section{Conclusion}
\label{sec:conclusion}

The ability to learn continuously and indefinitely in the presence of change, and without access to supervision, on a resource-constrained device is a crucial trait for future sensor systems.
%The development of systems which evince such ``liflelong learning'' capabilities is a critical step in bringing the promises of AIoT to fruition. 
In this work, we design and deploy the first end-to-end system named {\Method} to learn continuously from real-world data streams without labels. Our approach is based on Hyperdimensional Computing (HDC), an emerging neurally-inspired paradigm for lightweight edge computing.
\Method is built on a two-tier memory hierarchy including a working and a long-term memory, with collaborative components of novelty detection, online \prototype update and \prototype merging for optimal lifelong learning performance.
%, which represents information using high-dimensional, distributed, and low-precision vectors. In our approach, all computation is performed on these HD representations of data, which are obtained from raw sensory input via a specialized ``encoding function.'' The low-precision and distributed nature of the HD representations makes them well-suited to parallel processing and a promising alternative to conventional approaches based on deep neural networks. \textcolor{red}{Maybe say something about results here}
%Building on memory hierarchy from cognitive science, our system contains two ``tiers'' of memory to capture fine-grained and coarser patterns. We further propose a novel merging mechanism based on spectral clustering \cite{von2007tutorial}. % for merging short and long-term memories. 
%a short-term working memory that is responsible for constructing fine-grained summaries of the data stream over short time-horizons, and a long-term memory which stores coarser summaries extracted over time from the working memory. We propose a novel mechanism based on spectral clustering \cite{von2007tutorial} for merging short and long-term memories. 
%In contrast to existing NN-based lifelong learning techniques, {\Method} is a lightweight but effective lifelong learner.
We further propose two extensions to {\Method}, \SemiMethod and \EffMethod, to handle scarce labeled samples and power constraints.
Practical deployments on typical edge platforms and three IoT scenarios demonstrate {\Method}'s improvement of up to {\AccRes} on unsupervised clustering accuracy and up to {\EnergyRes} on energy efficiency compared to state-of-the-art NN-based unsupervised lifelong learning baselines~\cite{fini2021self,madaan2022representational,smith2019unsupervised}.
%Further experiments show that {\SemiMethod} improves the pairwise clustering accuracy by up to 50\% over the SemiHD~\cite{imani2019semihd} baseline, while {\EffMethod} preserves performance of more than 90\% with only 20\% of full \Method capacity.
%\revise{Add limitation of the method.}

%\vspace{-2mm}

%In contrast to most prior work on lifelong learning in IoT settings, our basic approach is fully unsupervised. 
%That is to say, it requires no direct intervention to label data. However, in many settings small quantities of such data is available. 
%We propose an extension to our basic model, that we call {\SemiMethod} that can exploit this data and improve performance over the fully unsupervised baseline. \textcolor{maybe quantify here}. Additionally, to allow one to trade-off between resource usage, power in particular, and predictive performance, we propose a further extension which adaptively modifies the HD encoding dimension using an approach based on online masking, that we show improves \textcolor{red}{something about performance improvement here}.

\begin{acks}
The authors would like to thank the anonymous shepherd, reviewers, and our colleague Xiyuan Zhang for their valuable feedback.
This work was supported in part by National Science Foundation under Grants \#2003279, \#1826967, \#2100237, \#2112167, \#1911095, \#2112665, and in part by PRISM and CoCoSys, centers in JUMP 2.0, an SRC program sponsored by DARPA.
\end{acks}

\bibliographystyle{ACM-Reference-Format}
\bibliography{sample-base}

%\appendix

\end{document}

%% file: preamble.tex
\def\R{{\mathbb{R}}}

\def\X{{\mathcal X}}

\def\H{{\mathcal H}}

\def\G{{\mathcal G}}

\def\D{{\mathcal D}}

\DeclareMathOperator*{\argmax}{argmax}
\DeclareMathOperator*{\argmin}{argmin}

%% file: main.bbl

\begin{thebibliography}{67}

%%% the Web address.

\ifx \showCODEN    \undefined \def \showCODEN     #1{\unskip}     \fi
\ifx \showDOI      \undefined \def \showDOI       #1{#1}\fi
\ifx \showISBNx    \undefined \def \showISBNx     #1{\unskip}     \fi
\ifx \showISBNxiii \undefined \def \showISBNxiii  #1{\unskip}     \fi
\ifx \showISSN     \undefined \def \showISSN      #1{\unskip}     \fi
\ifx \showLCCN     \undefined \def \showLCCN      #1{\unskip}     \fi
\ifx \shownote     \undefined \def \shownote      #1{#1}          \fi
\ifx \showarticletitle \undefined \def \showarticletitle #1{#1}   \fi
\ifx \showURL      \undefined \def \showURL       {\relax}        \fi
% invisible to TeX
\providecommand\bibfield[2]{#2}
\providecommand\bibinfo[2]{#2}
\providecommand\natexlab[1]{#1}
\providecommand\showeprint[2][]{arXiv:#2}

\bibitem[jet(2023)]%
        {jetsontx2}
 \bibinfo{year}{2023}\natexlab{}.
\newblock \bibinfo{title}{{Jetson TX2 Module}}.
\newblock
  \bibinfo{howpublished}{\url{https://developer.nvidia.com/embedded/jetson-tx2}}.
\newblock
\newblock
\shownote{[Online]}.


\bibitem[rpi(2023a)]%
        {rpi4b}
 \bibinfo{year}{2023}\natexlab{a}.
\newblock \bibinfo{title}{{Raspberry Pi 4B}}.
\newblock
  \bibinfo{howpublished}{\url{https://www.raspberrypi.com/products/raspberry-pi-4-model-b/}}.
\newblock
\newblock
\shownote{[Online]}.


\bibitem[rpi(2023b)]%
        {rpi0}
 \bibinfo{year}{2023}\natexlab{b}.
\newblock \bibinfo{title}{{Raspberry Pi Zero 2 W}}.
\newblock
  \bibinfo{howpublished}{\url{https://www.raspberrypi.com/products/raspberry-pi-zero-2-w/}}.
\newblock
\newblock
\shownote{[Online]}.


\bibitem[Avargu{\`e}s-Weber et~al\mbox{.}(2012)]%
        {avargues2012simultaneous}
\bibfield{author}{\bibinfo{person}{Aurore Avargu{\`e}s-Weber} {et~al\mbox{.}}}
  \bibinfo{year}{2012}\natexlab{}.
\newblock \showarticletitle{Simultaneous mastering of two abstract concepts by
  the miniature brain of bees}.
\newblock \bibinfo{journal}{\emph{Proceedings of the National Academy of
  Sciences}} \bibinfo{volume}{109}, \bibinfo{number}{19}
  (\bibinfo{year}{2012}), \bibinfo{pages}{7481--7486}.
\newblock


\bibitem[Baddeley(1992)]%
        {baddeley1992working}
\bibfield{author}{\bibinfo{person}{Alan Baddeley}.}
  \bibinfo{year}{1992}\natexlab{}.
\newblock \showarticletitle{Working memory}.
\newblock \bibinfo{journal}{\emph{Science}} \bibinfo{volume}{255},
  \bibinfo{number}{5044} (\bibinfo{year}{1992}), \bibinfo{pages}{556--559}.
\newblock


\bibitem[Banos and Saez(2014)]%
        {misc_mhealth_dataset_319}
\bibfield{author}{\bibinfo{person}{Garcia~Rafael Banos, Oresti} {and}
  \bibinfo{person}{Alejandro Saez}.} \bibinfo{year}{2014}\natexlab{}.
\newblock \bibinfo{title}{{MHEALTH Dataset}}.
\newblock \bibinfo{howpublished}{UCI Machine Learning Repository}.
\newblock
\newblock
\shownote{{DOI}: https://doi.org/10.24432/C5TW22}.


\bibitem[Bischl et~al\mbox{.}(2023)]%
        {bischl2023hyperparameter}
\bibfield{author}{\bibinfo{person}{Bernd Bischl} {et~al\mbox{.}}}
  \bibinfo{year}{2023}\natexlab{}.
\newblock \showarticletitle{Hyperparameter optimization: Foundations,
  algorithms, best practices, and open challenges}.
\newblock \bibinfo{journal}{\emph{Wiley Interdisciplinary Reviews: Data Mining
  and Knowledge Discovery}} \bibinfo{volume}{13}, \bibinfo{number}{2}
  (\bibinfo{year}{2023}), \bibinfo{pages}{e1484}.
\newblock


\bibitem[Bricken et~al\mbox{.}(2023)]%
        {bricken2023sparse}
\bibfield{author}{\bibinfo{person}{Trenton Bricken} {et~al\mbox{.}}}
  \bibinfo{year}{2023}\natexlab{}.
\newblock \showarticletitle{Sparse Distributed Memory is a Continual Learner}.
  In \bibinfo{booktitle}{\emph{International Conference on Learning
  Representations}}.
\newblock


\bibitem[Cai et~al\mbox{.}(2020)]%
        {cai2020tinytl}
\bibfield{author}{\bibinfo{person}{Han Cai} {et~al\mbox{.}}}
  \bibinfo{year}{2020}\natexlab{}.
\newblock \showarticletitle{Tinytl: Reduce memory, not parameters for efficient
  on-device learning}.
\newblock \bibinfo{journal}{\emph{Advances in Neural Information Processing
  Systems}}  \bibinfo{volume}{33} (\bibinfo{year}{2020}),
  \bibinfo{pages}{11285--11297}.
\newblock


\bibitem[Chen et~al\mbox{.}(2016)]%
        {chen2016smart}
\bibfield{author}{\bibinfo{person}{Ning Chen} {et~al\mbox{.}}}
  \bibinfo{year}{2016}\natexlab{}.
\newblock \showarticletitle{Smart urban surveillance using fog computing}. In
  \bibinfo{booktitle}{\emph{2016 IEEE/ACM Symposium on Edge Computing (SEC)}}.
  IEEE, \bibinfo{pages}{95--96}.
\newblock


\bibitem[Dutta et~al\mbox{.}(2022)]%
        {dutta2022hdnn}
\bibfield{author}{\bibinfo{person}{Arpan Dutta} {et~al\mbox{.}}}
  \bibinfo{year}{2022}\natexlab{}.
\newblock \showarticletitle{Hdnn-pim: Efficient in memory design of
  hyperdimensional computing with feature extraction}. In
  \bibinfo{booktitle}{\emph{Proceedings of the Great Lakes Symposium on VLSI
  2022}}. \bibinfo{pages}{281--286}.
\newblock


\bibitem[Essa and Abdelmaksoud(2023)]%
        {essa2023temporal}
\bibfield{author}{\bibinfo{person}{Ehab Essa} {and} \bibinfo{person}{Islam~R
  Abdelmaksoud}.} \bibinfo{year}{2023}\natexlab{}.
\newblock \showarticletitle{Temporal-channel convolution with self-attention
  network for human activity recognition using wearable sensors}.
\newblock \bibinfo{journal}{\emph{Knowledge-Based Systems}}
  \bibinfo{volume}{278} (\bibinfo{year}{2023}), \bibinfo{pages}{110867}.
\newblock


\bibitem[et~al(2022)]%
        {madaan2022representational}
\bibfield{author}{\bibinfo{person}{Divyam~Madaan et al}.}
  \bibinfo{year}{2022}\natexlab{}.
\newblock \showarticletitle{Representational Continuity for Unsupervised
  Continual Learning}. In \bibinfo{booktitle}{\emph{International Conference on
  Learning Representations}}.
\newblock


\bibitem[Fini et~al\mbox{.}(2022)]%
        {fini2021self}
\bibfield{author}{\bibinfo{person}{Enrico Fini} {et~al\mbox{.}}}
  \bibinfo{year}{2022}\natexlab{}.
\newblock \showarticletitle{Self-supervised models are continual learners}. In
  \bibinfo{booktitle}{\emph{Proceedings of the IEEE/CVF Conference on Computer
  Vision and Pattern Recognition}}.
\newblock


\bibitem[Gim and Ko(2022)]%
        {gim2022memory}
\bibfield{author}{\bibinfo{person}{In Gim} {and} \bibinfo{person}{JeongGil
  Ko}.} \bibinfo{year}{2022}\natexlab{}.
\newblock \showarticletitle{Memory-efficient DNN training on mobile devices}.
  In \bibinfo{booktitle}{\emph{Proceedings of the 20th Annual International
  Conference on Mobile Systems, Applications and Services}}.
  \bibinfo{pages}{464--476}.
\newblock


\bibitem[Grill et~al\mbox{.}(2020)]%
        {grill2020bootstrap}
\bibfield{author}{\bibinfo{person}{Jean-Bastien Grill} {et~al\mbox{.}}}
  \bibinfo{year}{2020}\natexlab{}.
\newblock \showarticletitle{Bootstrap your own latent-a new approach to
  self-supervised learning}.
\newblock \bibinfo{journal}{\emph{Advances in neural information processing
  systems}}  \bibinfo{volume}{33} (\bibinfo{year}{2020}),
  \bibinfo{pages}{21271--21284}.
\newblock


\bibitem[Halko et~al\mbox{.}(2011)]%
        {halko2011finding}
\bibfield{author}{\bibinfo{person}{Nathan Halko}, \bibinfo{person}{Per-Gunnar
  Martinsson}, {and} \bibinfo{person}{Joel~A Tropp}.}
  \bibinfo{year}{2011}\natexlab{}.
\newblock \showarticletitle{Finding structure with randomness: Probabilistic
  algorithms for constructing approximate matrix decompositions}.
\newblock \bibinfo{journal}{\emph{SIAM review}} \bibinfo{volume}{53},
  \bibinfo{number}{2} (\bibinfo{year}{2011}), \bibinfo{pages}{217--288}.
\newblock


\bibitem[Hersche et~al\mbox{.}(2022)]%
        {hersche2022constrained}
\bibfield{author}{\bibinfo{person}{Michael Hersche} {et~al\mbox{.}}}
  \bibinfo{year}{2022}\natexlab{}.
\newblock \showarticletitle{Constrained few-shot class-incremental learning}.
  In \bibinfo{booktitle}{\emph{Proceedings of the IEEE/CVF Conference on
  Computer Vision and Pattern Recognition}}. \bibinfo{pages}{9057--9067}.
\newblock


\bibitem[Hioki(2023)]%
        {powermeter}
\bibfield{author}{\bibinfo{person}{Hioki}.} \bibinfo{year}{2023}\natexlab{}.
\newblock \bibinfo{title}{{Hioki3334 Powermeter}}.
\newblock
  \bibinfo{howpublished}{\url{https://www.hioki.com/en/products/detail/?product_key=5812}}.
\newblock


\bibitem[Howard et~al\mbox{.}(2019)]%
        {howard2019searching}
\bibfield{author}{\bibinfo{person}{Andrew Howard} {et~al\mbox{.}}}
  \bibinfo{year}{2019}\natexlab{}.
\newblock \showarticletitle{Searching for mobilenetv3}. In
  \bibinfo{booktitle}{\emph{Proceedings of the IEEE/CVF International
  Conference on Computer Vision}}. \bibinfo{pages}{1314--1324}.
\newblock


\bibitem[Imani et~al\mbox{.}(2019a)]%
        {imani2019hdcluster}
\bibfield{author}{\bibinfo{person}{Mohsen Imani} {et~al\mbox{.}}}
  \bibinfo{year}{2019}\natexlab{a}.
\newblock \showarticletitle{Hdcluster: An accurate clustering using
  brain-inspired high-dimensional computing}. In
  \bibinfo{booktitle}{\emph{Design, Automation \& Test in Europe Conference \&
  Exhibition (DATE)}}. IEEE, \bibinfo{pages}{1591--1594}.
\newblock


\bibitem[Imani et~al\mbox{.}(2019b)]%
        {imani2019semihd}
\bibfield{author}{\bibinfo{person}{Mohsen Imani} {et~al\mbox{.}}}
  \bibinfo{year}{2019}\natexlab{b}.
\newblock \showarticletitle{Semihd: Semi-supervised learning using
  hyperdimensional computing}. In \bibinfo{booktitle}{\emph{IEEE/ACM
  International Conference on Computer-Aided Design (ICCAD)}}. IEEE,
  \bibinfo{pages}{1--8}.
\newblock


\bibitem[Imani et~al\mbox{.}(2017)]%
        {imani2017voicehd}
\bibfield{author}{\bibinfo{person}{Mohsen Imani}, \bibinfo{person}{Deqian
  Kong}, \bibinfo{person}{Abbas Rahimi}, {and} \bibinfo{person}{Tajana
  Rosing}.} \bibinfo{year}{2017}\natexlab{}.
\newblock \showarticletitle{Voicehd: Hyperdimensional computing for efficient
  speech recognition}. In \bibinfo{booktitle}{\emph{IEEE International
  Conference on Rebooting Computing (ICRC)}}. IEEE, \bibinfo{pages}{1--8}.
\newblock


\bibitem[Kanerva(2009)]%
        {kanerva2009hyperdimensional}
\bibfield{author}{\bibinfo{person}{Pentti Kanerva}.}
  \bibinfo{year}{2009}\natexlab{}.
\newblock \showarticletitle{Hyperdimensional computing: An introduction to
  computing in distributed representation with high-dimensional random
  vectors}.
\newblock \bibinfo{journal}{\emph{Cognitive Computation}}  \bibinfo{volume}{1}
  (\bibinfo{year}{2009}), \bibinfo{pages}{139--159}.
\newblock


\bibitem[Khaleghi et~al\mbox{.}(2020)]%
        {khaleghi2020prive}
\bibfield{author}{\bibinfo{person}{Behnam Khaleghi}, \bibinfo{person}{Mohsen
  Imani}, {and} \bibinfo{person}{Tajana Rosing}.}
  \bibinfo{year}{2020}\natexlab{}.
\newblock \showarticletitle{Prive-hd: Privacy-preserved hyperdimensional
  computing}. In \bibinfo{booktitle}{\emph{ACM/IEEE Design Automation
  Conference (DAC)}}. IEEE, \bibinfo{pages}{1--6}.
\newblock


\bibitem[Kim et~al\mbox{.}(2019)]%
        {kim2019efficient}
\bibfield{author}{\bibinfo{person}{Hyeji Kim}, \bibinfo{person}{Muhammad
  Umar~Karim Khan}, {and} \bibinfo{person}{Chong-Min Kyung}.}
  \bibinfo{year}{2019}\natexlab{}.
\newblock \showarticletitle{Efficient neural network compression}. In
  \bibinfo{booktitle}{\emph{Proceedings of the IEEE/CVF conference on computer
  vision and pattern recognition}}. \bibinfo{pages}{12569--12577}.
\newblock


\bibitem[Kim et~al\mbox{.}(2018)]%
        {kim2018efficient}
\bibfield{author}{\bibinfo{person}{Yeseong Kim}, \bibinfo{person}{Mohsen
  Imani}, {and} \bibinfo{person}{Tajana~S Rosing}.}
  \bibinfo{year}{2018}\natexlab{}.
\newblock \showarticletitle{Efficient human activity recognition using
  hyperdimensional computing}. In \bibinfo{booktitle}{\emph{Proceedings of the
  8th International Conference on the Internet of Things}}.
  \bibinfo{pages}{1--6}.
\newblock


\bibitem[Kirkpatrick et~al\mbox{.}(2017)]%
        {kirkpatrick2017overcoming}
\bibfield{author}{\bibinfo{person}{James Kirkpatrick} {et~al\mbox{.}}}
  \bibinfo{year}{2017}\natexlab{}.
\newblock \showarticletitle{Overcoming catastrophic forgetting in neural
  networks}.
\newblock \bibinfo{journal}{\emph{Proceedings of the national academy of
  sciences}} (\bibinfo{year}{2017}).
\newblock


\bibitem[Krizhevsky et~al\mbox{.}(2009)]%
        {krizhevsky2009learning}
\bibfield{author}{\bibinfo{person}{Alex Krizhevsky}, \bibinfo{person}{Geoffrey
  Hinton}, {et~al\mbox{.}}} \bibinfo{year}{2009}\natexlab{}.
\newblock \showarticletitle{Learning multiple layers of features from tiny
  images}.
\newblock  (\bibinfo{year}{2009}).
\newblock


\bibitem[Kwon~et al(2023)]%
        {kwon2023lifelearner}
\bibfield{author}{\bibinfo{person}{Young~D Kwon~et al}.}
  \bibinfo{year}{2023}\natexlab{}.
\newblock \showarticletitle{LifeLearner: Hardware-Aware Meta Continual Learning
  System for Embedded Computing Platforms}. In
  \bibinfo{booktitle}{\emph{Proceedings of the 21st ACM Conference on Embedded
  Networked Sensor Systems}}.
\newblock


\bibitem[Lee et~al\mbox{.}(2020)]%
        {Lee2020A}
\bibfield{author}{\bibinfo{person}{Soochan Lee} {et~al\mbox{.}}}
  \bibinfo{year}{2020}\natexlab{}.
\newblock \showarticletitle{A Neural Dirichlet Process Mixture Model for
  Task-Free Continual Learning}. In \bibinfo{booktitle}{\emph{International
  Conference on Learning Representations}}.
\newblock


\bibitem[Lin et~al\mbox{.}(2020)]%
        {lin2020mcunet}
\bibfield{author}{\bibinfo{person}{Ji Lin} {et~al\mbox{.}}}
  \bibinfo{year}{2020}\natexlab{}.
\newblock \showarticletitle{Mcunet: Tiny deep learning on iot devices}.
\newblock \bibinfo{journal}{\emph{Advances in Neural Information Processing
  Systems}}  \bibinfo{volume}{33} (\bibinfo{year}{2020}),
  \bibinfo{pages}{11711--11722}.
\newblock


\bibitem[Lin et~al\mbox{.}(2021)]%
        {lin2021memory}
\bibfield{author}{\bibinfo{person}{Ji Lin} {et~al\mbox{.}}}
  \bibinfo{year}{2021}\natexlab{}.
\newblock \showarticletitle{Memory-efficient patch-based inference for tiny
  deep learning}.
\newblock \bibinfo{journal}{\emph{Advances in Neural Information Processing
  Systems}}  \bibinfo{volume}{34} (\bibinfo{year}{2021}),
  \bibinfo{pages}{2346--2358}.
\newblock


\bibitem[Lin et~al\mbox{.}(2022)]%
        {lin2022device}
\bibfield{author}{\bibinfo{person}{Ji Lin} {et~al\mbox{.}}}
  \bibinfo{year}{2022}\natexlab{}.
\newblock \showarticletitle{On-device training under 256kb memory}.
\newblock \bibinfo{journal}{\emph{Advances in Neural Information Processing
  Systems}}  \bibinfo{volume}{35} (\bibinfo{year}{2022}),
  \bibinfo{pages}{22941--22954}.
\newblock


\bibitem[Lopez-Paz and Ranzato(2017)]%
        {lopez2017gradient}
\bibfield{author}{\bibinfo{person}{David Lopez-Paz} {and}
  \bibinfo{person}{Marc'Aurelio Ranzato}.} \bibinfo{year}{2017}\natexlab{}.
\newblock \showarticletitle{Gradient episodic memory for continual learning}.
\newblock \bibinfo{journal}{\emph{Advances in neural information processing
  systems}}  \bibinfo{volume}{30} (\bibinfo{year}{2017}).
\newblock


\bibitem[McCloskey and Cohen(1989)]%
        {mccloskey1989catastrophic}
\bibfield{author}{\bibinfo{person}{Michael McCloskey} {and}
  \bibinfo{person}{Neal~J Cohen}.} \bibinfo{year}{1989}\natexlab{}.
\newblock \showarticletitle{Catastrophic interference in connectionist
  networks: The sequential learning problem}.
\newblock In \bibinfo{booktitle}{\emph{Psychology of learning and motivation}}.
  Vol.~\bibinfo{volume}{24}. \bibinfo{publisher}{Elsevier},
  \bibinfo{pages}{109--165}.
\newblock


\bibitem[Mohaimenuzzaman et~al\mbox{.}(2023)]%
        {mohaimenuzzaman2023environmental}
\bibfield{author}{\bibinfo{person}{Md Mohaimenuzzaman} {et~al\mbox{.}}}
  \bibinfo{year}{2023}\natexlab{}.
\newblock \showarticletitle{Environmental Sound Classification on the Edge: A
  Pipeline for Deep Acoustic Networks on Extremely Resource-Constrained
  Devices}.
\newblock \bibinfo{journal}{\emph{Pattern Recognition}}  \bibinfo{volume}{133}
  (\bibinfo{year}{2023}), \bibinfo{pages}{109025}.
\newblock


\bibitem[Moin et~al\mbox{.}(2021)]%
        {moin2021wearable}
\bibfield{author}{\bibinfo{person}{Ali Moin} {et~al\mbox{.}}}
  \bibinfo{year}{2021}\natexlab{}.
\newblock \showarticletitle{A wearable biosensing system with in-sensor
  adaptive machine learning for hand gesture recognition}.
\newblock \bibinfo{journal}{\emph{Nature Electronics}} \bibinfo{volume}{4},
  \bibinfo{number}{1} (\bibinfo{year}{2021}), \bibinfo{pages}{54--63}.
\newblock


\bibitem[Neill(2020)]%
        {neill2020overview}
\bibfield{author}{\bibinfo{person}{James~O' Neill}.}
  \bibinfo{year}{2020}\natexlab{}.
\newblock \showarticletitle{An overview of neural network compression}.
\newblock \bibinfo{journal}{\emph{arXiv preprint arXiv:2006.03669}}
  (\bibinfo{year}{2020}).
\newblock


\bibitem[Ng et~al\mbox{.}(2001)]%
        {ng2001spectral}
\bibfield{author}{\bibinfo{person}{Andrew Ng}, \bibinfo{person}{Michael
  Jordan}, {and} \bibinfo{person}{Yair Weiss}.}
  \bibinfo{year}{2001}\natexlab{}.
\newblock \showarticletitle{On spectral clustering: Analysis and an algorithm}.
\newblock \bibinfo{journal}{\emph{Advances in neural information processing
  systems}}  \bibinfo{volume}{14} (\bibinfo{year}{2001}).
\newblock


\bibitem[Osipov et~al\mbox{.}(2022)]%
        {osipov2022hyperseed}
\bibfield{author}{\bibinfo{person}{Evgeny Osipov} {et~al\mbox{.}}}
  \bibinfo{year}{2022}\natexlab{}.
\newblock \showarticletitle{Hyperseed: Unsupervised learning with vector
  symbolic architectures}.
\newblock \bibinfo{journal}{\emph{IEEE Transactions on Neural Networks and
  Learning Systems}} (\bibinfo{year}{2022}).
\newblock


\bibitem[Parisi et~al\mbox{.}(2019)]%
        {parisi2019continual}
\bibfield{author}{\bibinfo{person}{German~I Parisi}, \bibinfo{person}{Ronald
  Kemker}, \bibinfo{person}{Jose~L Part}, \bibinfo{person}{Christopher Kanan},
  {and} \bibinfo{person}{Stefan Wermter}.} \bibinfo{year}{2019}\natexlab{}.
\newblock \showarticletitle{Continual lifelong learning with neural networks: A
  review}.
\newblock \bibinfo{journal}{\emph{Neural networks}}  \bibinfo{volume}{113}
  (\bibinfo{year}{2019}), \bibinfo{pages}{54--71}.
\newblock


\bibitem[Paszke et~al\mbox{.}(2019)]%
        {paszke2019pytorch}
\bibfield{author}{\bibinfo{person}{Adam Paszke} {et~al\mbox{.}}}
  \bibinfo{year}{2019}\natexlab{}.
\newblock \showarticletitle{Pytorch: An imperative style, high-performance deep
  learning library}.
\newblock \bibinfo{journal}{\emph{Advances in neural information processing
  systems}}  \bibinfo{volume}{32} (\bibinfo{year}{2019}).
\newblock


\bibitem[Piczak(2015)]%
        {piczak2015esc}
\bibfield{author}{\bibinfo{person}{Karol~J Piczak}.}
  \bibinfo{year}{2015}\natexlab{}.
\newblock \showarticletitle{ESC: Dataset for environmental sound
  classification}. In \bibinfo{booktitle}{\emph{Proceedings of the 23rd ACM
  international conference on Multimedia}}. \bibinfo{pages}{1015--1018}.
\newblock


\bibitem[Profentzas et~al\mbox{.}(2022)]%
        {profentzas2022minilearn}
\bibfield{author}{\bibinfo{person}{Christos Profentzas},
  \bibinfo{person}{Magnus Almgren}, {and} \bibinfo{person}{Olaf Landsiedel}.}
  \bibinfo{year}{2022}\natexlab{}.
\newblock \showarticletitle{MiniLearn: On-Device Learning for Low-Power IoT
  Devices}. In \bibinfo{booktitle}{\emph{International Conference on Embedded
  Wireless Systems and Networks}}.
\newblock


\bibitem[Rao et~al\mbox{.}(2019)]%
        {rao2019continual}
\bibfield{author}{\bibinfo{person}{Dushyant Rao}, \bibinfo{person}{Francesco
  Visin}, \bibinfo{person}{Andrei Rusu}, \bibinfo{person}{Razvan Pascanu},
  \bibinfo{person}{Yee~Whye Teh}, {and} \bibinfo{person}{Raia Hadsell}.}
  \bibinfo{year}{2019}\natexlab{}.
\newblock \showarticletitle{Continual unsupervised representation learning}.
\newblock \bibinfo{journal}{\emph{Advances in neural information processing
  systems}}  \bibinfo{volume}{32} (\bibinfo{year}{2019}).
\newblock


\bibitem[Ren et~al\mbox{.}(2021)]%
        {ren2021tinyol}
\bibfield{author}{\bibinfo{person}{Haoyu Ren}, \bibinfo{person}{Darko Anicic},
  {and} \bibinfo{person}{Thomas~A Runkler}.} \bibinfo{year}{2021}\natexlab{}.
\newblock \showarticletitle{Tinyol: Tinyml with online-learning on
  microcontrollers}. In \bibinfo{booktitle}{\emph{2021 International Joint
  Conference on Neural Networks (IJCNN)}}. IEEE, \bibinfo{pages}{1--8}.
\newblock


\bibitem[Russakovsky et~al\mbox{.}(2015)]%
        {russakovsky2015imagenet}
\bibfield{author}{\bibinfo{person}{Olga Russakovsky} {et~al\mbox{.}}}
  \bibinfo{year}{2015}\natexlab{}.
\newblock \showarticletitle{Imagenet large scale visual recognition challenge}.
\newblock \bibinfo{journal}{\emph{International journal of computer vision}}
  \bibinfo{volume}{115} (\bibinfo{year}{2015}), \bibinfo{pages}{211--252}.
\newblock


\bibitem[Rusu~et al(2016)]%
        {rusu2016progressive}
\bibfield{author}{\bibinfo{person}{Andrei~A Rusu~et al}.}
  \bibinfo{year}{2016}\natexlab{}.
\newblock \showarticletitle{Progressive neural networks}.
\newblock \bibinfo{journal}{\emph{arXiv preprint arXiv:1606.04671}}
  (\bibinfo{year}{2016}).
\newblock


\bibitem[Saha et~al\mbox{.}(2023)]%
        {saha2023tinyns}
\bibfield{author}{\bibinfo{person}{Swapnil~Sayan Saha} {et~al\mbox{.}}}
  \bibinfo{year}{2023}\natexlab{}.
\newblock \showarticletitle{TinyNS: Platform-Aware Neurosymbolic Auto Tiny
  Machine Learning}.
\newblock \bibinfo{journal}{\emph{ACM Transactions on Embedded Computing
  Systems}} (\bibinfo{year}{2023}).
\newblock


\bibitem[Sandler et~al\mbox{.}(2018)]%
        {sandler2018mobilenetv2}
\bibfield{author}{\bibinfo{person}{Mark Sandler} {et~al\mbox{.}}}
  \bibinfo{year}{2018}\natexlab{}.
\newblock \showarticletitle{Mobilenetv2: Inverted residuals and linear
  bottlenecks}. In \bibinfo{booktitle}{\emph{Proceedings of the IEEE conference
  on computer vision and pattern recognition}}. \bibinfo{pages}{4510--4520}.
\newblock


\bibitem[Shen et~al\mbox{.}(2021)]%
        {shen2021algorithmic}
\bibfield{author}{\bibinfo{person}{Yang Shen}, \bibinfo{person}{Sanjoy
  Dasgupta}, {and} \bibinfo{person}{Saket Navlakha}.}
  \bibinfo{year}{2021}\natexlab{}.
\newblock \showarticletitle{Algorithmic insights on continual learning from
  fruit flies}.
\newblock \bibinfo{journal}{\emph{arXiv preprint arXiv:2107.07617}}
  (\bibinfo{year}{2021}).
\newblock


\bibitem[Shunhou and Peng(2022)]%
        {shunhou2022aiot}
\bibfield{author}{\bibinfo{person}{Shun Shunhou} {and} \bibinfo{person}{Yang
  Peng}.} \bibinfo{year}{2022}\natexlab{}.
\newblock \showarticletitle{AIoT on Cloud}.
\newblock In \bibinfo{booktitle}{\emph{Digital Transformation in Cloud
  Computing}}. \bibinfo{publisher}{CRC Press}, \bibinfo{pages}{629--732}.
\newblock


\bibitem[Smith et~al\mbox{.}(2021)]%
        {smith2019unsupervised}
\bibfield{author}{\bibinfo{person}{James Smith} {et~al\mbox{.}}}
  \bibinfo{year}{2021}\natexlab{}.
\newblock \showarticletitle{Unsupervised Progressive Learning and the STAM
  Architecture}. In \bibinfo{booktitle}{\emph{Proceedings of the Thirtieth
  International Joint Conference on Artificial Intelligence, {IJCAI-21}}}.
  \bibinfo{pages}{2979--2987}.
\newblock


\bibitem[Sun et~al\mbox{.}(2020)]%
        {sun2020alexa}
\bibfield{author}{\bibinfo{person}{Ke Sun}, \bibinfo{person}{Chen Chen}, {and}
  \bibinfo{person}{Xinyu Zhang}.} \bibinfo{year}{2020}\natexlab{}.
\newblock \showarticletitle{" Alexa, stop spying on me!" speech privacy
  protection against voice assistants}. In
  \bibinfo{booktitle}{\emph{Proceedings of the 18th conference on Embedded
  Networked Sensor Systems}}. \bibinfo{pages}{298--311}.
\newblock


\bibitem[Thomas et~al\mbox{.}(2021)]%
        {thomas2021theoretical}
\bibfield{author}{\bibinfo{person}{Anthony Thomas}, \bibinfo{person}{Sanjoy
  Dasgupta}, {and} \bibinfo{person}{Tajana Rosing}.}
  \bibinfo{year}{2021}\natexlab{}.
\newblock \showarticletitle{A theoretical perspective on hyperdimensional
  computing}.
\newblock \bibinfo{journal}{\emph{Journal of Artificial Intelligence Research}}
   \bibinfo{volume}{72} (\bibinfo{year}{2021}), \bibinfo{pages}{215--249}.
\newblock


\bibitem[Tiezzi et~al\mbox{.}(2022)]%
        {ijcai2022p0483}
\bibfield{author}{\bibinfo{person}{Matteo Tiezzi} {et~al\mbox{.}}}
  \bibinfo{year}{2022}\natexlab{}.
\newblock \showarticletitle{Stochastic Coherence Over Attention Trajectory For
  Continuous Learning In Video Streams}. In
  \bibinfo{booktitle}{\emph{Proceedings of the Thirty-First International Joint
  Conference on Artificial Intelligence, {IJCAI-22}}}.
  \bibinfo{pages}{3480--3486}.
\newblock


\bibitem[Tiwari et~al\mbox{.}(2022)]%
        {tiwari2022gcr}
\bibfield{author}{\bibinfo{person}{Rishabh Tiwari} {et~al\mbox{.}}}
  \bibinfo{year}{2022}\natexlab{}.
\newblock \showarticletitle{Gcr: Gradient coreset based replay buffer selection
  for continual learning}. In \bibinfo{booktitle}{\emph{Proceedings of the
  IEEE/CVF Conference on Computer Vision and Pattern Recognition}}.
  \bibinfo{pages}{99--108}.
\newblock


\bibitem[Von~Luxburg(2007)]%
        {von2007tutorial}
\bibfield{author}{\bibinfo{person}{Ulrike Von~Luxburg}.}
  \bibinfo{year}{2007}\natexlab{}.
\newblock \showarticletitle{A tutorial on spectral clustering}.
\newblock \bibinfo{journal}{\emph{Statistics and computing}}
  \bibinfo{volume}{17} (\bibinfo{year}{2007}), \bibinfo{pages}{395--416}.
\newblock


\bibitem[Wang et~al\mbox{.}(2019)]%
        {wang2019deep}
\bibfield{author}{\bibinfo{person}{Erwei Wang} {et~al\mbox{.}}}
  \bibinfo{year}{2019}\natexlab{}.
\newblock \showarticletitle{Deep neural network approximation for custom
  hardware: Where we've been, where we're going}.
\newblock \bibinfo{journal}{\emph{ACM Computing Surveys (CSUR)}}
  \bibinfo{volume}{52}, \bibinfo{number}{2} (\bibinfo{year}{2019}),
  \bibinfo{pages}{1--39}.
\newblock


\bibitem[Wang et~al\mbox{.}(2022)]%
        {wang2022melon}
\bibfield{author}{\bibinfo{person}{Qipeng Wang} {et~al\mbox{.}}}
  \bibinfo{year}{2022}\natexlab{}.
\newblock \showarticletitle{Melon: Breaking the memory wall for
  resource-efficient on-device machine learning}. In
  \bibinfo{booktitle}{\emph{Proceedings of the 20th Annual International
  Conference on Mobile Systems, Applications and Services}}.
  \bibinfo{pages}{450--463}.
\newblock


\bibitem[Weiss et~al\mbox{.}(2016)]%
        {weiss2016smartwatch}
\bibfield{author}{\bibinfo{person}{Gary~M Weiss} {et~al\mbox{.}}}
  \bibinfo{year}{2016}\natexlab{}.
\newblock \showarticletitle{Smartwatch-based activity recognition: A machine
  learning approach}. In \bibinfo{booktitle}{\emph{2016 IEEE-EMBS International
  Conference on Biomedical and Health Informatics (BHI)}}. IEEE,
  \bibinfo{pages}{426--429}.
\newblock


\bibitem[Xie et~al\mbox{.}(2016)]%
        {xie2016unsupervised}
\bibfield{author}{\bibinfo{person}{Junyuan Xie}, \bibinfo{person}{Ross
  Girshick}, {and} \bibinfo{person}{Ali Farhadi}.}
  \bibinfo{year}{2016}\natexlab{}.
\newblock \showarticletitle{Unsupervised deep embedding for clustering
  analysis}. In \bibinfo{booktitle}{\emph{International Conference on Machine
  Learning}}. PMLR, \bibinfo{pages}{478--487}.
\newblock


\bibitem[Xu et~al\mbox{.}(2022)]%
        {xu2022mandheling}
\bibfield{author}{\bibinfo{person}{Daliang Xu} {et~al\mbox{.}}}
  \bibinfo{year}{2022}\natexlab{}.
\newblock \showarticletitle{Mandheling: Mixed-precision on-device dnn training
  with dsp offloading}. In \bibinfo{booktitle}{\emph{Proceedings of the 28th
  Annual International Conference on Mobile Computing And Networking}}.
  \bibinfo{pages}{214--227}.
\newblock


\bibitem[Xu et~al\mbox{.}(2023)]%
        {xu2023fsl}
\bibfield{author}{\bibinfo{person}{Weihong Xu}, \bibinfo{person}{Jaeyoung
  Kang}, {and} \bibinfo{person}{Tajana Rosing}.}
  \bibinfo{year}{2023}\natexlab{}.
\newblock \showarticletitle{FSL-HD: Accelerating Few-Shot Learning on ReRAM
  using Hyperdimensional Computing}. In \bibinfo{booktitle}{\emph{2023 Design,
  Automation \& Test in Europe Conference \& Exhibition (DATE)}}. IEEE,
  \bibinfo{pages}{1--6}.
\newblock


\bibitem[Zhang et~al\mbox{.}(2020a)]%
        {zhang2020class}
\bibfield{author}{\bibinfo{person}{Junting Zhang} {et~al\mbox{.}}}
  \bibinfo{year}{2020}\natexlab{a}.
\newblock \showarticletitle{Class-incremental learning via deep model
  consolidation}. In \bibinfo{booktitle}{\emph{Proceedings of the IEEE/CVF
  Winter Conference on Applications of Computer Vision}}.
  \bibinfo{pages}{1131--1140}.
\newblock


\bibitem[Zhang et~al\mbox{.}(2020b)]%
        {zhang2020mdldroidlite}
\bibfield{author}{\bibinfo{person}{Yu Zhang}, \bibinfo{person}{Tao Gu}, {and}
  \bibinfo{person}{Xi Zhang}.} \bibinfo{year}{2020}\natexlab{b}.
\newblock \showarticletitle{MDLdroidLite: A release-and-inhibit control
  approach to resource-efficient deep neural networks on mobile devices}. In
  \bibinfo{booktitle}{\emph{Proceedings of the 18th Conference on Embedded
  Networked Sensor Systems}}. \bibinfo{pages}{463--475}.
\newblock


\end{thebibliography}
